\documentclass[runningheads,a4paper]{llncs}
\usepackage{amsmath}
\usepackage{amssymb}
\usepackage{array,multirow,graphicx}
\usepackage{url}
\usepackage{apacite}
\usepackage{subcaption} 
\usepackage{csquotes}
\usepackage[table,xcdraw]{xcolor}
\usepackage{lscape}
\usepackage{amsmath}
\usepackage[figuresright]{rotating}
\usepackage{eucal}
\usepackage{array}
\usepackage{makecell}
\usepackage{booktabs}
\usepackage{adjustbox}
\usepackage{graphicx}
\usepackage{ragged2e}
\graphicspath{ {./figures/} }
\usepackage{float}
\newcommand{\keywords}[1]{\par\addvspace\baselineskip
\noindent\keywordname\enspace\ignorespaces#1}

\usepackage{adjustbox}
\usepackage{booktabs}
\usepackage{comment}
\usepackage{xcolor}

\usepackage{acronym}
\usepackage[utf8]{inputenc}
\usepackage{xspace}
	
\newcommand*{\ie}		{i.e.,\ }

\acrodef{cnn} 		[\textsc{CNN\xspace}]				{Convolutional Neural Network}
\acrodef{resnet} 		[\textsc{ResNet\xspace}]				{residual neural network}
\acrodef{lstm}		[\textsc{LSTM\xspace}]				{Long Short-Term Memory Networks}
\acrodef{lrcn}		[\textsc{LRCN\xspace}]				{Long-term Recurrent Convolutional Networks}
\acrodef{svm}		[\textsc{SVM\xspace}]				{Support Vector Machine}
\acrodef{dbb}		[\textsc{DBB\xspace}]				{Dynamic Batch Balance}
\acrodef{wma}		[\textsc{WMA\xspace}]				{Weighted Moving Average}

\acrodef{stft}		[\textsc{STFT\xspace}]				{Short Time Fourier transform}
\acrodef{mfcc}		[\textsc{MFCC\xspace}]				{Mel-frequency cepstral coefficient}
\acrodef{mir}		[\textsc{MIR\xspace}]				{Music Information Retrieval}
\acrodef{zcr}		[\textsc{ZCR\xspace}]				{Zero-Crossing Rate}

\acrodef{relu}		[\textsc{ReLU\xspace}]				{Rectified Linear Unit}
\acrodef{tp}		[\textsc{TP\xspace}]				{True Positives}
\acrodef{tn}		[\textsc{TN\xspace}]				{True Negatives}
\acrodef{fp}		[\textsc{FP\xspace}]				{False Positive}
\acrodef{fn}		[\textsc{FN\xspace}]				{False Negative}

\begin{document}

\mainmatter  % start of an individual contribution

\title {Deep Neural Network approaches for Analysing Videos of Music Performances }

\titlerunning{Running Title}

%\author{AuthorA\inst{1}\and AuthorB\inst{2} \thanks{Please place acknowledgement here.}}
%Prakash Chandra Chhipa, Richa Upadhya,  Killian Murphy, Federico Visi,  Stefan Östersjö, Marcus Liwicki
\author{Foteini Simistira Liwicki\inst{1}, Richa Upadhyay\inst{1}, Prakash Chandra Chhipa\inst{1},   Killian Murphy\inst{2}, Federico Visi\inst{3},  Stefan Östersjö\inst{3} \and Marcus Liwicki\inst{1}}
%UPADHYAY
% if the names of the authors are too long for the running head, please use the format: AuthorA et al.
%\authorrunning{AuthorA and AuthorB (or AuthorA et al. if too long)}

\institute{Machine Learning Group, Luleå University of Technology, Sweden \and SAMOVAR laboratory, Télécom SudParis, Institut Polytechnique de Paris, France \and GEMM-Gesture Embodiment and Machines in Music, Luleå University of Technology, Sweden\\ \email{foteini.liwicki@ltu.se}}

% killian.murphy@telecom-sudparis.eu


\maketitle

\begin{abstract}
This paper presents a framework to automate the labelling process for gestures in musical performance videos with a 3D \ac{cnn}. 
While this idea was proposed in a previous study, this paper introduces several novelties:
(i) Presents a novel method to overcome the class imbalance challenge and make learning possible for co-existent gestures by batch balancing approach and spatial-temporal representations of gestures. 
(ii) Performs a detailed study on 7 and 18 categories of gestures generated during the performance (guitar play) of musical pieces that have been video-recorded. 
(iii) Investigates the possibility to use audio features.
(iv) Extends the analysis to multiple videos.
%The current work is an extension of a previous work in (Anonymous, 2020), which uses a multi-model approach on only video information and a pretrained 3D \ac{resnet} to recognize the categories of gestures and obtained an F1 score of 0.386 on the validation set, with the possibility of performance generalization. 
The novel methods significantly improve the performance of gesture identification by 12\,\%, when compared to the previous work (51\,\% in this study over 39\,\% in previous work).
We successfully validate the proposed methods on 7 super classes (72\,\%), an ensemble of the 18 gestures/classes, and additional videos (75\,\%).

%\textit{abstract} environment.
\end{abstract}
%\important{Ensuring Blind Review Authors should avoid explicit reference to their names and their published work in any part of the version submitted for review.}\\\\
%\important{Given that ‘Papers should be expanded to the length and quality of a journal article’, I think something in the order of 8,000 words would be appropriate (i.e, it should represent a clear expansion of the conference version). Page-count will vary according to the number of figures/tables, etc.}

\section{Introduction}
%\todo{[everyone can put text here] } 

This paper presents a set of experiments aimed at identifying expressive gestures in music performance videos by means of machine learning methods. 

%The study was carried out by the AI group at Luleå University of Technology in collaboration with researchers of Gesture Embodiment and Machines in Music (GEMM), a cross-disciplinary research cluster, also at the Luleå University of Technology.

Understanding the role of gestures in music performance is an open challenge, made more difficult by the variability of musicians' movements and the time and expertise required to analyse and label video recordings of musicians playing. To facilitate research on gestures, recordings of performances are labelled with the corresponding gestures~\cite{macritchie2013inferring,Ostersjo2019}. These tasks often require many hours of human labor. Using deep learning methods to automate labelling and aid human expert annotators might improve the labelling process, and is a less explored research field.

%there is only little research on automating such labeling by machine learning and deep learning methods.

The current work is an extension of a previous work in~\cite{anonym}, which uses a simple \ac{cnn} on only video information to recognize the categories of gestures and obtained an F1 score of 39\,\% on the validation set, with the possibility of performance generalization. 
This paper introduces several novel contributions:
\begin{itemize}
    \item A novel method to overcome the class imbalance challenge is introduced that makes learning possible for co-existent gestures via batch balancing approach and spatial-temporal representations of gestures. 
    The novel methods significantly improve the performance (F-measure) of gesture identification by 12\,\%, when compared to the previous work (51\,\% in this study over 39\,\% in previous work).
    \item It performs a detailed study on 7 and 18 categories of gestures generated during the performance (guitar play) of musical pieces that have been video-recorded. 
    Here, an F-measure of 72\,\% is achieved.
    \item It investigates the possibility of using audio features. It is evident from the literature \cite{hara3dcnns} that, for the purpose of recognizing gestures or movements of an individual in a task, video or images have a very significant role. 
    The learning algorithms perform well for such tasks when fed with the video frames, ostensibly because identifying movements relies on visual information. 
    In this particular application, music (\ie audio) is expected to be connected to the gestures involved in its performance.
    Hence, audio can be used as added information to estimate the gesture labels in the videos. This work also analyses a multi-modal system that utilizes video as well as audio content to classify gestures.
    \item It extends the analysis to multiple videos, achieving an F-measure of 75\,\%.
\end{itemize}

%(something related to multi-modal system)

%\todo{it would be useful to have an overview of the paper structure here}

The remainder of this paper is organized as follows. 
Section \ref{sota} presents the state of the art in video analysis and music performance analysis. 
Section \ref{dataset} presents the data, and Section \ref{methodology} illustrates the various methods employed in this work. 
Section \ref{implementation} elaborates on the implementation of the methods for gesture identification. 
Section \ref{discussion} presents a discussion of the results. 
Finally, the conclusion is in Section \ref{conclusion}. 
\section{State of the Art}  \label{sota}
\subsection{Video Analysis}
%\todo{[Federico] I used the text you wrote for the ICPR version of the paper (was rejected). Feel free to make changes or additions.}  \\

Image and video analysis with machine learning and deep learning architectures has shown very promising results.
One of the first major breakthroughs of deep learning techniques was in the field of image analysis.
ImageNet \cite{deng2009imagenet}, a frequently used large publicly available dataset, is a typical benchmark for image recognition tasks.

Since the early 2010s, the use of CNN architectures resulted in significantly better performance in classifying large image datasets such as ImageNet \cite{NIPS2012_4824}. 
The performance of such architectures improved year after year~\cite{schmidhuber2015deep,goodfellow2016deep}.

Building on the rapid success in image analysis, methods based on \ac{cnn} have been quickly adapted to neighboring fields, especially video analysis.
Machine learning techniques for video analysis have emerged in the field of action recognition in video streams, with several small datasets available, like HMDB-51 \cite{jhuang2011large} and UCF-101 \cite{soomro2012ucf101}.
Lately, larger publicly available datasets have been made available, like ActivityNet \cite{caba2015activitynet} and Kinetics \cite{carreira2017quo}. 
In the following paragraphs, we will provide a summary of the works that introduced the architectures used in this paper.
For a complete state-of-the-art overview, refer to~\cite{rodriguez2019video}.

In 2014, \cite{karpathy2014large} \ac{cnn}s have been successfully used for classifying 1 million YouTube videos with 487 classes (63.9\% accuracy) and were also applied to the UCF-101 dataset (63.3\%).
More complex architectures were designed to integrate \ac{lstm}{\cite{hochreiter1997long}} into \ac{cnn} models. The \ac{lrcn}~\cite{donahue2015long} achieved an accuracy of 87.6\% on the UCF-101 dataset.
Other advanced architectures use two-stream ConvNets fused with \ac{svm}~\cite{simonyan2014two} (88.0\% on UCF-101, 59.4\% on HMDB-51).

The most relevant related work uses 3D-\ac{cnn} architectures (called space-time \ac{cnn}s with long-term temporal convolutions)~\cite{varol2017long}, they achieve 92.7\% on UCF-101 and 67.2\% on HMDB-51.
%- CNN and deep bidirectional \cite{ullah2017action},  91.21\% on UCF-101, 87.64\% on HMDB-51.\\
%- \cite{carreira2017quo} they propose a two-stream inflated 3D ConvNet, average accuracy of 80.8\% and top-1 accuracy of 71.6\% on the Kinetics dataset, 80.2\% on HMDB-51 and 97.9\% on UCF-101.
For action detection in video streams, such a 3D variant of the \ac{cnn}s has been proven to be more efficient than the 2D variant, i.e.,~\cite{hara3dcnns} reports an accuracy of 94.5\% on the UCF-101, 70.2\% on the HMDB-51, and 78.4\% on the Kinetics datasets.
This architecture is used as the underlying base for the architecture presented in this paper.

\subsection{Music Performance Analysis}
%\todo{[ML group] Feel free to make changes or additions.}  \\
Performance analysis emerged as a field of study in music psychology and musicology in the late 1980s, with a focus in a wide range of perspectives, including the role of gesture, performer-performer interaction, social contexts, etc.~\cite{bowen1996performance}.
Theories of embodied music cognition posit that music is a multimodal medium experienced not only through sound, but also through visual and kinematic cues.
These theories have  given rise to a further study of musical performance through the analysis of movement data captured during performance \cite{godoy2010musical}. 
Research outcomes indicate that body motion during musical performances is not incidental, but instead contributes to the meaning of performed music \cite{macritchie2013inferring,Ostersjo2019}. 
Music and consciousness studies have combined perspectives from music psychology, philosophy and neurology, approaching further perspectives on the role of embodiment in musical performance \cite{herbert2019music}. 
Studies in these fields have primarily employed quantitative means such as video, motion capture, and sensor data in order to analyse different aspects of the musical experience \cite{jensenius2013action,visi2017musical}. 
In order to gain a deeper insight, recent multi-method approaches combine quantitative and qualitative data \cite{coorevits2015decomposing,Ostersjo2019}.

Recent work in machine learning for music mainly focuses on the design of gestural interaction with musical sound \cite{Visi2020b}.
Gesture analysis so far, requires additional body sensors (see also in the next section) and/or human labeling.

Although quantitative analysis of video data of music performance alone appears to be a fruitful field for further development, it remains an under-researched area, which could benefit from the employment of recent deep learning techniques. 

\section{Datasets}  \label{dataset}

%\todo {Federico, Stefan, Killian}

\subsection{Source of the dataset}
The dataset used in this paper consists of four annotated videos that were recorded and analyzed by experts in a previous music performance study \cite{coorevits2015decomposing}. The definition and annotation of the expressive gestures in the performance is the result of joint analysis sessions.

% \usepackage{graphicx}
\begin{table}[!ht]
    \small
    \footnotesize
    \caption{Technical characteristics of the videos}
\centering

\label{tab:dataset}
    \begin{adjustbox}{width=\textwidth}
%\resizebox{\textwidth}{!}{%
\begin{tabular}{rcccccc}
\hline
\textbf{\begin{tabular}[c]{@{}c@{}}Original\\ name \end{tabular}} &
  \textbf{Name} &
  \textbf{\begin{tabular}[c]{@{}c@{}} Duration\\ (sec)\end{tabular}} &
  \textbf{\begin{tabular}[c]{@{}c@{}}Frame-rate\\  (fps)\end{tabular}} &
  \textbf{\begin{tabular}[c]{@{}c@{}} No. of \\ frames\end{tabular}} &
  \textbf{Dimension} &
  \textbf{\begin{tabular}[c]{@{}c@{}}Sampling rate\\ (KHz)\end{tabular}} \\ \hline
  \hline
Three times concert   & video-1 & 480.55 & 25 & 12013 & 1080 x 1920 & 48 \\
Three times rehearsal & video-2 & 434.28 & 25 & 10587 & 1080 x 1920 & 48 \\
Four times concert    & video-3 & 662.29 & 25 & 16557 & 1080 x 1920 & 48 \\
Four times rehearsal  & video-4 & 628.93 & 25 & 15723 & 1080 x 1920 & 48 \\
\hline
\end{tabular}%
\end{adjustbox}
%}

\end{table}    % ref the table as  \ref{tab:dataset}
The four videos show the music performer playing Austerity Measures II, a piece for 10 string guitar by the composer David Gorton. 
The composition is structured in five distinct sections, across a total of 64 bars, and the preface to the piece describes how the performer should repeat this sequence, either three or four times, while progressively dropping more and more bars with each new iteration. 
This process presents the performer with the challenge of adapting the shaping of phrases to the gradual loss of musical material. 
The performer is the one to decide which bars to drop, according to a set of rules decided by the composer. 

There are four performances of the piece in the dataset, comprised by one concert performance and one rehearsal recording respectively, of the three and four times versions of the piece~(see Table \ref{tab:dataset}).

\cite{coorevits2015decomposing} used the design of the composition to study the relationship between expressive gestures and the phrasing in each performance. 
They designed an analysis method that could be used to objectively determine the phrasing and the expressive gestures that were used by the performer.

The analysis of the expressive gestures consisted of four music experts negotiating and agreeing upon which gestures of the musician during his representation were expressive gestures, through qualitative analysis of the audio and video. 
%(musician - Stefan Östersjö -, composer - David Gorton - and two musicologists - Esther Coorevits and Dirk Moelants) 
Only unanimously agreed upon expressive gestures were then annotated onto the videos using the ELAN \footnote{https://archive.mpi.nl/tla/elan} software and exported as tab-delimited text files.
Using a Python script, we then segmented time into 0.04 second time steps (the sample rate of the video is 25Hz) and assigned a boolean value for each gesture at each time step of the video indicating whether or not the gesture was being performed at that moment in time. This resulted in a dataset where for each video frame we have a line in our database that accounts for the presence or absence of the gesture during the corresponding time step. 
As the labels and video frames duration do not overlap perfectly, the gesture will be tagged as present in the frame if there is any overlap between the label and the duration of said frame.

\subsection{Data description}
The data used in this paper consists of four music performance videos and their corresponding gesture labels.
The technical description of the videos is included in Table \ref{tab:dataset}.
The representations were filmed in an environment with a glass background and the performer sitting down, facing the camera. 
There were also several sensors attached to the performer but their data was not used for the purpose of this study.

\begin{table}[!t]
    \small
    \footnotesize
\centering

\caption{Number of frames where each of the classes are present per video}
\label{tab:18classdistribution}
   % \begin{adjustbox}{width=\textwidth}
\begin{tabular}{lrrrr}
\hline
\textbf{Class}               & \textbf{video-1} & \textbf{video-2} & \textbf{video-3} & \textbf{video-4} \\ \hline
\hline
Facial expression            & 3369             & 1024              & 3098              & 1931              \\
Nodding                      & 1377             & 836               & 962               & 1269              \\
Right hand round             & 1159             & 547               & 1397              & 1135              \\
Expressive shoulder movement & 2029             & 1945              & 2065              & 3979              \\
Left hand gesture            & 1234             & 370               & 919               & 1252              \\
Lifting head                 & 1493             & 1454              & 2058              & 1735              \\
Minimal movement             & 0                & 4016              & 4085              & 4313              \\
Eyes closed                  & 1271             & 219               & 1248              & 1151              \\
Vibrato                      & 469              & 432               & 674               & 533               \\
Expressive preparation       & 907              & 0                 & 1225              & 602               \\
Freeze                       & 858              & 559               & 1826              & 1580              \\
Expressive head movement     & 261              & 319               & 325               & 557               \\
Frowning                     & 4481             & 1190              & 4215              & 3701              \\
Physical energy              & 838              & 144               & 0                 & 120               \\
Upbeat in head movement      & 0                & 66                & 142               & 90                \\
Repositioning guitar         & 0                & 191               & 641               & 1411              \\
Sympathetic body movement    & 0                & 0                 & 0                 & 0                 \\
Normal play                  & 4292             & 2156              & 4057              & 2732              \\
\hline
\end{tabular}
%\end{adjustbox}
\end{table}   % ref it as \ref{tab:18classdistribution}
\begin{table}[!t]
    \small
    \footnotesize
\centering

\caption{Number of occurrences for each gesture for each video - 18 classes}
\label{tab:gesture_occurrences_18}
   % \begin{adjustbox}{width=\textwidth}
\begin{tabular}{lrrrr}
\hline
\textbf{Class}               & \textbf{video-1} & \textbf{video-2} & \textbf{video-3} & \textbf{video-4} \\ \hline
\hline
Facial expression            & 42              & 14               & 17              & 13              \\
Nodding                      & 59               & 28               & 46               & 55               \\
Right hand round             & 18               & 11               & 20               & 20               \\
Expressive shoulder movement & 24              & 19              & 20              & 36              \\
Left hand gesture            & 21               & 8               & 16               & 16               \\
Lifting head                 & 22               & 20               & 22              & 24              \\
Minimal movement             & 0                & 18              & 10              & 11              \\
Eyes closed                  & 22               & 4               & 12               & 17               \\
Vibrato                      & 17               & 15               & 23               & 20               \\
Expressive preparation       & 17               & 0                & 17               & 10               \\
Freeze                       & 13               & 12               & 23              & 22              \\
Expressive head movement     & 8               & 5               & 7               & 7               \\
Frowning                     & 62              & 20               & 66              & 55              \\
Physical energy              & 23               & 4                & 0                & 3                \\
Upbeat in head movement      & 0                & 2                & 6                & 4               \\
Repositioning guitar         & 0                & 3               & 12               & 25               \\
Sympathetic body movement    & 0                & 0                & 0                & 0                \\
Normal play                  & -              & -              & -              & -             \\
\hline
\end{tabular}
%\end{adjustbox}
\end{table}

It is difficult to estimate what could be considered enough data for each gesture to be predicted with satisfactory performance. 
We have included tables \ref{tab:18classdistribution} and \ref{tab:gesture_occurrences_18} to give an indication of the quantity of data we have for each class compared to the others, but there are many elements to take into consideration to determine whether there is enough data. 
Therefore those tables are not intended to be taken at face value when trying to determine whether a gesture is represented sufficiently well or not.

Table \ref{tab:18classdistribution} details the number of frames in which each of the 18 gestures appears per video, while Table \ref{tab:gesture_occurrences_18} shows the number of occurrences for each gesture.

%This table \ref{tab:gesture_occurrences_18} when comparing it to table \ref{tab:18classdistribution} allows estimating how many frames on average the gestures are represented on (and so how long it appears on screen).
%This could help make sure that each gesture is represented for a duration suitable for detection, and that, having ample different repetitions of the same movement, it would make the dataset more complete.

%Indeed, the presence of each gesture will be evaluated on a sequence of 16 frames (see section~\ref{temp_smooth}, total duration of 0.64 seconds) so there might be differences between how well gestures are represented as they inherently each have different a duration and some might be harder to detect with only a shorter sample evaluated (starting from which value of X frames showing the gesture among the window of sixteen, does the prediction performance start to fall ?).
%Also by having many repetitions of the same gesture (in different conditions as other gestures might be done concurrently), the dataset can capture slight variations of the same gesture which can help make the prediction system be better at detection and be more robust (i.e. it would be better to have nodding when the head is slightly tilted to the right, and when the head is level, rather than just when the head is level).

The proportion of each gesture is low compared to the total number of frames (and the number of frames of normal play), however, we make the hypothesis that there should be enough representation of most classes to enable a predictor of reasonable performance to be trained for these expressive gestures.

\begin{figure}[!ht]
\centering
\includegraphics[width=0.6\linewidth, height=0.6\linewidth]{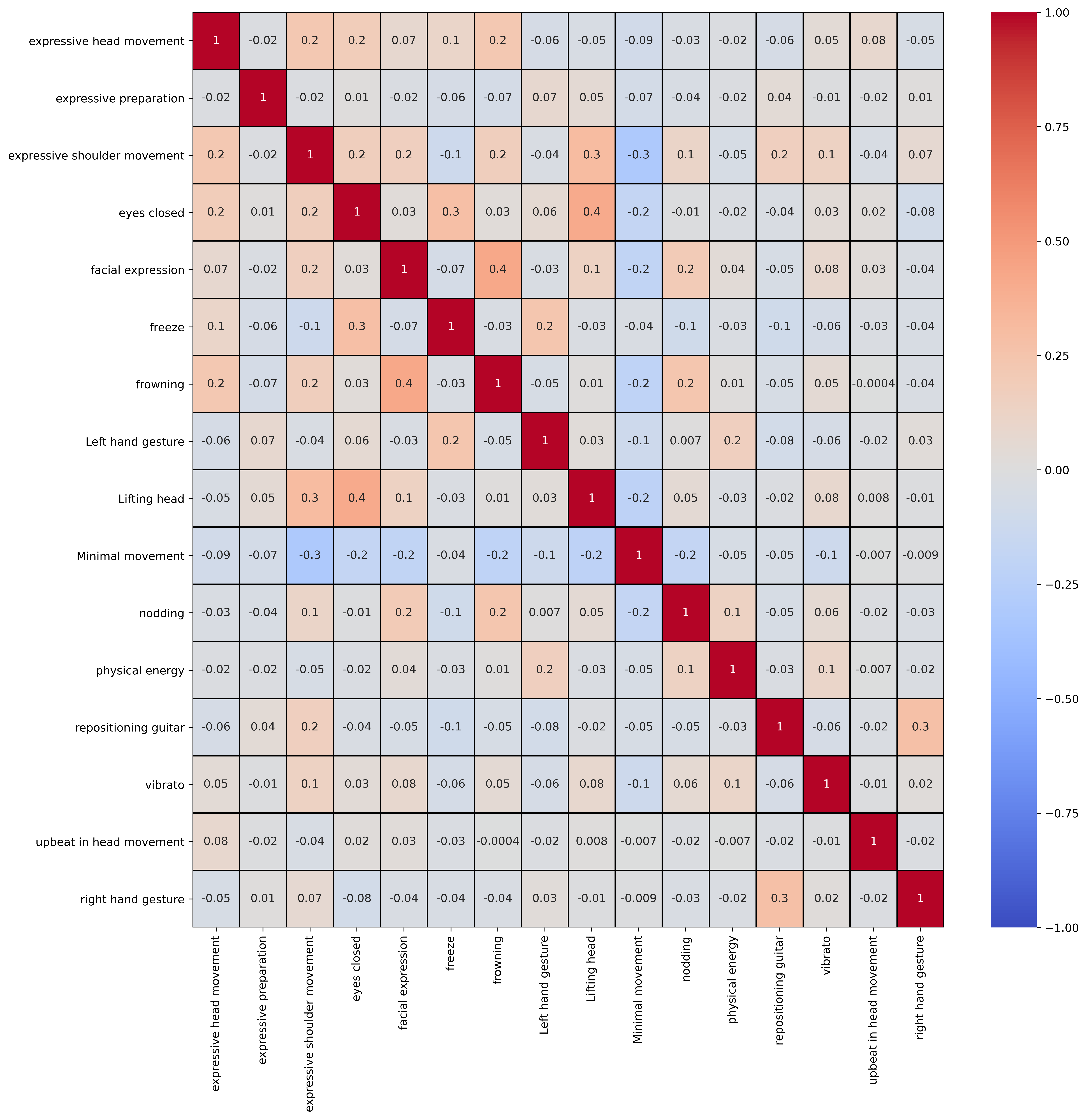}
\caption{Inter-correlation of expressive gestures for video-4}
\label{fig:4r_intercorr}
\end{figure}

The inter-correlation matrices of the expressive gestures are also included in Figure \ref{fig:3c_intercorr} (the inter-correlations of the other videos appear in the appendix Figures~\ref{fig:3r_intercorr},~\ref{fig:4c_intercorr}, and~\ref{fig:4r_intercorr}). The correlation between two gestures X and Y indicates whether tend to be present in the same frames or not. The value 1 means they always appear together when one of them is present and -1 means the two gestures will always have opposing tags.
We can see the relationships between each gesture and that potentially there is additional information brought by each gesture.

\begin{figure}[!t]
\begin{minipage}[b]{0.58\linewidth}
  \centering
  \includegraphics[scale = 0.19]{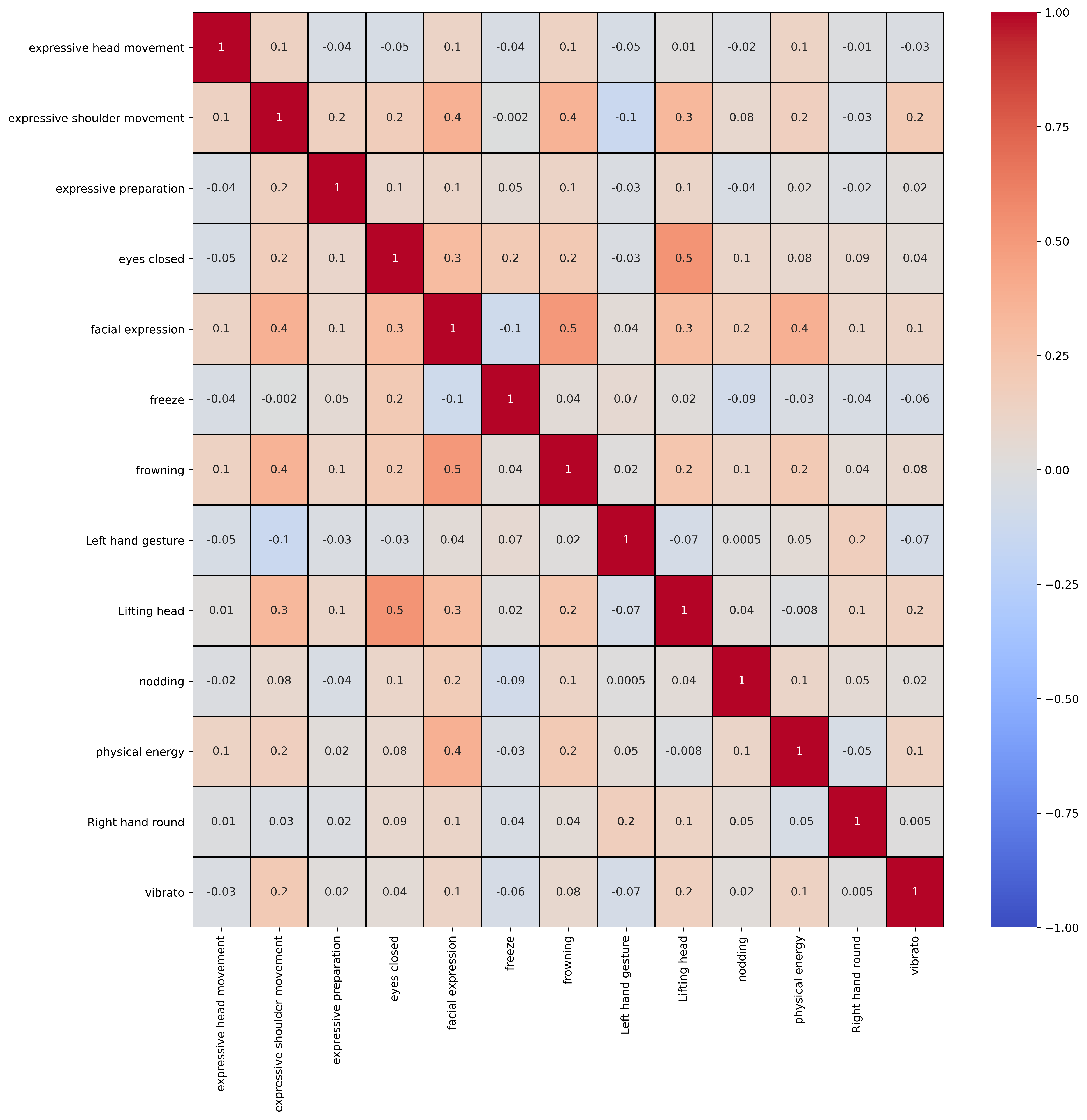}
  \caption{Inter-correlation of expressive gestures for video 1}
  \label{fig:3c_intercorr}
\end{minipage}
\hfill
\begin{minipage}[b]{0.41\linewidth}
  \centering
  \includegraphics[scale = 0.31]{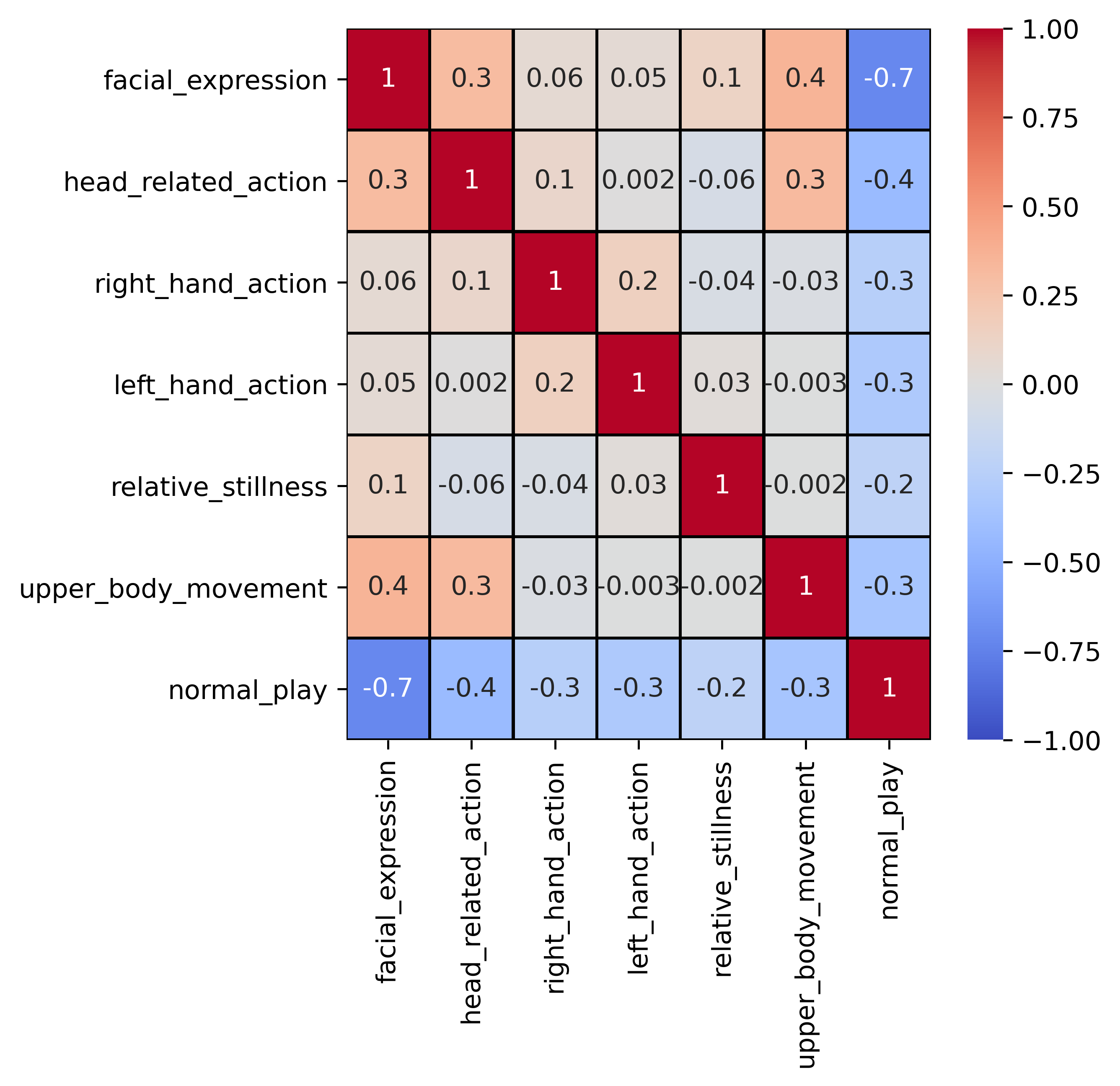}
\caption{Inter-correlation of super-classes for video-1}
\label{fig:3c_intercorr_superclass}
\end{minipage}
\end{figure}

Some points noted from analyzing these figures and table follow. 
The ``eyes closed", ``freeze", and ``lifting head" gestures seem to be moderately correlated together and relatively often all three appear at the same time. 
The ``frowning" and ``facial expression" gestures are also moderately correlated. 
Furthermore, ''minimal movement" is anti-correlated with most of the gestures, which is expected.
We can also note that video-1 has fewer classes, and overall most correlations are positive.
The above remark makes video-1 an outlier compared to the rest.

We have also mapped the expressive gestures to broader super-classes, regrouping several similar expressive gestures (often involving the same body parts), in an attempt to improve the performance of the prediction models. 
The mapping for these super-classes is in Table \ref{tab:mapping}, and the distribution of the super-classes is available in Table \ref{tab:7classdistribution}.

% \usepackage{graphicx}
\begin{table}[!t]
    \small
    \footnotesize
\centering
\caption{Mapping of 18 classes to super classes}
\label{tab:mapping}

\begin{tabular}{ll}
\hline
\textbf{Classes (as mentioned in Table \ref{tab:18classdistribution})}                                     & \textbf{Super classes} \\ \hline
\hline
Facial expression,  Eyes closed,  frowning              & Facial Expression      \\
\begin{tabular}[c]{@{}l@{}}Nodding, Lifting head, Expressive head movement,\\ Upbeat in head movement\end{tabular} & Head related action \\
Right hand round, Repositioning guitar,                 & Right Hand action      \\
Left hand gesture, Vibrato                              & Left hand action       \\
Minimal movement, Freeze                                & Relative stillness     \\
Expressive shoulder movement, Sympathetic body movement & Upper body movement    \\
Rest (any other class), Normal Play                     & Normal play         \\
\hline
\end{tabular}%

\end{table}
\begin{table}[!t]
    \small
    \footnotesize
\centering
\caption{Super class distribution (the numbers in the table indicate the number of data points per super class)}
\label{tab:7classdistribution}
\begin{tabular}{lrrrr}
\hline
\textbf{Super class} & \textbf{video-1} & \textbf{video-2} & \textbf{video-3} & \textbf{video-4} \\ \hline
\hline
Facial expression   & 348 & 121 & 356 & 314 \\
Head related action & 175 & 71  & 205 & 216 \\
Right hand action   & 73  & 40  & 119 & 89  \\
Left hand action    & 100 & 50  & 102 & 112 \\
Relative stillness  & 53  & 267 & 341 & 349 \\
Upper body movement & 127 & 124 & 129 & 249 \\
Normal play         & 280 & 165 & 271 & 184 \\
\hline
\end{tabular}
\end{table}   % ref it as \ref{tab:7classdistribution}

\begin{figure}
\begin{minipage}{\linewidth}
  \centering
  \includegraphics[scale = 0.41]{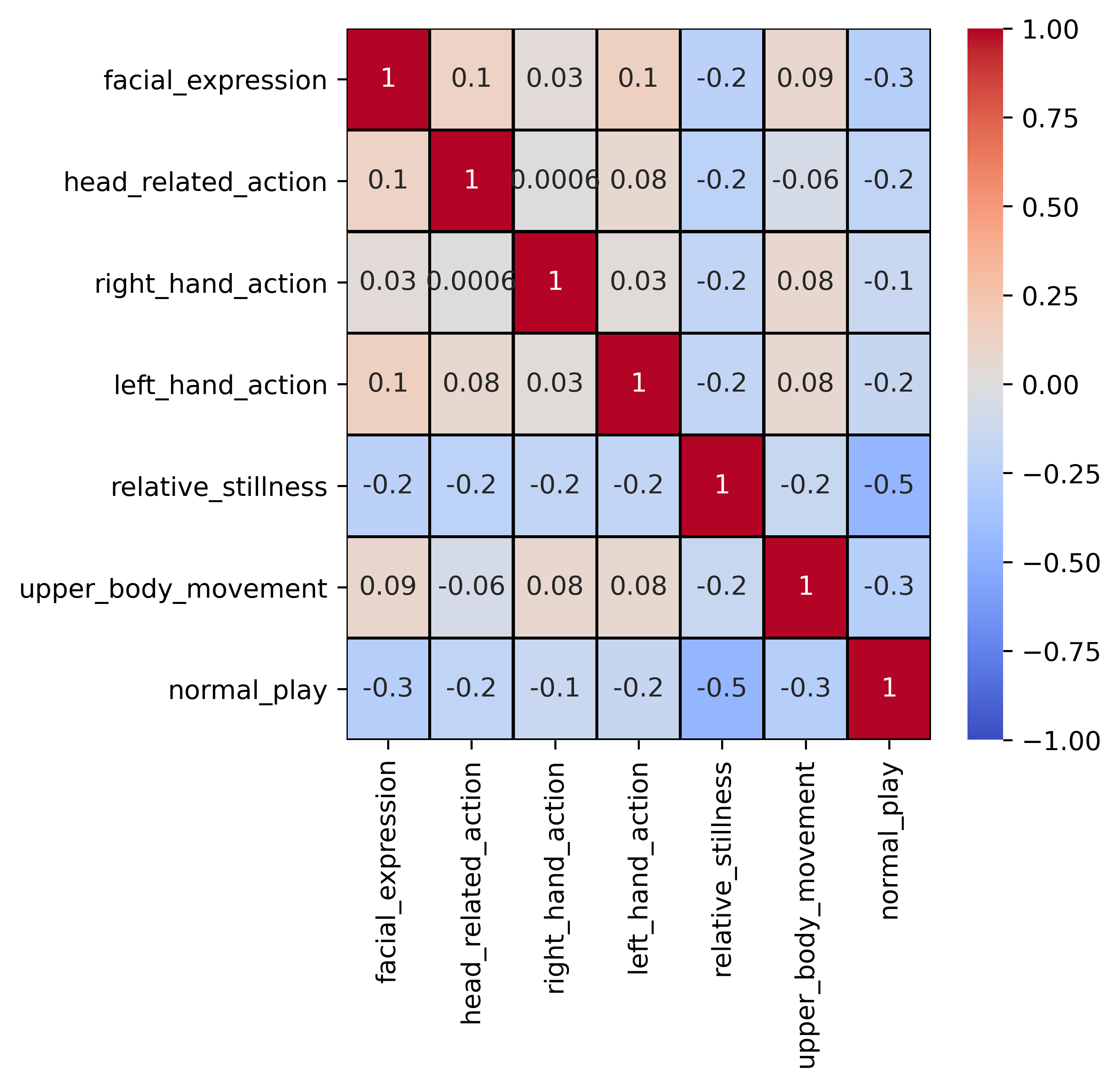}
  \caption{\centering{Inter-correlation of super-classes for video-2}}
  \label{fig:3r_intercorr_superclass}
\end{minipage}
\vspace{1em} % add some whitespace after the first figure

\begin{minipage}[b]{0.45\linewidth}
  \centering
  \includegraphics[scale = 0.23]{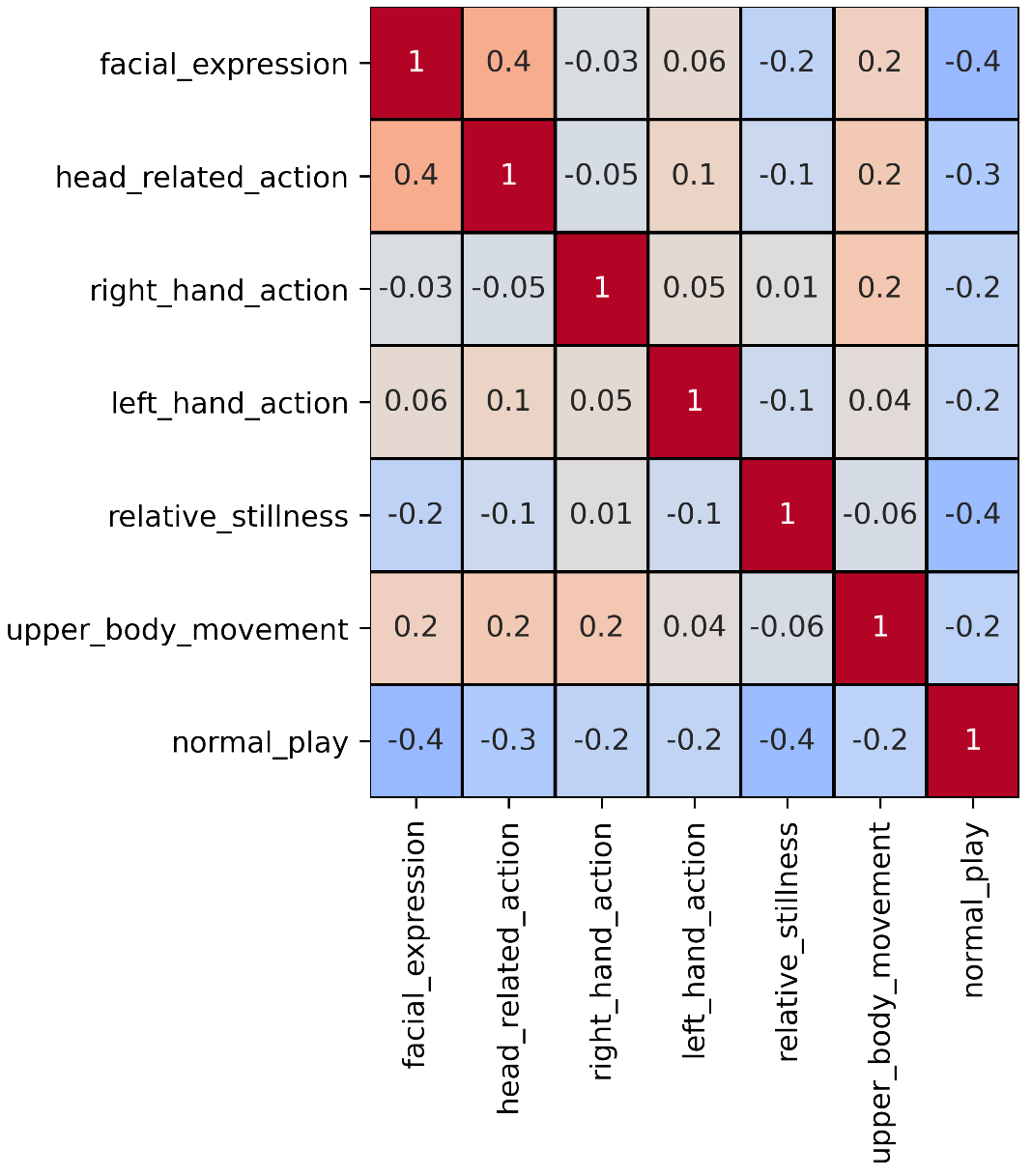}
    \caption{\centering{Inter-correlation of super-classes for video-3}}
  \label{fig:4c_intercorr_superclass}
\end{minipage}
\hfill
\begin{minipage}[b]{0.45\linewidth}
  \centering
  \includegraphics[scale = 0.41]{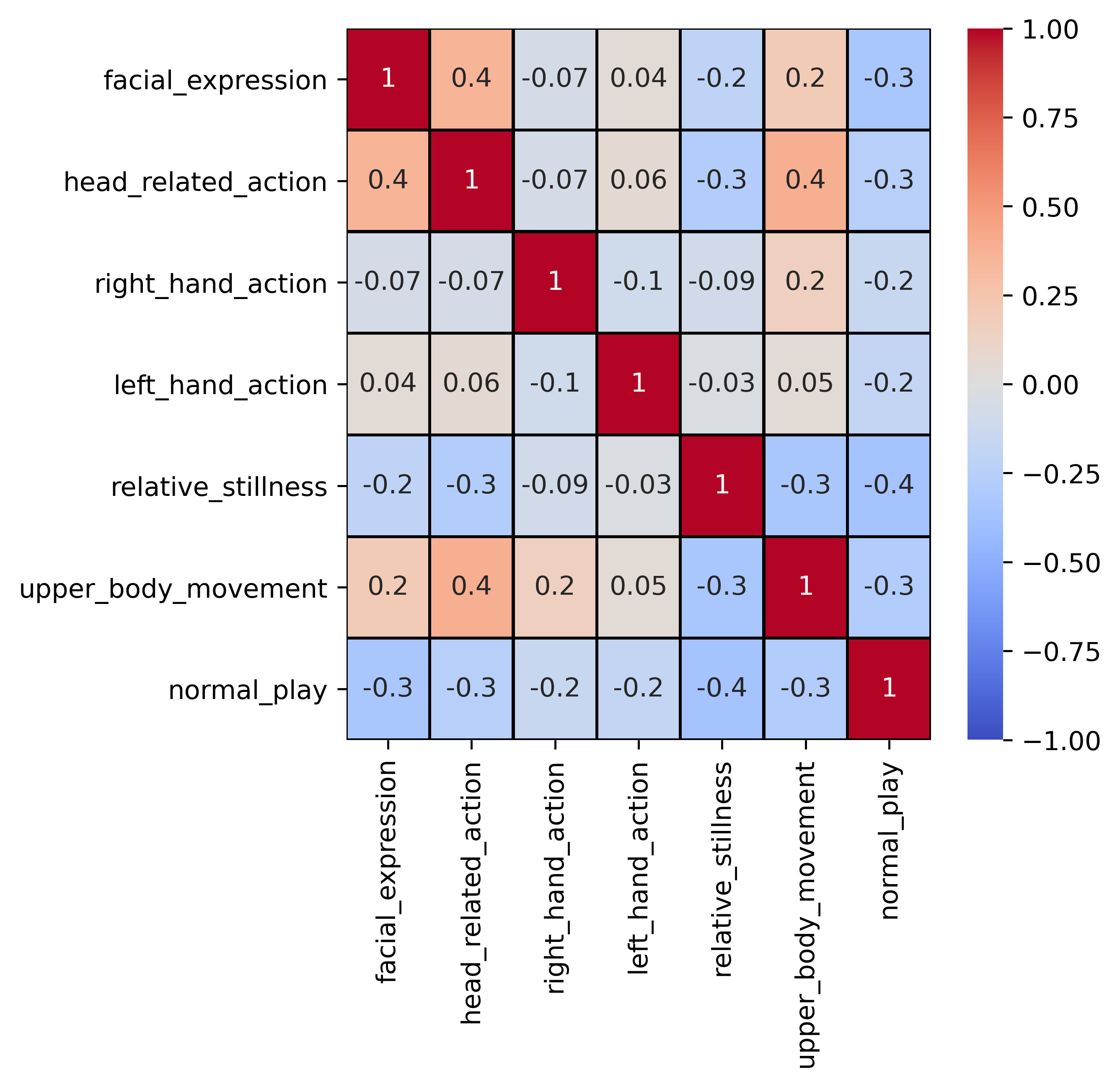}
  \caption{\centering{Inter-correlation of super-classes for video-4}}
  \label{fig:4r_intercorr_superclass}
\end{minipage}
\end{figure}

The inter-correlation matrices of the super-classes are also included in Figure \ref{fig:3c_intercorr_superclass}.
More details can be found in Figure \ref{fig:3r_intercorr_superclass}, Figure \ref{fig:4c_intercorr_superclass}, and Figure \ref{fig:4r_intercorr_superclass} in the Appendix.
It seems that both ''normal play" and ''relative stillness" are moderately anti-correlated with the rest of the classes, which is expected considering the other classes represent an exceptional movement. 
''Head related action" is moderately correlated with ''facial expression" and ''upper body movement", which seems natural given that head movement often occurs with both a change of facial expression and movement of the upper body.
We can also note that both hand movement-related classes seem to be mostly non-correlated with the other classes.

%Instrument: 12-String Guitar\\
%Musician: Stefan Östersjö\\
%Annotators: Esther Coorevits, David Gorton, Dirk Moelants, Stefan Östersjö\\
%Annotating software: ELAN

%[width=100mm,scale=0.5]
% [width=.7\linewidth, height=.7\linewidth]
%[width=0.25\textwidth]

\section{Methodology}  \label{methodology}
\subsection{Single Model} In~\cite{anonym}, a multiple model approach is presented in which each category/class of musical gesture
mapped with one 3D \ac{resnet} \ac{cnn} architecture. 
The outcome of each 3D \ac{resnet} architecture is an independent 2D vector. 
The first element represents the likelihood that the input belongs to the category and the second element represents the absence of the category. 
The current work focuses on simplifying both aspects, multiple models and dual outcomes, by adapting single 3D \ac{resnet} architecture with a single output vector representing the likelihood that input belongs to the category. 
This approach intends to obtain higher performance generalization on unseen examples with reduced computation for model training.

\subsection{Dynamic Batch Balance} \label{sec:BB}
The class distribution, mentioned in Table \ref{tab:18classdistribution} (see previous section) for detailed categories and super classes Table \ref{tab:7classdistribution} for composite categories, are highly imbalanced. 
The frequency of the musical gestures and their co-occurrence is not balanced, so just increasing the dataset's size does not help overcome the class imbalance. 
To overcome the majority classes' domination during training, this work emphasizes balancing the class distribution in batches during training through loss customization. 
Based on previous work in \cite{qiu2018nonuniform} about the non-uniform weighted loss on imbalanced images and \cite{dong2018imbalanced} about minority classes, we design loss customization dynamically based on the class distribution for each batch.

We represent $F_{b}^c$ as the loss factor, $S_b$ as the size of batch, $P_b^c$ as the set of positive examples, $N_b^c$ as the set of negative examples of musical gesture class $c$ in batch $b$ during training. 
$count(P_b^c)$ and $count(N_b^c)$ are the numbers of positive and negative examples respectively. 
Equation \ref{eq:loss_factor} defines the loss factor for batch $b$ for class $c$ corresponding to respective category of musical gesture as following:

\begin{equation} \label{eq:loss_factor}
 F_b^c =
\begin{cases}
 0 & \text{if} \: \text{$count(P_b^c) = 0$} \\
 1 &  \text{if} \: \text{$count(P_b^c) > 0$ \:}  \text{and\: $count(P_b^c) \geq  count(N_b^c$)} \\
 \text{$S_b$/$count(P_b^c)$} & \text{if} \: \text{$count(P_b^c) > 0$ \:} \text{and \: count($P_b^c) < count(N_b^c)$} \\
\end{cases}
\end{equation}

Further, we define the conditional loss quantity $Loss_b^c$ for class $c$ corresponding to respective category of musical gesture for batch $b$ in equation \ref{eq:conditional_loss} in which $criterion$ defines the loss method used.
$Pred(P_b^c)$ and $Target(P_b^c)$ are model prediction and ground truth on set of positive example(s) of class $c$ in batch $b$.
Similarly $Pred(N_b^c)$ and $Target(N_b^c)$ are model prediction and ground truth on set of negative example(s) of class $c$ in batch $b$.

\begin{equation} \label{eq:conditional_loss}
 Loss_b^c =
\begin{cases}
 0 & \text{if} \: \text{$F_b^c = 0$} \\
 \text{$criterion(Pred(P_b^c), Target(P_b^c))  +$ }  & \forall \: \text{$F_b^c \geq 1$} \\
 \text{$criterion(Pred(N_b^c), Target(N_b^c))$} &    \\
\end{cases}
\end{equation}

Each category of musical gesture is represented as a single 1D vector independently because each appearance of every musical gesture is independent of others, unlike multi-class classification. 
Thus we calculate the effective loss with respect to each class $c$ and accumulate it to obtain total loss for batch $b$, $BatchLoss_b$ to backpropagate to 3D \ac{resnet} shown in equation \ref{eq:batch_loss}.

\begin{equation} \label{eq:batch_loss}
 BatchLoss_b = \sum_{i=0}^c  F_b^i \times Loss_b^i
\end{equation}

The above-explained method for calculating batch-wise loss $BatchLoss_b$, is \ac{dbb}. 
This method adapts the imbalance of non-uniform class distribution for every batch, prevents the skewed over-training on majority classes, and represents the loss for each independent class belonging to categories of musical gesture.

Further, the \ac{dbb} is extendable to incorporate complete training dataset level class imbalance by redefining conditional loss quantity $Loss_b^c$ from equation \ref{eq:conditional_loss} with class weight $W^c$ in equation \ref{eq:conditional_loss_ext}.

\begin{equation} \label{eq:conditional_loss_ext}
 Loss_b^c =
\begin{cases}
 0 & \forall \: \text{$F_b^c = 0$} \\
 \text{$W^c \times criterion(Pred(P_b^c), Target(P_b^c))  +$ }  & \forall \: \text{$F_b^c \geq 1$} \\
 \text{$criterion(Pred(N_b^c), Target(N_b^c))$} &    \\
\end{cases}
\end{equation}

\begin{equation} \label{eq:class_level_weights}
 W^c = \frac{\sum_{i=0}^c count(P^i)}{count(P^c)}
\end{equation}

The equation \ref{eq:class_level_weights} defines the estimation of class weight $W^c$ for class $c$ where $P^c$ is set of positive examples of class $c$ and $count(P^c)$ is total example count of class $c$. 

\ac{dbb} serves as a framework for countering class imbalance in the dataset at batch and class levels. 
\ac{dbb} counters the class imbalance with loss customization during training where classes are coexistent, which makes oversampling methods for minority classes not so well defined theoretically and empirically. 
Further, \ac{dbb} is able to balance classes where data augmentation methods are limited by the nature of data and the coexistence of more than one class. \ac{dbb} is based on loss customization to improve the learning of the model during training, so it is not used during evaluation on validation and test set.

\subsection{Temporal Smoothing} \label{temp_smooth}
%Weighted moving average define the temporal and quantitative presence of gesture in precise manner

Label Smoothing is a regularization technique for deep neural networks that introduces noise for the labels, thus increases robustness. 
It generates labels in terms of distribution by applying weighted average across classes. 
According to \cite{DBLP}, noisy labels with bootstrapping makes the model robust towards label corruption while successfully extending the approach for deep neural networks. \cite{goibert2019adversarial} defines label smoothing as a method for improving adversarial robustness of supervised deep-learning models, which boosts the adversarial performance of the model along with generalization.
Specifically, \cite{NEURIPS2019_f1748d6b} investigates the role of label smoothing in the case of knowledge distillation in a teacher-student network. 
However, most of the existing approaches of label smoothing are majorly intended to introduce noise systemically in labels to attain a robust model against possible data corruption in the real-world and prevent over-fitting. 

In addition to the standard nature of labels for image classification, label information for temporal data such as video frames and audio frames contain temporal characteristics. 
Such temporal characteristics are based on the start and end time step of a label corresponding to a specific class.
Further, the presence of such labels is likely to coexist and create overlaps for a given time step. 
In our work, one example of a musical performance dataset is represented as a series of 16 video and (combined) audio frames. 
Thus it becomes crucial to be able to represent the label(s) based on their temporal characteristics.  

We define a perspective of label smoothing for more accurate representations of coexistent labels for temporal data based on the above defined temporal characteristics. 
Unlike the purpose of noise inclusion, our method of Temporal Smoothing, focuses on representing smoother labels for video frames data on the temporal dimension, which allows the label representation closer to ground truth. 
We formulate label smoothing for each example (16 consecutive frames) in the temporal dimension using \cite{perry2010weighted} \ac{wma} method. 
\ac{wma} models the actual presence of musical gestures in temporal ordered video frames and remains flexible for the coexistence of multiple musical gestures in possible overlaps. 
Furthermore, the current method assigns a greater weighting to the label presence in most recent video frames, and less weighting to the label presence in older video frames in the temporal order of 1 to 16 frames. 

Equation \ref{eq:temporal_smoothing} shows the temporal smoothing label $Label^c_\text{TS}$ for class $c$ of an example consisting of $n$ frames in temporal order where frame at $i=n$ is most recent. 
$Label_i^c$ represents the label of class $c$ at $i^th$ time step with possible value of $1$ if musical gesture corresponding to class $c$ is present, otherwise $0$.  

\begin{equation} \label{eq:temporal_smoothing}
 Label^c_\text{TS} = \frac{\sum_{i=n}^1 i \times Label_i^c}{\sum_{j=n}^1 j}
\end{equation}

\subsection{Multimodal Representation} \label{multimodal_representation}

 Representation of audio and video data in a structure that a computational model can understand and learn is a challenge.
 In this work, in order to exploit the audio information along with the video, for gesture classification, a joint representation of the video and audio features is used. 
 In such a representation, for the transfer of knowledge between the modalities, the unimodal features are combined into the same representation space. 
 Such a concatenation of the individual modalities is also referred to as early fusion \cite{Baltrusaitis2017}. 
 Deep neural networks are extensively used as a well-known as well as an efficient approach for unimodal data representations \cite{bengio2014representation}. 
 The subsequent layers of a deep neural network give an abstract representation of the data; that is why very often, the last layers of a deep neural network are used to illustrate high-level features of the input. 
 This work exploits the given fact and builds a multimodal representation of the video frames and audio by providing these modalities through deep neural networks and projecting the extracted latent features into a joint space. 
 These features are further supplied to another deep neural network for the classification of gestures. 
 Section \ref{arch} describes the neural network architectures used for generating the joint representations for both the modalities.

\subsubsection{Data Modalities}
As already discussed in Section \ref{dataset}, in this work, video recordings of a musician playing guitar are used for analyzing the performance by means of movements or gestures of the individual. 
Therefore, there are two modalities present in the data, \ie visual and audio information, which are utilized as follows: 

\begin{itemize}
\item{\textbf{Video Data}} \label{video_rep}
In order to extract higher-order features from the video, a model previously trained on a huge dataset (nearly 650,000 video clips \cite{hara3dcnns}) is used in this work to obtain representations that can further aid in classifying gestures during a musical performance. 
This is a classic example of transfer learning \cite{bengio2014representation} wherein the knowledge gained during training for one task is utilized as an onset for another related task. 
In a case where only the video frames are used for classification, \ie unimodal, pretrained model is used as a feature extractor, and these meta-features are further processed in a deep neural network (usually a fully connected neural network with multiple hidden layers) to classify the gestures. 
While in case another modality is also employed, then the joint representations (as discussed in section \ref{multimodal_representation}) are used as input to the deep neural network for classification.

\item{\textbf{Audio data}}

We used audio recorded by means of a contact microphone placed on the guitar while filming the performance. 
Such audio cannot be directly fed to the deep neural network due to its huge dimension.
Thus, to feed the deep neural network with scalable input, a set of audio features were extracted during data preprocessing. 

These features are then fed to a deep neural network to extract suitable hidden representations, which are then combined with the video representations (discussed in section \ref{video_rep}) and given as an input to a classifier network.   

The following low dimensional audio descriptors, also known as timbral texture features \cite{audiofeatures} are derived from the raw audio signal. 
For all the frequency domain (spectral) features, the \ac{stft} is calculated by dividing it into small overlapping segments and finding the Fourier Transform for each segment. 
\begin{itemize}
\setlength\itemsep{0.5em}
    \item[] \textbf{\ac{mfcc} and derivatives} The audio signals are non-stationary in nature, which means their frequency varies with time. 
    According to \cite{time_freq_analysis}, for such signals, Fourier transform does not gives an informative frequency domain representation. 
    For this reason, \ac{mfcc} are widely used for \ac{mir}. These are static characteristics that contain information only from a frame or segment. 
    The power spectrum of the \ac{stft} is passed through a mel filter banks (triangular filters); these filters have low bandwidth at low frequencies and high bandwidth at higher frequencies which basically reflects the behaviour of human hearing.
    The output frequencies from the filters are log-transformed, and the discrete Fourier transform of the resulting values gives \ac{mfcc}. 
    The shape of the spectral envelope is described by the \ac{mfcc} features, i.e., they capture the changes of the spectral envelope of the audio. 
    The first order and second-order derivatives of the \ac{mfcc} coefficients are also considered as audio descriptors as they capture changes of the cepstrum.
    
    \item[] \textbf{Spectral centroid} These represent the spectral position and shape. 
    The spectrum's center of gravity is represented by the spectral centroid. 
    Let $A_t[n]$ be the magnitude of Fourier transform at segment t and k is the frequency bin, the spectral centroid is expressed as; 
    \begin{equation}
        C_t = \dfrac{\sum_{i=1}^K n A_t[k] }{\sum_{i=1}^K  A_t[k] }
        \label{spec_centroid}
    \end{equation}
    \item[] \textbf{Spectral bandwidth} It is defined as the distribution of the power of the spectrum of the signal around its center frequency. 
    \item[] \textbf{Spectral roll-off} The frequency $F_t$ below which lies 85\% of the magnitude of the signal is spectral roll-off. It can be expressed as;
    \begin{equation}
        \sum_{i=0}^{S_f} A_t[k] = 0.85 \sum_{i=0}^{K} A_t[k]
    \end{equation}
    
    \item[] \textbf{Spectral flatness} This explains the noise versus sinusoidality of the signal. 
    It equals to one if the signal only contains noise and equals to zero if the signal is sinusoidal.
    
    \item[] \textbf{Spectral contrast} It is the difference between the peak and valley of the spectrum. 
    \item[] \textbf{\ac{zcr}} According to \cite{MITROVIC201071}, the \ac{zcr} is the measure of the dominant frequency of the signal, as it accounts for the number of zero crossings per second. 
    It is expressed as;
    \begin{equation}
        R_{zerocrossing} = \dfrac{1}{2} \sum_{n=1}^{N} \mid sign(x[n]) - sign(x[n-1]) \mid
    \end{equation}
    \item[] \textbf{Root-mean-square} This is basically a measure of  normalized signal energy in the time domain; for a discrete signal x(n), it is expressed as;
    \begin{equation}
        x_{rms} = \sqrt{\dfrac{1}{N} \sum_{n=0}^{N-1} x^2(n)}
    \end{equation}
   
\end{itemize}
\end{itemize}
\section{Implementation} \label{implementation}
This work is centered around two types of implementations, the first is unimodal, which is only using the video, and the second is bimodal, in which the audio is also used for training a model along with the video. 
For both cases, the following experiments in Table \ref{tab:exp} were designed for the 18 classes as well as for the 7 super classes.
\begin{table}[!ht]
\centering
\caption{Overview of the conducted experiments, approaches, and data distribution (the numbers in the table indicate the number of data points)}
\label{tab:exp}
\begin{tabular}{|c|c|c|c|c|}
\hline
\textbf{} & \textbf{} & \multicolumn{3}{c|}{\textbf{Data Distribution}} \\ \cline{3-5} 
\textbf{Data} & \textbf{Approaches} & \textbf{\begin{tabular}[c]{@{}c@{}}Train\\ set\end{tabular}} & \textbf{\begin{tabular}[c]{@{}c@{}}Valida-\\ tion set\end{tabular}} & \textbf{\begin{tabular}[c]{@{}c@{}}Test\\ set\end{tabular}} \\ \cline{3-5} 
\textbf{} & \textbf{} & \textbf{80\%} & \textbf{10\%} & \textbf{10\%} \\ \hline
Video-1 & Single Model (SM) &  &  &  \\ \cline{2-2}
 & Single Model (SM) + Batch Balancing (BB) &  &  &  \\ \cline{2-2}
 & \begin{tabular}[c]{@{}c@{}}Single Model (SM) + Batch Balancing (BB)+\\ Temporal Smoothening (TS)\end{tabular} & 600 & 75 & 75 \\ \cline{2-2}
 & \begin{tabular}[c]{@{}c@{}}Bi-modal (video+audio) + Single model (SM)+\\ Batch Balancing (BB) + Temporal Smoothening\\ (TS)\end{tabular} &  &  &  \\ \hline
Video-1 + & Single Model (SM) &  &  &  \\ \cline{2-2}
Video-2 + & Single Model (SM) + Batch Balancing (BB) &  &  &  \\ \cline{2-2}
\multicolumn{1}{|l|}{\begin{tabular}[c]{@{}l@{}}Video-3 +\\ Video-4\end{tabular}} & \begin{tabular}[c]{@{}c@{}}Single Model (SM) + Batch Balancing (BB) + \\ Temporal Smoothening (TS)\end{tabular} & 2758 & 344 & 344 \\ \cline{2-2}
 & \begin{tabular}[c]{@{}c@{}}Bi-modal (video+audio) + Single Model (SM) + \\ Batch Balancing (BB) + Temporal Smoothening\\ (TS)\end{tabular} &  &  &  \\ \hline
Leave &  & \multicolumn{2}{c|}{Video 1, 2, 3} & Video 4 \\ \cline{3-5} 
Video - 4 &  & 2218 & 246 & 928 \\ \cline{1-1} \cline{3-5} 
Leave &  & \multicolumn{2}{c|}{Video 2, 3, 4} & Video 1 \\ \cline{3-5} 
Video - 1 & Single Model (SM) + Batch Balancing (BB) & 2425 & 269 & 750 \\ \cline{1-1} \cline{3-5} 
Leave & + Temporal Smoothening (TS) & \multicolumn{2}{c|}{Video 1, 3, 4} & Video 2 \\ \cline{3-5} 
Video - 2 & (only for 7 super classes) & 2491 & 276 & 678 \\ \cline{1-1} \cline{3-5} 
Leave &  & \multicolumn{2}{c|}{Video 1, 2, 4} & Video 3 \\ \cline{3-5} 
Video - 3 &  & 2171 & 241 & 1034 \\ \hline
\end{tabular}
\end{table}

\subsection{Preparing data}
Image frames are extracted from all the videos at the given frame rate. 
16 successive frames make one data sample (16 frames = 0.64 seconds of a video clip) and there is no frame overlap while creating the input data.
The reason for taking specifically 16 consecutive frames is because of the input requirements of 3-D \ac{cnn} architecture \cite{hara3dcnns} used in this work.
Every data point (of 16 frames) is assigned labels as per the gestures present in those frames. 
Note that as discussed in Section \ref{temp_smooth}, many gestures can co-exist in one data point. 
These frames are fed to the network in batches of 32 samples for training, validation, and testing. 
Prior to this the images are resized to 112 x 112 \cite{hara3dcnns}. 
Now the dimension of each data sample is 3 channels x 16 frames x 112 pixels x 112 pixels. 
As for the audio data, it is sliced into segments of 0.64 seconds corresponding to the video frames. 
For calculating the low dimensional audio features, the 0.64 audio segments are divided into smaller overlapping segments called windows.
The size of this window should be so small that the spectrum has stable frequency characteristics. 
The dimension of audio features for each 0.64 second segment is 28 x 108, where 28 are the number of windows (window size = 2048) used to calculate the features and 108 constitutes 27 key features (20 \ac{mfcc} + 7 rest) along with the first, second, and third-order derivatives of all the features.

%%%%%%%%%%%%%%%%%%%%%%%%%%%%%%%%%%%%
\begin{figure}[!ht]
  \begin{subfigure}[b]{0.4\textwidth}
    \includegraphics[scale = 0.55]{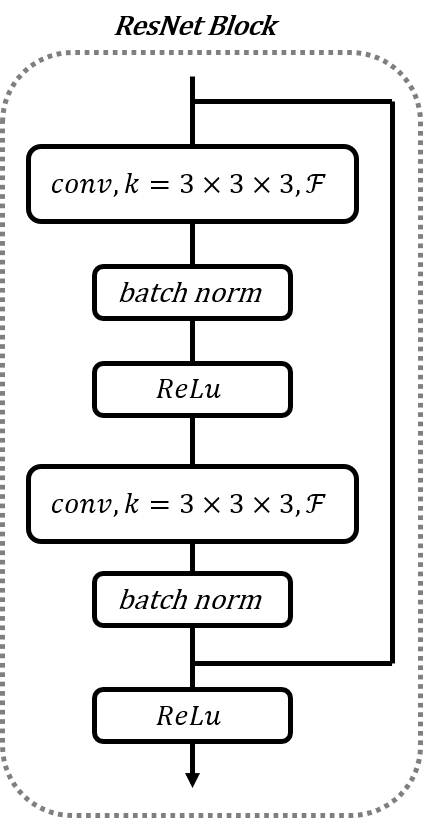}
    \caption{Basic ResNet block}
    \label{fig:resnetblock}
  \end{subfigure}
  \hspace{3em}
  \begin{subfigure}[b]{0.4\textwidth}
    \includegraphics[scale = 0.6]{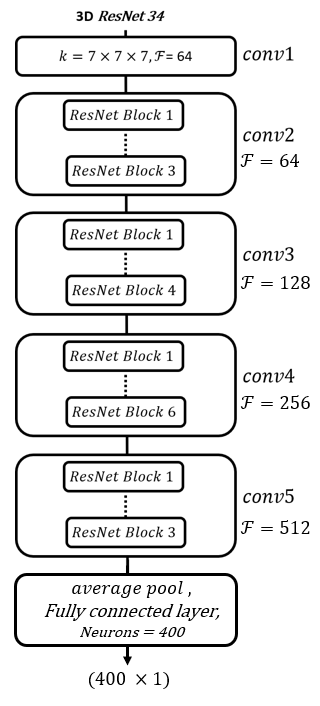}
    \caption{3D ResNet-34 architecture}
    \label{fig:resnet34}
  \end{subfigure}
  \caption{(a) shows the \ac{resnet} block used in this work, (b) shows the 3 dimensional \ac{resnet} 34 architecture, which is composed of a number of \ac{resnet} blocks. Here F is the number of the convolutional filter feature maps.  Spatio-temporal
down-sampling is performed by conv3\_1, conv4\_1, and conv5\_1  \ac{resnet} blocks with a stride of two, while in conv1 the temporal stride is one and spatial stride is two. }
%  \caption{(a) shows the ResNet block used in this work, (b) shows the 3 dimensional ResNet 34 architecture, which is composed of a number of ResNet blocks. Here $\mathcal{F}$ as mentioned in \cite{hara3dcnns} is the number of feature maps of convolutional filter}
\end{figure}
%%%%%%%%%%%%%%%%%%%%%%%%%%%%%%%%%%%%%%%
\subsection{Architectures} \label{arch}
\begin{figure}[!ht]
  \begin{subfigure}[b]{0.4\textwidth}
    \includegraphics[scale = 0.35]{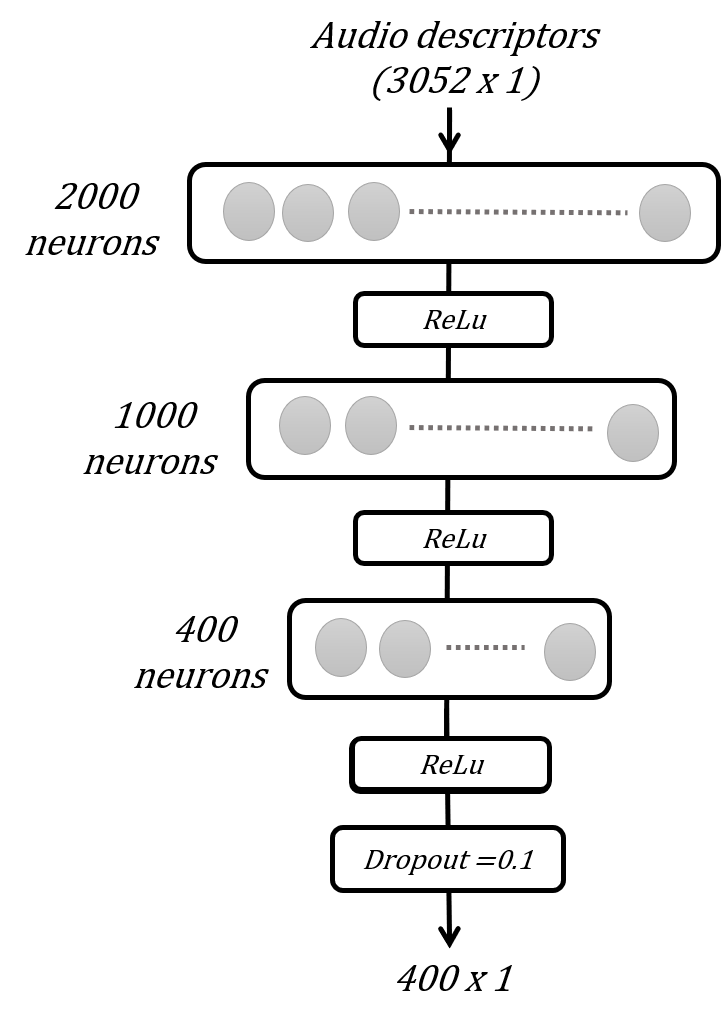}
    \caption{Fully connect deep neural network}
    \label{fig:DNN}
  \end{subfigure}
  \hspace{1em}
  \begin{subfigure}[b]{0.55\textwidth}
    \includegraphics[scale = 0.4]{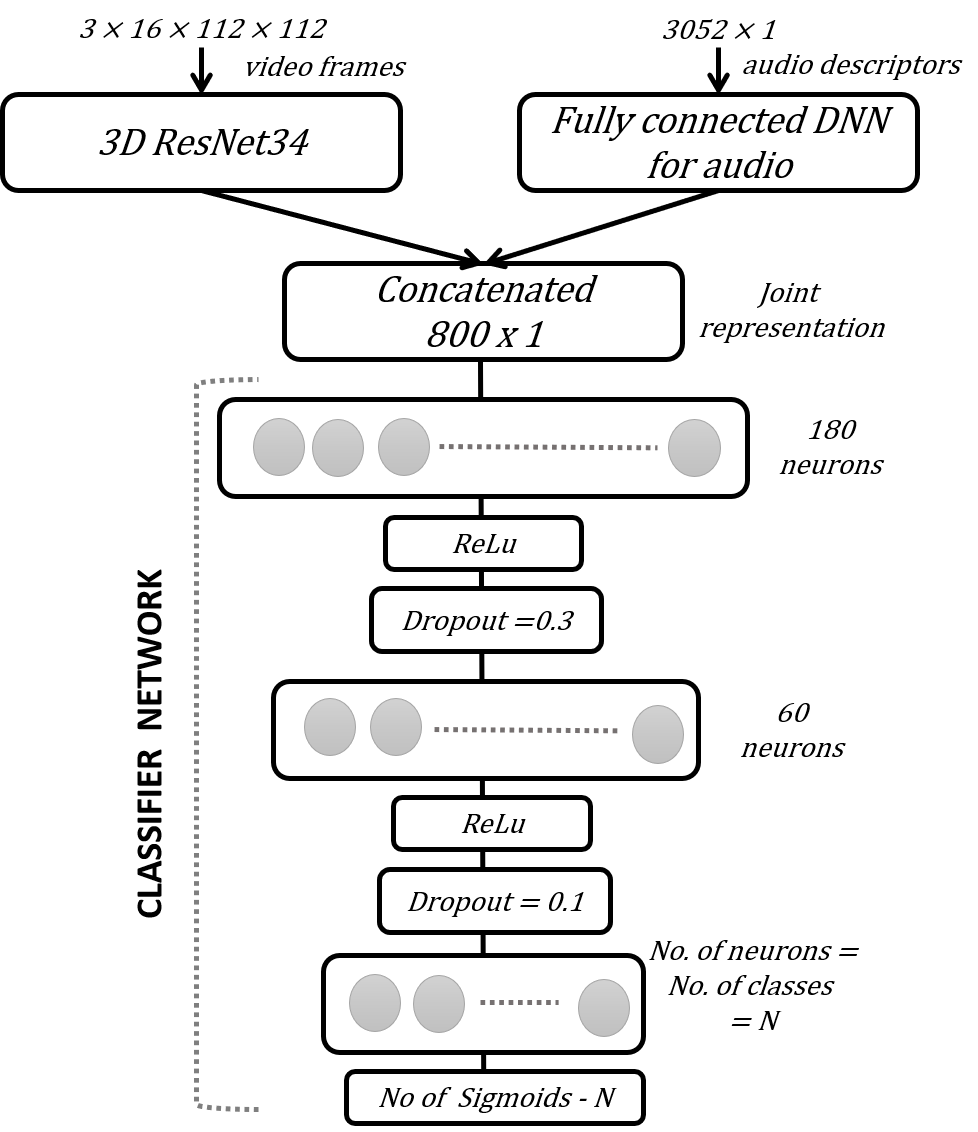}
    \caption{Multimodal architecture}
    \label{fig:multimodal}
  \end{subfigure}
    \caption{(a) represents the architecture of the model used to find the audio representations, (b) shows the multimodal architecture along with the classifier network. Here N is the number of classes/labels in this case the number of gestures to be classified, so it can be either 7 ( Table \ref{tab:7classdistribution}) or 18 (Table \ref{tab:18classdistribution})  }
\end{figure}

For both types of implementations, while working with the video information, a pretrained 3D-ResNet34 model \cite{hara3dcnns}\cite{kataoka2020megascale} is employed.
\ac{resnet}s have proven to be the most efficient architectures for image classification. 
The model uses 34 3D convolutional layers for human activity recognition in videos. 
It is trained on a very large dataset of high-quality videos i.e. Kinetics 400 \cite{carreira2017quo} which classifies 400 different activities in the video clips. 
Since it is a 3D \ac{cnn}, it processes 16 frames as one data point. 
The composition of the model is illustrated in Figure \ref{fig:resnetblock}. 
When a data sample (i.e. 16 consecutive frames of the video) is passed through the 3D \ac{resnet} architecture, 400 features are obtained at the output. 
The output of the 3D \ac{resnet} model is utilized distinctly for unimodal and bimodal implementations;
\begin{itemize}
    \item Unimodal: the 400 features are fed to the classifier network for identifying the gestures.
    \item Bimodal: the 400 features are integrated with audio representations, as discussed in section \ref{multimodal_representation} and thereafter given as input to the classifier. 
\end{itemize}
%%%%%%
For extracting the representations for audio, the low dimensional audio features (28 x 109) are unrolled in a one-dimensional vector (3052 x 1) and given as input to a fully connected deep neural network shown in Figure \ref{fig:DNN}. 
It is a fully connected neural network with three hidden layers along with \ac{relu} as the activation function and also a dropout layer at the end to avoid over-fitting during training. 
The primary idea of this network is to extract hidden representations in the data and scale down the dimension to 400, equal to what is obtained from the 3D \ac{resnet}. 
%%%%%%%%%%%%%%%%%%%%%%%%%%%%%%%%%%%
%%%%%%%%%%%%%%%%%%%%%%%%%%%%%%%%%%%%%

Hidden representations of the audio and video data are extracted using networks illustrated in Figure \ref{fig:DNN} and Figure \ref{fig:resnet34}, thereafter for bimodal implementation, they are together given to the classifier network, as depicted in Figure \ref{fig:multimodal}. 
Whereas for unimodal, video only implementation, the 400 hidden features extracted from 3D \ac{resnet} are directly supplied to the classifier network. 
It is also a fully connected neural network with three hidden layers along with \ac{relu} activation function. 
Dropout layers are also added after the activation layer to regularize the network so as to overcome the problem of over-fitting.  
At the output of the classifier network, there are \enquote{N} neurons, depending on the number of gestures.
These outputs are then fed to \emph{N} sigmoid activation functions, respectively, which limit the resulting values between 0 and 1. 
After applying a suitable threshold, the \emph{N} outputs are binarized, which represents the presence `1' or absence `0' of a gesture. 
Multiple sigmoids will help to identify when more than one gestures occur simultaneously.

\subsection{Network Training}
For training and evaluation of the network, the dataset is divided into three exclusive, non-overlapping sets, these are train set, validation set, and test set. 
The division ratio and the number of samples in each set are included in Table~\ref{tab:exp}, above.
For the video frames, a pretrained 3D ResNet34 model shared by \cite{hara3dcnns} is used to extract the features at the time of training. 
This means the weights (or parameters) of the original model (for all the blocks till \enquote{\textit{conv4}})are retained. 
In order to adapt the 3D ResNet34 for this work, the block-4 i.e. \enquote{\textit{conv5}} in the architecture in Figure \ref{fig:resnet34} and the fully-connected layer after it is kept trainable so that the associated weights of these layers are fine-tuned for this task. 
This also considerably reduces the training time. The validation set is used to evaluate the performance of the network during training. 

The implementation of this work is done using PyTorch \cite{pytorch} python library, and each experiment was performed on the GPU cluster, which uses NVIDIA GTX 1080 TI processors.  
Other specifications for training are as follows;
\begin{itemize}
    \item[.] Batch size = 32
    \item[.]  Learning rate = 0.001
    \item[.]  Number of epochs = 3000
    \item[.]  weight decay (or L2 normalization) = $\mathrm{1e^{-2}}$
    \item[.]  Loss function = Mean Square Error (MSE) loss
    \item[.]  Optimizer = \enquote{Adam} derived from adaptive moment estimation \cite{kingma2017adam}
\end{itemize}

Gesture categories in frames are overlapping, making binary cross-entropy loss unsuitable, and further, temporal smoothing transforms labels from discrete value to continuous, in the range of 0 to 1. With this specification, MSE loss is used for each binary classifier.

\subsection{Evaluation Metrics}
To quantify the performance of the trained model on the test set, evaluation metrics are employed. 
As discussed in Section \ref{sec:BB}, the dataset used in this work is imbalanced and sparsed for the gestures, so the most commonly used metrics like accuracy can give misleading conclusions because it is insensitive to skewed data. 
For this reason, in this work \textit{Precision, Recall, and F-1 score} are used as the evaluation metrics. More specifically, this work focuses on recognizing the gestures in video dataset of musical performance so true positive (TP) defines the presence of corresponding gesture in frame. Absence of gesture is not considered to evaluate the performance to concretely define the evaluation.
However, the 'normal play' category is designed for the frames not having any gestures. 
Every output neuron has a sigmoid activation function, so it gives values between 0 and 1, \ie binary classification for every gesture. 
So every gesture/class Precision and Recall are calculated; using these class-wise values average F-1 score for the model is determined. 
Consider that 1 denotes presence (positive class) and 0 denotes absence (negative class) of a gesture; after doing a binary classification, the following values can be defined \cite{metrics}:
\begin{itemize}
\item\ac{tp} - instances when model correctly predicts the positive class (presence of corresponding gesture).
\item\ac{tn} - instances when model correctly predicts the negative class (absence of corresponding gesture).
\item\ac{fp} - instances when model falsely predicts the positive class.
\item\ac{fn} - instances when model falsely predicts the negative class.
\end{itemize}
These are helpful in calculated metrics for skewed data, given by:
\begin{itemize}

    \item[] \textbf{Precision} gives the fraction of instance predicted as positive class that actually belong to the positive class.
    \begin{equation}
        Precision = \dfrac{TP}{TP + FP}
    \end{equation}
    \item[] \textbf{Recall} defines how well the positive class was detected out of the actual positive class.
    \begin{equation}
        Recall = \dfrac{TP}{TP + FN}
    \end{equation}
    \item[] \textbf{F-1 Score} is basically precision and recall combined in one score, it looks for a balance between both the above metrics and is given by the harmonic mean of precision and recall.
    \begin{equation}
        F1_{score} = \dfrac{2 \times Precision \times Recall}{Precision + Recall}
    \end{equation}
\end{itemize}

\section{Results and Discussion}  \label{discussion}
\begin{table}[!ht]
\centering
\caption{The average F1 score for training and testing for all the experiments}
\label{tab:F1_all}
\begin{tabular}{c|c|cc|cc|c|c|c|c|}
\cline{2-10}
\textbf{} & \multicolumn{1}{r|}{\textbf{Methodology}} & \multicolumn{2}{c|}{\textbf{SM}} & \multicolumn{2}{c|}{\textbf{SM + BB}} & \multicolumn{2}{c|}{\textbf{SM + BB}} & \multicolumn{2}{c|}{\textbf{Bimodal+}} \\
\textbf{} & \textbf{} & \textbf{} & \textbf{} & \textbf{} & \textbf{} & \multicolumn{2}{c|}{\textbf{+TS}} & \multicolumn{2}{c|}{\textbf{SM + BB + TS}} \\ \cline{3-10} 
\textbf{} & \multicolumn{1}{r|}{\textbf{(referred  as)}} & \multicolumn{2}{c|}{\textbf{(a)}} & \multicolumn{2}{c|}{\textbf{(b)}} & \multicolumn{2}{c|}{\textbf{(c)}} & \multicolumn{2}{c|}{\textbf{(d)}} \\ \hline
\multicolumn{1}{|c|}{\textbf{No.}} & \textbf{Experiments} & \multicolumn{1}{c|}{\textbf{Train}} & \textbf{Test} & \multicolumn{1}{c|}{\textbf{Train}} & \textbf{Test} & \textbf{Train} & \textbf{Test} & \textbf{Train} & \textbf{Test} \\ \hline
\multicolumn{1}{|c|}{\textbf{1}} & \textbf{\begin{tabular}[c]{@{}c@{}}Video-1,\\  classes = 18\end{tabular}} & \multicolumn{1}{c|}{0.69} & 0.47 & \multicolumn{1}{c|}{0.70} & 0.47 & 0.67 & 0.53 & 0.63 & 0.51 \\ \hline
\multicolumn{1}{|c|}{\textbf{2}} & \textbf{\begin{tabular}[c]{@{}c@{}}Video-1,\\  classes = 7\end{tabular}} & \multicolumn{1}{c|}{0.96} & 0.78 & \multicolumn{1}{c|}{0.92} & 0.74 & 0.96 & 0.84 & 0.82 & 0.72 \\ \hline
\multicolumn{1}{|c|}{\textbf{3}} & \textbf{\begin{tabular}[c]{@{}c@{}}All videos,\\ classes =   18\end{tabular}} & \multicolumn{1}{c|}{0.63} & 0.55 & \multicolumn{1}{c|}{0.72} & 0.62 & 0.73 & 0.66 & 0.49 & 0.53 \\ \hline
\multicolumn{1}{|c|}{\textbf{4}} & \textbf{\begin{tabular}[c]{@{}c@{}}All videos,\\ classes = 7\end{tabular}} & \multicolumn{1}{c|}{0.92} & 0.84 & \multicolumn{1}{c|}{0.90} & 0.85 & 0.95 & 0.91 & 0.78 & 0.75 \\ \hline
\end{tabular}
\end{table}
%%%%%%%%%%%%%%%%%%%%%%%%%%%%%%%%%%%%
\begin{figure}[!ht]
\centering
  \begin{subfigure}[b]{0.45\textwidth}
    \includegraphics[width=5.5cm,height=5.5cm,keepaspectratio]{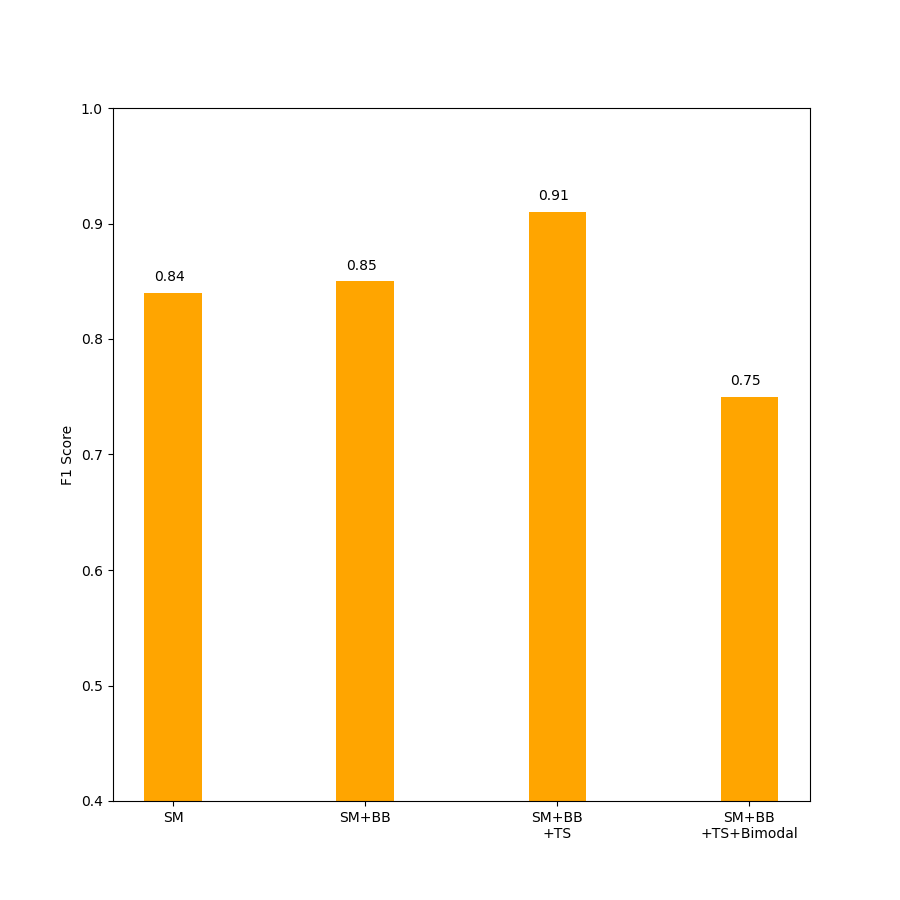}
    \caption{\centering{Evaluation of model train on all data for 7 classes}}
    \label{fig:7classes_allvideos_test}
  \end{subfigure}
  \hspace{1em}
  \begin{subfigure}[b]{0.45\textwidth}
    \includegraphics[width=5.5cm,height=5.5cm,keepaspectratio]{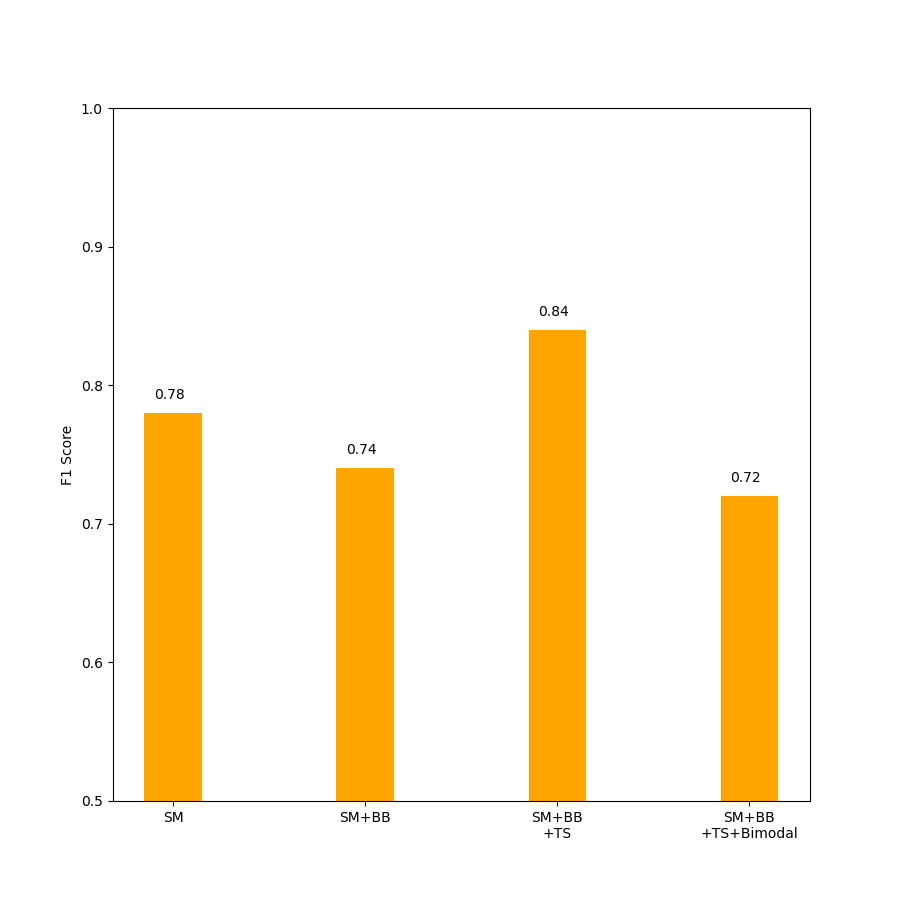}
    \caption{\centering{Evaluation of model trained on video-1 for 7 classes}}
    \label{fig:7classes_1video_test}
  \end{subfigure}
  \caption{ These plots show F-1 score evaluation results (a) on test set of all videos combined (Video-1 +Video-2 + Video-3 + Video-4) (b) on test set of Video-1. It displays results of four experiments of proposed methods, that are SM, SM+BB, SM+BB+TS and Bi-modal+SM+BB+TS as detailed in Table \ref{tab:exp} for 7 super classes}
\end{figure}
%%%%%%%%%%%%%%%%%%%%%%%%%%%%%%%%%%%%%%
\begin{figure}[!ht]
\centering
  \begin{subfigure}[b]{0.45\textwidth}
    \includegraphics[width=5.5cm,height=5.5cm,keepaspectratio]{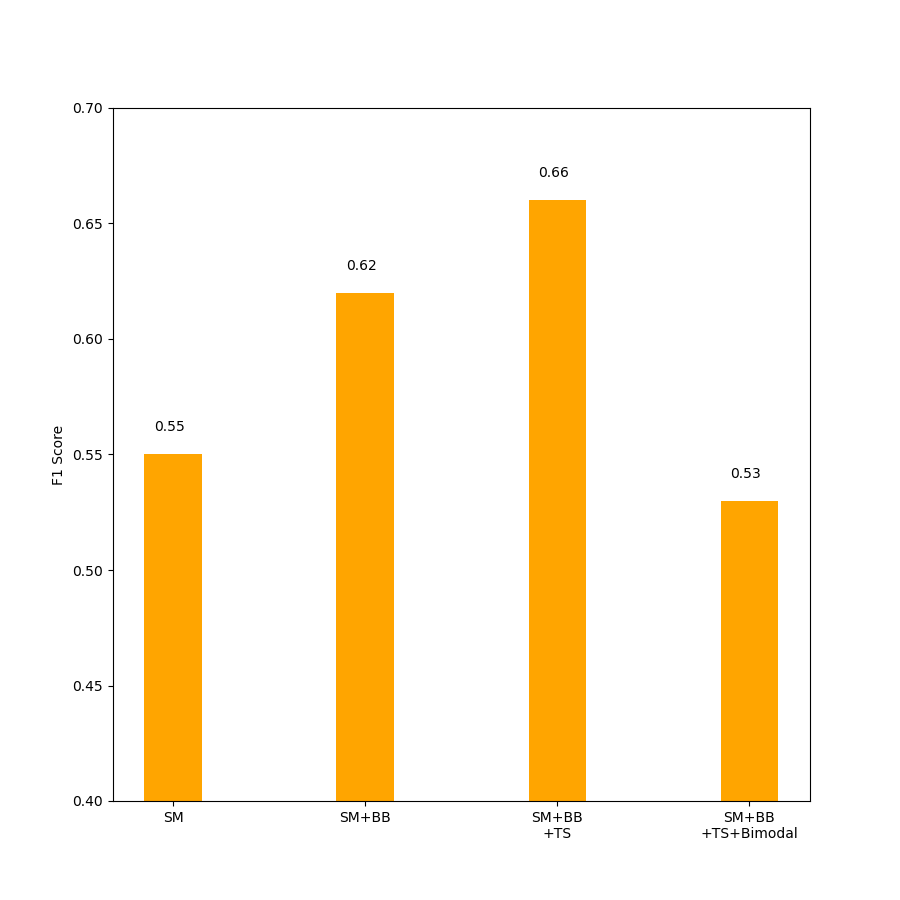}
    \caption{\centering{Evaluation of model train on all data for 18 classes}}
    \label{fig:18classes_allvideos_test}
  \end{subfigure}
  \begin{subfigure}[b]{0.45\textwidth}
    \includegraphics[width=5.5cm,height=5.5cm,keepaspectratio]{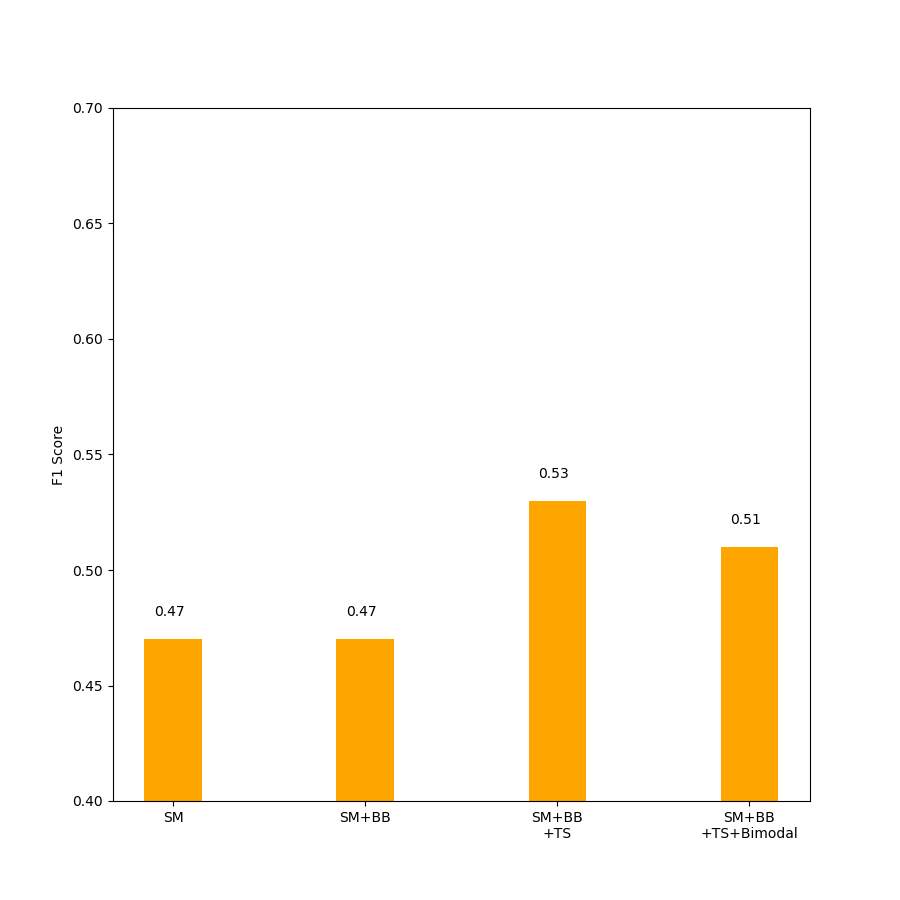}
    \caption{\centering{18 classes model on test set of Video-1}}
    \label{fig:18classes_1video_test}
  \end{subfigure}
  \caption{These plots show F-1 score evaluation results (a) on test set of all videos combined (Video-1 +Video-2 + Video-3 + Video-4) (b) on test set of Video-1. It displays results of four experiments of proposed methods, that are SM, SM+BB, SM+BB+TS and Bi-modal+SM+BB+TS as detailed in Table \ref{tab:exp} for 18 classes}
\end{figure}
%%%%%%%%%%%%%%%%%%%%%%%%%%%%%%%%%%%%%%%

In this section, we discuss the results of described experiments in Table \ref{tab:exp} that incorporates the proposed methods in different settings of dataset and classes.
The analysis of the results is further segregated based on the dataset(s) being used for experiments as a single video dataset (Video-1), all videos dataset (Video-1, Video-2, Video-3, and Video-4), and leave-one video out dataset, respectively.
Within the described analogy of dataset settings, we cover results for 18 classes and 7 classes, respectively, except for leave-one video out dataset setting. 
Further, It is to be noted that proposed methods are incorporated in accumulation during the design of the experiment to perform the comparative analysis while avoiding a large number of experiments.

1(a), 1(b), 1(c), and 1(d) in Table \ref{tab:F1_all} and \ref{fig:18classes_1video_test} shows the method performance in terms of F1 score on single video dataset for 18 classes.
1(c) obtains significant performance improvement by 6\% over baseline 1(a) on test set carried out by Temporal smoothing and preventing model to over-fit during training. 
However, the performance of \ac{dbb} in 1(b) remains the same as baseline. 
Also, 1(d) shows improved performance over baseline by 4\% but remains less than 1(c). 
Next, the performance of 7 class models shown in 2(a), 2(b), 2(c), and 2(d) respectively in Table \ref{tab:F1_all} and Figure \ref{fig:7classes_1video_test} shows the 2(c) outperform on baseline and other methods by 6\% and 8\% respectively. 

Further, the results from all video datasets, for both 18 and 7 classes, consistently shown significant improvement in Table \ref{tab:F1_all} and \ref{fig:18classes_allvideos_test}.
Referring 3(a), 3(b), and 3(c) on train set and test set shows presented methods show F1 score improvement linearly where \ac{dbb} in 3b(b) obtains 7\% improvement over baseline 3(a), and Temporal Smoothing obtains further improvement of 4\% over 3(b). 
However, the performance of 3(d) (adding audio) is low and the model remain underfit. 
Results on 7 classes model for all videos dataset in 4(a), 4(b), 4(c) also shows linear improvement in the F1 score.
Referring Figure \ref{fig:7classes_allvideos_test}, 4(c) outperformed on 4(a) and 4(b) by 7\% and 6\% respectively.
However, 4(d) remain with lower F1 than all methods.

In all the experiments mentioned above on single video and all videos dataset, the proposed methods have shown significant improvement consistently compared to the Single Method of the previous work.
\ac{dbb} and Temporal Smoothing contributed to this improvement. 
The proposed methods adapt the accumulation towards better performance generalization.
It is also shown in Table \ref{tab:3TC_7classes}, Table \ref{tab:3TC_18classes} that precision and recall of every individual class has improved in balanced manner for 7 classes and 18 classes models on single video dataset.
The same observation holds true for precision and recall for all videos dataset in Table \ref{tab:4videos_7classes} and Table \ref{tab:allvideos18classes}. 
All experiments related to audio features in Bi-modal show an under-fitting issue and require more exploration of network architecture and training process to leverage the capability of audio features.
This indicates that current audio features are not suited for this task and a future study on audio features for performance analysis would be helpful.

\begin{table}[!ht]
\centering
\caption{The average F1 score for the leave one out experiments}
\label{tab:F1_leaveoneout}
\begin{tabular}{|c|c|c|c|}
\hline\textbf{No.} & \textbf{Experiments} & \multicolumn{2}{c|}{\textbf{SM + BB+ TS}} \\ \cline{3-4} 
\textbf{} & \textbf{} & \textbf{Train} & \textbf{Test} \\ \hline
1 & Train on   video 1,2 \& 3, test on video 4 & 0.97 & 0.21 \\ \hline
2 & Train on   video 2, 3 \& 4, test on video 1 & 0.96 & 0.27 \\ \hline
3 & Train on   video 1, 3\& 4, test on video 2 & 0.97 & 0.19 \\ \hline
4 & Train on   video 1,2 \& 4, test on video 3 & 0.98 & 0.29 \\ \hline
\end{tabular}
\end{table}
Table \ref{tab:F1_leaveoneout} presents the results of experiments using `leave one video out' described in the last section of Table \ref{tab:exp} where the method being used includes single Model, \ac{dbb}, and Temporal Smoothing. 
In this data setting, all the frames of one video (the entire video, respectively) is used for testing while the other three videos are used in training. 
Results of leave one out experiments in all four cases show poor performance generalization on an unseen test video for all four experiments. 
It shows an average F1 score 0.21, 0.27, 0.19, and 0.29 when model training by leaving test videos as video 4, video 1, video 2, and video 3, respectively. 
While being better than random guessing, there is still a lot of room for improvement.
One of the reasons for poor generalization is related to different data distribution as mentioned in the dataset section.
Furthermore, two of the videos have plain uniform background whereas others have a noisy background with the presence of people. 
When the 'leave one out' protocol is followed, it always exposes the model to train on only one dataset with similar characteristics, whereas the other two videos belong to different nature of background. 
Hence, the model does not get sufficient training exposure for the video being tested. 
This reveals the further work directions to evaluate proposed methods on balanced data environment and identifying features that are less sensitive with a background in general.
Sophisticated data augmentation strategies would also be a promising line of future research.

\section{Conclusion}  \label{conclusion}
This paper presented a study of the effectiveness of deep learning methods (3D \ac{cnn}) on the task of labelling performance gestures in music performance videos.
The study comprises several methods and tailored extensions to the task at hand and performs experiments on single videos, four videos, and previously unseen videos.
Furthermore, the usefulness of including audio features is investigated.

The results of testing various models on small (video-1 only) and larger datasets (all 4 videos combined) suggest that the improved models presented in this work significantly outperform those presented previously \cite{anonym}.
Improvements include: (i) use of Single Model with 1D Vector output, (ii) \ac{dbb}, and (iii) Temporal Smoothing.
The final F-measure on four videos is 53\,\% for 18 classes and 75\,\% for the 7 more general classes of performance gestures. 
Results improvement by linear accumulations of mentioned methods convey that: 
\begin{enumerate}
    \item \ac{dbb} reduces the impact of class imbalance (which is not possible with oversampling of minority classes),
    \item Single Model with 1D Vector simplifies network to prevent overfitting, and
    \item Temporal Smoothing obtains the robust representations of musical gestures based on their temporal appearance and time duration in each sample
\end{enumerate}

It is noticeable that the experiments which involve proposed methods as the Single Model with 1D Vector output, \ac{dbb}, and Temporal Smoothing all together give the best results.
However, further improvements can be made with regards to the musical gesture recognition and also a better generalization of the model since the results seem to be still over-fitted.
Similarly, in the bi-modal experiments including audio, there is more room for improvement, \ie explore suitable features, network architecture, and training parameters, in order to overcome the underfitting effect. 
Using additional audio features and also employing a pretrained model for music information retrieval could be helpful. 

The main outcome of this study is that it is possible to automate the process of labelling performance gestures in music performance videos.
In a realistic scenario, as of today, one could annotate a part of the video and the rest of the annotation could be automated (or at least partially automated with corrections added by a human expert). This would result in significant time savings in the labelling process of videos for research in music performance.

In the future, by using more diverse data (in terms of musical instruments used, music being performed, backgrounds, and other characteristics) we expect to further improve the performance of the deep learning approaches and obtain more reliable automated labelling.

\section*{Acknowledgements}
 % The authors would like to thank Pedro Malo for his valuable contribution towards the realization of this work.
 The authors would like to thank Pedro Malo Perise for his valuable contribution towards the realization of this work.
 
\bibliographystyle{apacite}
\bibliography{main.bib}

\begin{thebibliography}{}

\bibitem [\protect \citeauthoryear {%
Baltru{\v{s}}aitis%
, Ahuja%
\BCBL {}\ \BBA {} Morency%
}{%
Baltru{\v{s}}aitis%
\ \protect \BOthers {.}}{%
{\protect \APACyear {2017}}%
}]{%
Baltrusaitis2017}
\APACinsertmetastar {%
Baltrusaitis2017}%
\begin{APACrefauthors}%
Baltru{\v{s}}aitis, T.%
, Ahuja, C.%
\BCBL {}\ \BBA {} Morency, L\BPBI P.%
\end{APACrefauthors}%
\unskip\
\newblock
\APACrefYearMonthDay{2017}{}{}.
\newblock
{\BBOQ}\APACrefatitle {{Multimodal machine learning: A survey and taxonomy}}
  {{Multimodal machine learning: A survey and taxonomy}}.{\BBCQ}
\newblock
\APACjournalVolNumPages{arXiv}{41}{2}{423--443}.
\PrintBackRefs{\CurrentBib}

\bibitem [\protect \citeauthoryear {%
Bengio%
, Courville%
\BCBL {}\ \BBA {} Vincent%
}{%
Bengio%
\ \protect \BOthers {.}}{%
{\protect \APACyear {2014}}%
}]{%
bengio2014representation}
\APACinsertmetastar {%
bengio2014representation}%
\begin{APACrefauthors}%
Bengio, Y.%
, Courville, A.%
\BCBL {}\ \BBA {} Vincent, P.%
\end{APACrefauthors}%
\unskip\
\newblock
\APACrefYearMonthDay{2014}{}{}.
\newblock
\APACrefbtitle {Representation Learning: A Review and New Perspectives.}
  {Representation learning: A review and new perspectives.}
\PrintBackRefs{\CurrentBib}

\bibitem [\protect \citeauthoryear {%
Bowen%
}{%
Bowen%
}{%
{\protect \APACyear {1996}}%
}]{%
bowen1996performance}
\APACinsertmetastar {%
bowen1996performance}%
\begin{APACrefauthors}%
Bowen, J\BPBI A.%
\end{APACrefauthors}%
\unskip\
\newblock
\APACrefYearMonthDay{1996}{}{}.
\newblock
{\BBOQ}\APACrefatitle {Performance practice versus performance analysis: Why
  should performers study performance} {Performance practice versus performance
  analysis: Why should performers study performance}.{\BBCQ}
\newblock
\APACjournalVolNumPages{Performance Practice Review}{9}{1}{3}.
\PrintBackRefs{\CurrentBib}

\bibitem [\protect \citeauthoryear {%
Caba~Heilbron%
, Escorcia%
, Ghanem%
\BCBL {}\ \BBA {} Carlos~Niebles%
}{%
Caba~Heilbron%
\ \protect \BOthers {.}}{%
{\protect \APACyear {2015}}%
}]{%
caba2015activitynet}
\APACinsertmetastar {%
caba2015activitynet}%
\begin{APACrefauthors}%
Caba~Heilbron, F.%
, Escorcia, V.%
, Ghanem, B.%
\BCBL {}\ \BBA {} Carlos~Niebles, J.%
\end{APACrefauthors}%
\unskip\
\newblock
\APACrefYearMonthDay{2015}{}{}.
\newblock
{\BBOQ}\APACrefatitle {Activitynet: A large-scale video benchmark for human
  activity understanding} {Activitynet: A large-scale video benchmark for human
  activity understanding}.{\BBCQ}
\newblock
\BIn{} \APACrefbtitle {Proceedings of the ieee conference on computer vision
  and pattern recognition} {Proceedings of the ieee conference on computer
  vision and pattern recognition}\ (\BPGS\ 961--970).
\PrintBackRefs{\CurrentBib}

\bibitem [\protect \citeauthoryear {%
Carreira%
\ \BBA {} Zisserman%
}{%
Carreira%
\ \BBA {} Zisserman%
}{%
{\protect \APACyear {2017}}%
}]{%
carreira2017quo}
\APACinsertmetastar {%
carreira2017quo}%
\begin{APACrefauthors}%
Carreira, J.%
\BCBT {}\ \BBA {} Zisserman, A.%
\end{APACrefauthors}%
\unskip\
\newblock
\APACrefYearMonthDay{2017}{}{}.
\newblock
{\BBOQ}\APACrefatitle {Quo vadis, action recognition? a new model and the
  kinetics dataset} {Quo vadis, action recognition? a new model and the
  kinetics dataset}.{\BBCQ}
\newblock
\BIn{} \APACrefbtitle {proceedings of the IEEE Conference on Computer Vision
  and Pattern Recognition} {proceedings of the ieee conference on computer
  vision and pattern recognition}\ (\BPGS\ 6299--6308).
\PrintBackRefs{\CurrentBib}

\bibitem [\protect \citeauthoryear {%
Coorevits%
, Moelants%
, {\"O}stersj{\"o}%
, Gorton%
\BCBL {}\ \BBA {} Leman%
}{%
Coorevits%
\ \protect \BOthers {.}}{%
{\protect \APACyear {2015}}%
}]{%
coorevits2015decomposing}
\APACinsertmetastar {%
coorevits2015decomposing}%
\begin{APACrefauthors}%
Coorevits, E.%
, Moelants, D.%
, {\"O}stersj{\"o}, S.%
, Gorton, D.%
\BCBL {}\ \BBA {} Leman, M.%
\end{APACrefauthors}%
\unskip\
\newblock
\APACrefYearMonthDay{2015}{}{}.
\newblock
{\BBOQ}\APACrefatitle {Decomposing a composition: On the multi-layered analysis
  of expressive music performance} {Decomposing a composition: On the
  multi-layered analysis of expressive music performance}.{\BBCQ}
\newblock
\BIn{} \APACrefbtitle {International Symposium on Computer Music
  Multidisciplinary Research} {International symposium on computer music
  multidisciplinary research}\ (\BPGS\ 167--189).
\PrintBackRefs{\CurrentBib}

\bibitem [\protect \citeauthoryear {%
Deng%
\ \protect \BOthers {.}}{%
Deng%
\ \protect \BOthers {.}}{%
{\protect \APACyear {2009}}%
}]{%
deng2009imagenet}
\APACinsertmetastar {%
deng2009imagenet}%
\begin{APACrefauthors}%
Deng, J.%
, Dong, W.%
, Socher, R.%
, Li, L\BHBI J.%
, Li, K.%
\BCBL {}\ \BBA {} Fei-Fei, L.%
\end{APACrefauthors}%
\unskip\
\newblock
\APACrefYearMonthDay{2009}{}{}.
\newblock
{\BBOQ}\APACrefatitle {Imagenet: A large-scale hierarchical image database}
  {Imagenet: A large-scale hierarchical image database}.{\BBCQ}
\newblock
\BIn{} \APACrefbtitle {2009 IEEE conference on computer vision and pattern
  recognition} {2009 ieee conference on computer vision and pattern
  recognition}\ (\BPGS\ 248--255).
\PrintBackRefs{\CurrentBib}

\bibitem [\protect \citeauthoryear {%
Donahue%
\ \protect \BOthers {.}}{%
Donahue%
\ \protect \BOthers {.}}{%
{\protect \APACyear {2015}}%
}]{%
donahue2015long}
\APACinsertmetastar {%
donahue2015long}%
\begin{APACrefauthors}%
Donahue, J.%
, Anne~Hendricks, L.%
, Guadarrama, S.%
, Rohrbach, M.%
, Venugopalan, S.%
, Saenko, K.%
\BCBL {}\ \BBA {} Darrell, T.%
\end{APACrefauthors}%
\unskip\
\newblock
\APACrefYearMonthDay{2015}{}{}.
\newblock
{\BBOQ}\APACrefatitle {Long-term recurrent convolutional networks for visual
  recognition and description} {Long-term recurrent convolutional networks for
  visual recognition and description}.{\BBCQ}
\newblock
\BIn{} \APACrefbtitle {Proceedings of the IEEE conference on computer vision
  and pattern recognition} {Proceedings of the ieee conference on computer
  vision and pattern recognition}\ (\BPGS\ 2625--2634).
\PrintBackRefs{\CurrentBib}

\bibitem [\protect \citeauthoryear {%
Dong%
, Gong%
\BCBL {}\ \BBA {} Zhu%
}{%
Dong%
\ \protect \BOthers {.}}{%
{\protect \APACyear {2018}}%
}]{%
dong2018imbalanced}
\APACinsertmetastar {%
dong2018imbalanced}%
\begin{APACrefauthors}%
Dong, Q.%
, Gong, S.%
\BCBL {}\ \BBA {} Zhu, X.%
\end{APACrefauthors}%
\unskip\
\newblock
\APACrefYearMonthDay{2018}{}{}.
\newblock
{\BBOQ}\APACrefatitle {Imbalanced deep learning by minority class incremental
  rectification} {Imbalanced deep learning by minority class incremental
  rectification}.{\BBCQ}
\newblock
\APACjournalVolNumPages{IEEE transactions on pattern analysis and machine
  intelligence}{41}{6}{1367--1381}.
\PrintBackRefs{\CurrentBib}

\bibitem [\protect \citeauthoryear {%
God{\o}y%
\ \BBA {} Leman%
}{%
God{\o}y%
\ \BBA {} Leman%
}{%
{\protect \APACyear {2010}}%
}]{%
godoy2010musical}
\APACinsertmetastar {%
godoy2010musical}%
\begin{APACrefauthors}%
God{\o}y, R\BPBI I.%
\BCBT {}\ \BBA {} Leman, M.%
\end{APACrefauthors}%
\unskip\
\newblock
\APACrefYear{2010}.
\newblock
\APACrefbtitle {Musical gestures: Sound, movement, and meaning} {Musical
  gestures: Sound, movement, and meaning}.
\newblock
\APACaddressPublisher{}{Routledge}.
\PrintBackRefs{\CurrentBib}

\bibitem [\protect \citeauthoryear {%
Goibert%
\ \BBA {} Dohmatob%
}{%
Goibert%
\ \BBA {} Dohmatob%
}{%
{\protect \APACyear {2019}}%
}]{%
goibert2019adversarial}
\APACinsertmetastar {%
goibert2019adversarial}%
\begin{APACrefauthors}%
Goibert, M.%
\BCBT {}\ \BBA {} Dohmatob, E.%
\end{APACrefauthors}%
\unskip\
\newblock
\APACrefYearMonthDay{2019}{}{}.
\newblock
{\BBOQ}\APACrefatitle {Adversarial robustness via label-smoothing} {Adversarial
  robustness via label-smoothing}.{\BBCQ}
\newblock
\APACjournalVolNumPages{arXiv preprint arXiv:1906.11567}{}{}{}.
\PrintBackRefs{\CurrentBib}

\bibitem [\protect \citeauthoryear {%
Goodfellow%
, Bengio%
\BCBL {}\ \BBA {} Courville%
}{%
Goodfellow%
\ \protect \BOthers {.}}{%
{\protect \APACyear {2016}}%
}]{%
goodfellow2016deep}
\APACinsertmetastar {%
goodfellow2016deep}%
\begin{APACrefauthors}%
Goodfellow, I.%
, Bengio, Y.%
\BCBL {}\ \BBA {} Courville, A.%
\end{APACrefauthors}%
\unskip\
\newblock
\APACrefYear{2016}.
\newblock
\APACrefbtitle {Deep learning} {Deep learning}.
\newblock
\APACaddressPublisher{}{MIT press}.
\PrintBackRefs{\CurrentBib}

\bibitem [\protect \citeauthoryear {%
Gorton%
\ \BBA {} {\"{O}}stersj{\"{o}}%
}{%
Gorton%
\ \BBA {} {\"{O}}stersj{\"{o}}%
}{%
{\protect \APACyear {2019}}%
}]{%
Ostersjo2019}
\APACinsertmetastar {%
Ostersjo2019}%
\begin{APACrefauthors}%
Gorton, D.%
\BCBT {}\ \BBA {} {\"{O}}stersj{\"{o}}, S.%
\end{APACrefauthors}%
\unskip\
\newblock
\APACrefYearMonthDay{2019}{nov}{}.
\newblock
{\BBOQ}\APACrefatitle {{Austerity Measures I:}} {{Austerity Measures
  I:}}.{\BBCQ}
\newblock
\BIn{} C.~Laws, W.~Brooks, D.~Gorton, N\BPBI T.~Thuy,
  S.~{\"{O}}stersj{\"{o}}\BCBL {}\ \BBA {} J\BPBI J.~Wells\ (\BEDS),
  \APACrefbtitle {Voices, Bodies, Practices} {Voices, bodies, practices}\
  (\BPGS\ 29--80).
\newblock
\APACaddressPublisher{}{Universitaire Pers Leuven}.
\newblock
\begin{APACrefDOI} \doi{10.2307/j.ctvmd83kv.6} \end{APACrefDOI}
\PrintBackRefs{\CurrentBib}

\bibitem [\protect \citeauthoryear {%
Hara%
, Kataoka%
\BCBL {}\ \BBA {} Satoh%
}{%
Hara%
\ \protect \BOthers {.}}{%
{\protect \APACyear {2018}}%
}]{%
hara3dcnns}
\APACinsertmetastar {%
hara3dcnns}%
\begin{APACrefauthors}%
Hara, K.%
, Kataoka, H.%
\BCBL {}\ \BBA {} Satoh, Y.%
\end{APACrefauthors}%
\unskip\
\newblock
\APACrefYearMonthDay{2018}{}{}.
\newblock
{\BBOQ}\APACrefatitle {{Can Spatiotemporal 3D CNNs Retrace the History of 2D
  CNNs and ImageNet?}} {{Can Spatiotemporal 3D CNNs Retrace the History of 2D
  CNNs and ImageNet?}}{\BBCQ}
\newblock
\APACjournalVolNumPages{Proceedings of the IEEE Computer Society Conference on
  Computer Vision and Pattern Recognition}{}{}{6546--6555}.
\newblock
\begin{APACrefDOI} \doi{10.1109/CVPR.2018.00685} \end{APACrefDOI}
\PrintBackRefs{\CurrentBib}

\bibitem [\protect \citeauthoryear {%
{He}%
\ \BBA {} {Garcia}%
}{%
{He}%
\ \BBA {} {Garcia}%
}{%
{\protect \APACyear {2009}}%
}]{%
metrics}
\APACinsertmetastar {%
metrics}%
\begin{APACrefauthors}%
{He}, H.%
\BCBT {}\ \BBA {} {Garcia}, E\BPBI A.%
\end{APACrefauthors}%
\unskip\
\newblock
\APACrefYearMonthDay{2009}{}{}.
\newblock
{\BBOQ}\APACrefatitle {Learning from Imbalanced Data} {Learning from imbalanced
  data}.{\BBCQ}
\newblock
\APACjournalVolNumPages{IEEE Transactions on Knowledge and Data
  Engineering}{21}{9}{1263-1284}.
\newblock
\begin{APACrefDOI} \doi{10.1109/TKDE.2008.239} \end{APACrefDOI}
\PrintBackRefs{\CurrentBib}

\bibitem [\protect \citeauthoryear {%
Herbert%
, Clarke%
\BCBL {}\ \BBA {} Clarke%
}{%
Herbert%
\ \protect \BOthers {.}}{%
{\protect \APACyear {2019}}%
}]{%
herbert2019music}
\APACinsertmetastar {%
herbert2019music}%
\begin{APACrefauthors}%
Herbert, R.%
, Clarke, D.%
\BCBL {}\ \BBA {} Clarke, E.%
\end{APACrefauthors}%
\unskip\
\newblock
\APACrefYear{2019}.
\newblock
\APACrefbtitle {Music and Consciousness 2: Worlds, Practices, Modalities}
  {Music and consciousness 2: Worlds, practices, modalities}.
\newblock
\APACaddressPublisher{}{Oxford University Press}.
\PrintBackRefs{\CurrentBib}

\bibitem [\protect \citeauthoryear {%
Hochreiter%
\ \BBA {} Schmidhuber%
}{%
Hochreiter%
\ \BBA {} Schmidhuber%
}{%
{\protect \APACyear {1997}}%
}]{%
hochreiter1997long}
\APACinsertmetastar {%
hochreiter1997long}%
\begin{APACrefauthors}%
Hochreiter, S.%
\BCBT {}\ \BBA {} Schmidhuber, J.%
\end{APACrefauthors}%
\unskip\
\newblock
\APACrefYearMonthDay{1997}{}{}.
\newblock
{\BBOQ}\APACrefatitle {Long short-term memory} {Long short-term memory}.{\BBCQ}
\newblock
\APACjournalVolNumPages{Neural computation}{9}{8}{1735--1780}.
\PrintBackRefs{\CurrentBib}

\bibitem [\protect \citeauthoryear {%
Jensenius%
}{%
Jensenius%
}{%
{\protect \APACyear {2013}}%
}]{%
jensenius2013action}
\APACinsertmetastar {%
jensenius2013action}%
\begin{APACrefauthors}%
Jensenius, A\BPBI R.%
\end{APACrefauthors}%
\unskip\
\newblock
\APACrefYearMonthDay{2013}{}{}.
\newblock
{\BBOQ}\APACrefatitle {An action--sound approach to teaching interactive music}
  {An action--sound approach to teaching interactive music}.{\BBCQ}
\newblock
\APACjournalVolNumPages{Organised Sound}{18}{2}{178--189}.
\PrintBackRefs{\CurrentBib}

\bibitem [\protect \citeauthoryear {%
Jhuang%
, Garrote%
, Poggio%
, Serre%
\BCBL {}\ \BBA {} Hmdb%
}{%
Jhuang%
\ \protect \BOthers {.}}{%
{\protect \APACyear {2011}}%
}]{%
jhuang2011large}
\APACinsertmetastar {%
jhuang2011large}%
\begin{APACrefauthors}%
Jhuang, H.%
, Garrote, H.%
, Poggio, E.%
, Serre, T.%
\BCBL {}\ \BBA {} Hmdb, T.%
\end{APACrefauthors}%
\unskip\
\newblock
\APACrefYearMonthDay{2011}{}{}.
\newblock
{\BBOQ}\APACrefatitle {HMDB: A large video database for human motion
  recognition} {Hmdb: A large video database for human motion
  recognition}.{\BBCQ}
\newblock
\BIn{} \APACrefbtitle {Proc. of IEEE International Conference on Computer
  Vision} {Proc. of ieee international conference on computer vision}\
  (\BVOL~4, \BPG~6).
\PrintBackRefs{\CurrentBib}

\bibitem [\protect \citeauthoryear {%
Karpathy%
\ \protect \BOthers {.}}{%
Karpathy%
\ \protect \BOthers {.}}{%
{\protect \APACyear {2014}}%
}]{%
karpathy2014large}
\APACinsertmetastar {%
karpathy2014large}%
\begin{APACrefauthors}%
Karpathy, A.%
, Toderici, G.%
, Shetty, S.%
, Leung, T.%
, Sukthankar, R.%
\BCBL {}\ \BBA {} Fei-Fei, L.%
\end{APACrefauthors}%
\unskip\
\newblock
\APACrefYearMonthDay{2014}{}{}.
\newblock
{\BBOQ}\APACrefatitle {Large-scale video classification with convolutional
  neural networks} {Large-scale video classification with convolutional neural
  networks}.{\BBCQ}
\newblock
\BIn{} \APACrefbtitle {Proceedings of the IEEE conference on Computer Vision
  and Pattern Recognition} {Proceedings of the ieee conference on computer
  vision and pattern recognition}\ (\BPGS\ 1725--1732).
\PrintBackRefs{\CurrentBib}

\bibitem [\protect \citeauthoryear {%
Kataoka%
, Wakamiya%
, Hara%
\BCBL {}\ \BBA {} Satoh%
}{%
Kataoka%
\ \protect \BOthers {.}}{%
{\protect \APACyear {2020}}%
}]{%
kataoka2020megascale}
\APACinsertmetastar {%
kataoka2020megascale}%
\begin{APACrefauthors}%
Kataoka, H.%
, Wakamiya, T.%
, Hara, K.%
\BCBL {}\ \BBA {} Satoh, Y.%
\end{APACrefauthors}%
\unskip\
\newblock
\APACrefYearMonthDay{2020}{}{}.
\newblock
\APACrefbtitle {Would Mega-scale Datasets Further Enhance Spatiotemporal 3D
  CNNs?} {Would mega-scale datasets further enhance spatiotemporal 3d cnns?}
\PrintBackRefs{\CurrentBib}

\bibitem [\protect \citeauthoryear {%
Kingma%
\ \BBA {} Ba%
}{%
Kingma%
\ \BBA {} Ba%
}{%
{\protect \APACyear {2017}}%
}]{%
kingma2017adam}
\APACinsertmetastar {%
kingma2017adam}%
\begin{APACrefauthors}%
Kingma, D\BPBI P.%
\BCBT {}\ \BBA {} Ba, J.%
\end{APACrefauthors}%
\unskip\
\newblock
\APACrefYearMonthDay{2017}{}{}.
\newblock
\APACrefbtitle {Adam: A Method for Stochastic Optimization.} {Adam: A method
  for stochastic optimization.}
\PrintBackRefs{\CurrentBib}

\bibitem [\protect \citeauthoryear {%
Krizhevsky%
, Sutskever%
\BCBL {}\ \BBA {} Hinton%
}{%
Krizhevsky%
\ \protect \BOthers {.}}{%
{\protect \APACyear {2012}}%
}]{%
NIPS2012_4824}
\APACinsertmetastar {%
NIPS2012_4824}%
\begin{APACrefauthors}%
Krizhevsky, A.%
, Sutskever, I.%
\BCBL {}\ \BBA {} Hinton, G\BPBI E.%
\end{APACrefauthors}%
\unskip\
\newblock
\APACrefYearMonthDay{2012}{}{}.
\newblock
{\BBOQ}\APACrefatitle {ImageNet Classification with Deep Convolutional Neural
  Networks} {Imagenet classification with deep convolutional neural
  networks}.{\BBCQ}
\newblock
\BIn{} F.~Pereira, C\BPBI J\BPBI C.~Burges, L.~Bottou\BCBL {}\ \BBA {} K\BPBI
  Q.~Weinberger\ (\BEDS), \APACrefbtitle {Advances in Neural Information
  Processing Systems 25} {Advances in neural information processing systems
  25}\ (\BPGS\ 1097--1105).
\newblock
\APACaddressPublisher{}{Curran Associates, Inc.}
\newblock
\begin{APACrefURL}
  \url{http://papers.nips.cc/paper/4824-imagenet-classification-with-deep-convolutional-neural-networks.pdf}
  \end{APACrefURL}
\PrintBackRefs{\CurrentBib}

\bibitem [\protect \citeauthoryear {%
MacRitchie%
, Buck%
\BCBL {}\ \BBA {} Bailey%
}{%
MacRitchie%
\ \protect \BOthers {.}}{%
{\protect \APACyear {2013}}%
}]{%
macritchie2013inferring}
\APACinsertmetastar {%
macritchie2013inferring}%
\begin{APACrefauthors}%
MacRitchie, J.%
, Buck, B.%
\BCBL {}\ \BBA {} Bailey, N\BPBI J.%
\end{APACrefauthors}%
\unskip\
\newblock
\APACrefYearMonthDay{2013}{}{}.
\newblock
{\BBOQ}\APACrefatitle {Inferring musical structure through bodily gestures}
  {Inferring musical structure through bodily gestures}.{\BBCQ}
\newblock
\APACjournalVolNumPages{Musicae Scientiae}{17}{1}{86--108}.
\PrintBackRefs{\CurrentBib}

\bibitem [\protect \citeauthoryear {%
Mitrović%
, Zeppelzauer%
\BCBL {}\ \BBA {} Breiteneder%
}{%
Mitrović%
\ \protect \BOthers {.}}{%
{\protect \APACyear {2010}}%
}]{%
MITROVIC201071}
\APACinsertmetastar {%
MITROVIC201071}%
\begin{APACrefauthors}%
Mitrović, D.%
, Zeppelzauer, M.%
\BCBL {}\ \BBA {} Breiteneder, C.%
\end{APACrefauthors}%
\unskip\
\newblock
\APACrefYearMonthDay{2010}{}{}.
\newblock
{\BBOQ}\APACrefatitle {Chapter 3 - Features for Content-Based Audio Retrieval}
  {Chapter 3 - features for content-based audio retrieval}.{\BBCQ}
\newblock
\BIn{} \APACrefbtitle {Advances in Computers: Improving the Web} {Advances in
  computers: Improving the web}\ (\BVOL~78, \BPG~71-150).
\newblock
\APACaddressPublisher{}{Elsevier}.
\PrintBackRefs{\CurrentBib}

\bibitem [\protect \citeauthoryear {%
M\"{u}ller%
, Kornblith%
\BCBL {}\ \BBA {} Hinton%
}{%
M\"{u}ller%
\ \protect \BOthers {.}}{%
{\protect \APACyear {2019}}%
}]{%
NEURIPS2019_f1748d6b}
\APACinsertmetastar {%
NEURIPS2019_f1748d6b}%
\begin{APACrefauthors}%
M\"{u}ller, R.%
, Kornblith, S.%
\BCBL {}\ \BBA {} Hinton, G\BPBI E.%
\end{APACrefauthors}%
\unskip\
\newblock
\APACrefYearMonthDay{2019}{}{}.
\newblock
{\BBOQ}\APACrefatitle {When does label smoothing help?} {When does label
  smoothing help?}{\BBCQ}
\newblock
\BIn{} H.~Wallach, H.~Larochelle, A.~Beygelzimer, F.~d\textquotesingle
  Alch\'{e}-Buc, E.~Fox\BCBL {}\ \BBA {} R.~Garnett\ (\BEDS), \APACrefbtitle
  {Advances in Neural Information Processing Systems} {Advances in neural
  information processing systems}\ (\BVOL~32).
\newblock
\APACaddressPublisher{}{Curran Associates, Inc.}
\PrintBackRefs{\CurrentBib}

\bibitem [\protect \citeauthoryear {%
Paszke%
\ \protect \BOthers {.}}{%
Paszke%
\ \protect \BOthers {.}}{%
{\protect \APACyear {2019}}%
}]{%
pytorch}
\APACinsertmetastar {%
pytorch}%
\begin{APACrefauthors}%
Paszke, A.%
, Gross, S.%
, Massa, F.%
, Lerer, A.%
, Bradbury, J.%
, Chanan, G.%
\BDBL {}Chintala, S.%
\end{APACrefauthors}%
\unskip\
\newblock
\APACrefYearMonthDay{2019}{}{}.
\newblock
{\BBOQ}\APACrefatitle {PyTorch: An Imperative Style, High-Performance Deep
  Learning Library} {Pytorch: An imperative style, high-performance deep
  learning library}.{\BBCQ}
\newblock
\BIn{} H.~Wallach, H.~Larochelle, A.~Beygelzimer, F.~d\textquotesingle
  Alch\'{e}-Buc, E.~Fox\BCBL {}\ \BBA {} R.~Garnett\ (\BEDS), \APACrefbtitle
  {Advances in Neural Information Processing Systems 32} {Advances in neural
  information processing systems 32}\ (\BPGS\ 8024--8035).
\newblock
\APACaddressPublisher{}{Curran Associates, Inc.}
\PrintBackRefs{\CurrentBib}

\bibitem [\protect \citeauthoryear {%
Perry%
}{%
Perry%
}{%
{\protect \APACyear {2010}}%
}]{%
perry2010weighted}
\APACinsertmetastar {%
perry2010weighted}%
\begin{APACrefauthors}%
Perry, M\BPBI B.%
\end{APACrefauthors}%
\unskip\
\newblock
\APACrefYearMonthDay{2010}{}{}.
\newblock
{\BBOQ}\APACrefatitle {The Weighted Moving Average Technique} {The weighted
  moving average technique}.{\BBCQ}
\newblock
\APACjournalVolNumPages{Wiley Encyclopedia of Operations Research and
  Management Science}{}{}{}.
\PrintBackRefs{\CurrentBib}

\bibitem [\protect \citeauthoryear {%
{Pukhova}%
, {Kustov}%
\BCBL {}\ \BBA {} {Ferrini}%
}{%
{Pukhova}%
\ \protect \BOthers {.}}{%
{\protect \APACyear {2018}}%
}]{%
time_freq_analysis}
\APACinsertmetastar {%
time_freq_analysis}%
\begin{APACrefauthors}%
{Pukhova}, V\BPBI M.%
, {Kustov}, T\BPBI V.%
\BCBL {}\ \BBA {} {Ferrini}, G.%
\end{APACrefauthors}%
\unskip\
\newblock
\APACrefYearMonthDay{2018}{}{}.
\newblock
{\BBOQ}\APACrefatitle {Time-frequency analysis of non-stationary signals}
  {Time-frequency analysis of non-stationary signals}.{\BBCQ}
\newblock
\BIn{} \APACrefbtitle {2018 IEEE Conference of Russian Young Researchers in
  Electrical and Electronic Engineering (EIConRus)} {2018 ieee conference of
  russian young researchers in electrical and electronic engineering
  (eiconrus)}\ (\BPG~1141-1145).
\newblock
\begin{APACrefDOI} \doi{10.1109/EIConRus.2018.8317292} \end{APACrefDOI}
\PrintBackRefs{\CurrentBib}

\bibitem [\protect \citeauthoryear {%
Qiu%
\ \BBA {} Song%
}{%
Qiu%
\ \BBA {} Song%
}{%
{\protect \APACyear {2018}}%
}]{%
qiu2018nonuniform}
\APACinsertmetastar {%
qiu2018nonuniform}%
\begin{APACrefauthors}%
Qiu, Q.%
\BCBT {}\ \BBA {} Song, Z.%
\end{APACrefauthors}%
\unskip\
\newblock
\APACrefYearMonthDay{2018}{}{}.
\newblock
{\BBOQ}\APACrefatitle {A nonuniform weighted loss function for imbalanced image
  classification} {A nonuniform weighted loss function for imbalanced image
  classification}.{\BBCQ}
\newblock
\BIn{} \APACrefbtitle {Proceedings of the 2018 international conference on
  image and graphics processing} {Proceedings of the 2018 international
  conference on image and graphics processing}\ (\BPGS\ 78--82).
\PrintBackRefs{\CurrentBib}

\bibitem [\protect \citeauthoryear {%
{R.}%
, {S.}%
\BCBL {}\ \BBA {} {D.}%
}{%
{R.}%
\ \protect \BOthers {.}}{%
{\protect \APACyear {2018}}%
}]{%
audiofeatures}
\APACinsertmetastar {%
audiofeatures}%
\begin{APACrefauthors}%
{R.}, G\BPBI S.%
, {S.}, B\BPBI S.%
\BCBL {}\ \BBA {} {D.}, S\BPBI S.%
\end{APACrefauthors}%
\unskip\
\newblock
\APACrefYearMonthDay{2018}{}{}.
\newblock
{\BBOQ}\APACrefatitle {Cepstral (MFCC) Feature and Spectral (Timbral) Features
  Analysis for Musical Instrument Sounds} {Cepstral (mfcc) feature and spectral
  (timbral) features analysis for musical instrument sounds}.{\BBCQ}
\newblock
\BIn{} \APACrefbtitle {2018 IEEE Global Conference on Wireless Computing and
  Networking (GCWCN)} {2018 ieee global conference on wireless computing and
  networking (gcwcn)}\ (\BPG~109-113).
\newblock
\begin{APACrefDOI} \doi{10.1109/GCWCN.2018.8668628} \end{APACrefDOI}
\PrintBackRefs{\CurrentBib}

\bibitem [\protect \citeauthoryear {%
Reed%
\ \protect \BOthers {.}}{%
Reed%
\ \protect \BOthers {.}}{%
{\protect \APACyear {2015}}%
}]{%
DBLP}
\APACinsertmetastar {%
DBLP}%
\begin{APACrefauthors}%
Reed, S.%
, Lee, H.%
, Anguelov, D.%
, Szegedy, C.%
, Erhan, D.%
\BCBL {}\ \BBA {} Rabinovich, A.%
\end{APACrefauthors}%
\unskip\
\newblock
\APACrefYearMonthDay{2015}{}{}.
\newblock
\APACrefbtitle {Training Deep Neural Networks on Noisy Labels with
  Bootstrapping.} {Training deep neural networks on noisy labels with
  bootstrapping.}
\PrintBackRefs{\CurrentBib}

\bibitem [\protect \citeauthoryear {%
Rodr{\'\i}guez-Moreno%
, Mart{\'\i}nez-Otzeta%
, Sierra%
, Rodriguez%
\BCBL {}\ \BBA {} Jauregi%
}{%
Rodr{\'\i}guez-Moreno%
\ \protect \BOthers {.}}{%
{\protect \APACyear {2019}}%
}]{%
rodriguez2019video}
\APACinsertmetastar {%
rodriguez2019video}%
\begin{APACrefauthors}%
Rodr{\'\i}guez-Moreno, I.%
, Mart{\'\i}nez-Otzeta, J\BPBI M.%
, Sierra, B.%
, Rodriguez, I.%
\BCBL {}\ \BBA {} Jauregi, E.%
\end{APACrefauthors}%
\unskip\
\newblock
\APACrefYearMonthDay{2019}{}{}.
\newblock
{\BBOQ}\APACrefatitle {Video Activity Recognition: State-of-the-Art} {Video
  activity recognition: State-of-the-art}.{\BBCQ}
\newblock
\APACjournalVolNumPages{Sensors}{19}{14}{3160}.
\PrintBackRefs{\CurrentBib}

\bibitem [\protect \citeauthoryear {%
Schmidhuber%
}{%
Schmidhuber%
}{%
{\protect \APACyear {2015}}%
}]{%
schmidhuber2015deep}
\APACinsertmetastar {%
schmidhuber2015deep}%
\begin{APACrefauthors}%
Schmidhuber, J.%
\end{APACrefauthors}%
\unskip\
\newblock
\APACrefYearMonthDay{2015}{}{}.
\newblock
{\BBOQ}\APACrefatitle {Deep learning in neural networks: An overview} {Deep
  learning in neural networks: An overview}.{\BBCQ}
\newblock
\APACjournalVolNumPages{Neural networks}{61}{}{85--117}.
\PrintBackRefs{\CurrentBib}

\bibitem [\protect \citeauthoryear {%
Simistira~Liwicki%
, Liwicki%
, Perise%
, Visi%
\BCBL {}\ \BBA {} {\"O}stersj{\"o}%
}{%
Simistira~Liwicki%
\ \protect \BOthers {.}}{%
{\protect \APACyear {2020}}%
}]{%
anonym}
\APACinsertmetastar {%
anonym}%
\begin{APACrefauthors}%
Simistira~Liwicki, F.%
, Liwicki, M.%
, Perise, P\BPBI M.%
, Visi, F.%
\BCBL {}\ \BBA {} {\"O}stersj{\"o}, S.%
\end{APACrefauthors}%
\unskip\
\newblock
\APACrefYearMonthDay{2020}{}{}.
\newblock
{\BBOQ}\APACrefatitle {Analysing Musical Performance in Videos Using Deep
  Neural Networks} {Analysing musical performance in videos using deep neural
  networks}.{\BBCQ}
\newblock
\BIn{} \APACrefbtitle {1st Joint Conference on AI Music Creativity, AIMC,
  Online, 2020, October 19-23.} {1st joint conference on ai music creativity,
  aimc, online, 2020, october 19-23.}
\PrintBackRefs{\CurrentBib}

\bibitem [\protect \citeauthoryear {%
Simonyan%
\ \BBA {} Zisserman%
}{%
Simonyan%
\ \BBA {} Zisserman%
}{%
{\protect \APACyear {2014}}%
}]{%
simonyan2014two}
\APACinsertmetastar {%
simonyan2014two}%
\begin{APACrefauthors}%
Simonyan, K.%
\BCBT {}\ \BBA {} Zisserman, A.%
\end{APACrefauthors}%
\unskip\
\newblock
\APACrefYearMonthDay{2014}{}{}.
\newblock
{\BBOQ}\APACrefatitle {Two-stream convolutional networks for action recognition
  in videos} {Two-stream convolutional networks for action recognition in
  videos}.{\BBCQ}
\newblock
\BIn{} \APACrefbtitle {Advances in neural information processing systems}
  {Advances in neural information processing systems}\ (\BPGS\ 568--576).
\PrintBackRefs{\CurrentBib}

\bibitem [\protect \citeauthoryear {%
Soomro%
, Zamir%
\BCBL {}\ \BBA {} Shah%
}{%
Soomro%
\ \protect \BOthers {.}}{%
{\protect \APACyear {2012}}%
}]{%
soomro2012ucf101}
\APACinsertmetastar {%
soomro2012ucf101}%
\begin{APACrefauthors}%
Soomro, K.%
, Zamir, A\BPBI R.%
\BCBL {}\ \BBA {} Shah, M.%
\end{APACrefauthors}%
\unskip\
\newblock
\APACrefYearMonthDay{2012}{}{}.
\newblock
{\BBOQ}\APACrefatitle {UCF101: A dataset of 101 human actions classes from
  videos in the wild} {Ucf101: A dataset of 101 human actions classes from
  videos in the wild}.{\BBCQ}
\newblock
\APACjournalVolNumPages{arXiv preprint arXiv:1212.0402}{}{}{}.
\PrintBackRefs{\CurrentBib}

\bibitem [\protect \citeauthoryear {%
Varol%
, Laptev%
\BCBL {}\ \BBA {} Schmid%
}{%
Varol%
\ \protect \BOthers {.}}{%
{\protect \APACyear {2017}}%
}]{%
varol2017long}
\APACinsertmetastar {%
varol2017long}%
\begin{APACrefauthors}%
Varol, G.%
, Laptev, I.%
\BCBL {}\ \BBA {} Schmid, C.%
\end{APACrefauthors}%
\unskip\
\newblock
\APACrefYearMonthDay{2017}{}{}.
\newblock
{\BBOQ}\APACrefatitle {Long-term temporal convolutions for action recognition}
  {Long-term temporal convolutions for action recognition}.{\BBCQ}
\newblock
\APACjournalVolNumPages{IEEE transactions on pattern analysis and machine
  intelligence}{40}{6}{1510--1517}.
\PrintBackRefs{\CurrentBib}

\bibitem [\protect \citeauthoryear {%
F.~Visi%
, Schramm%
, Coorevits%
\BCBL {}\ \BBA {} Miranda%
}{%
F.~Visi%
\ \protect \BOthers {.}}{%
{\protect \APACyear {2017}}%
}]{%
visi2017musical}
\APACinsertmetastar {%
visi2017musical}%
\begin{APACrefauthors}%
Visi, F.%
, Schramm, R.%
, Coorevits, E.%
\BCBL {}\ \BBA {} Miranda, E\BPBI R.%
\end{APACrefauthors}%
\unskip\
\newblock
\APACrefYearMonthDay{2017}{}{}.
\newblock
{\BBOQ}\APACrefatitle {Musical instruments, body movement, space, and motion
  data: Music as an emergent multimodal choreography.} {Musical instruments,
  body movement, space, and motion data: Music as an emergent multimodal
  choreography.}{\BBCQ}
\newblock
\APACjournalVolNumPages{Human Technology}{13}{1}{}.
\PrintBackRefs{\CurrentBib}

\bibitem [\protect \citeauthoryear {%
F\BPBI G.~Visi%
\ \BBA {} Tanaka%
}{%
F\BPBI G.~Visi%
\ \BBA {} Tanaka%
}{%
{\protect \APACyear {2020}}%
}]{%
Visi2020b}
\APACinsertmetastar {%
Visi2020b}%
\begin{APACrefauthors}%
Visi, F\BPBI G.%
\BCBT {}\ \BBA {} Tanaka, A.%
\end{APACrefauthors}%
\unskip\
\newblock
\APACrefYearMonthDay{2020}{}{}.
\newblock
{\BBOQ}\APACrefatitle {{Interactive Machine Learning of Musical Gesture}}
  {{Interactive Machine Learning of Musical Gesture}}.{\BBCQ}
\newblock
\BIn{} E\BPBI R.~Miranda\ (\BED), \APACrefbtitle {Handbook of Music and AI.}
  {Handbook of music and ai.}
\newblock
\APACaddressPublisher{}{Springer, in press}.
\PrintBackRefs{\CurrentBib}

\end{thebibliography}
\newpage
\section*{Appendix}
\label{appendix}
\setcounter{table}{0}
\renewcommand{\thetable}{A\arabic{table}}

\let\cleardoublepage\clearpage 

% \begin{figure}[H]
% \centering
% \includegraphics[scale = 0.5]{3r_intercorr_superclass}
% \caption{Inter-correlation of super-classes for video 2}
% \label{fig:3r_intercorr_superclass}
% \end{figure}

% \begin{figure}[H]
% \centering
% \includegraphics[scale = 0.5]{4c_intercorr_superclass}
% \caption{Inter-correlation of super-classes for video 3}
% \label{fig:4c_intercorr_superclass}
% \end{figure}

% \begin{figure}[H]
% \centering
% \includegraphics[scale = 0.5]{4r_intercorr_superclass}
% \caption{Inter-correlation of super-classes for video 4}
% \label{fig:4r_intercorr_superclass}
% \end{figure}

%\begin{comment}

\begin{figure}[!ht]
\centering
\includegraphics[width=0.6\linewidth, height=0.6\linewidth]{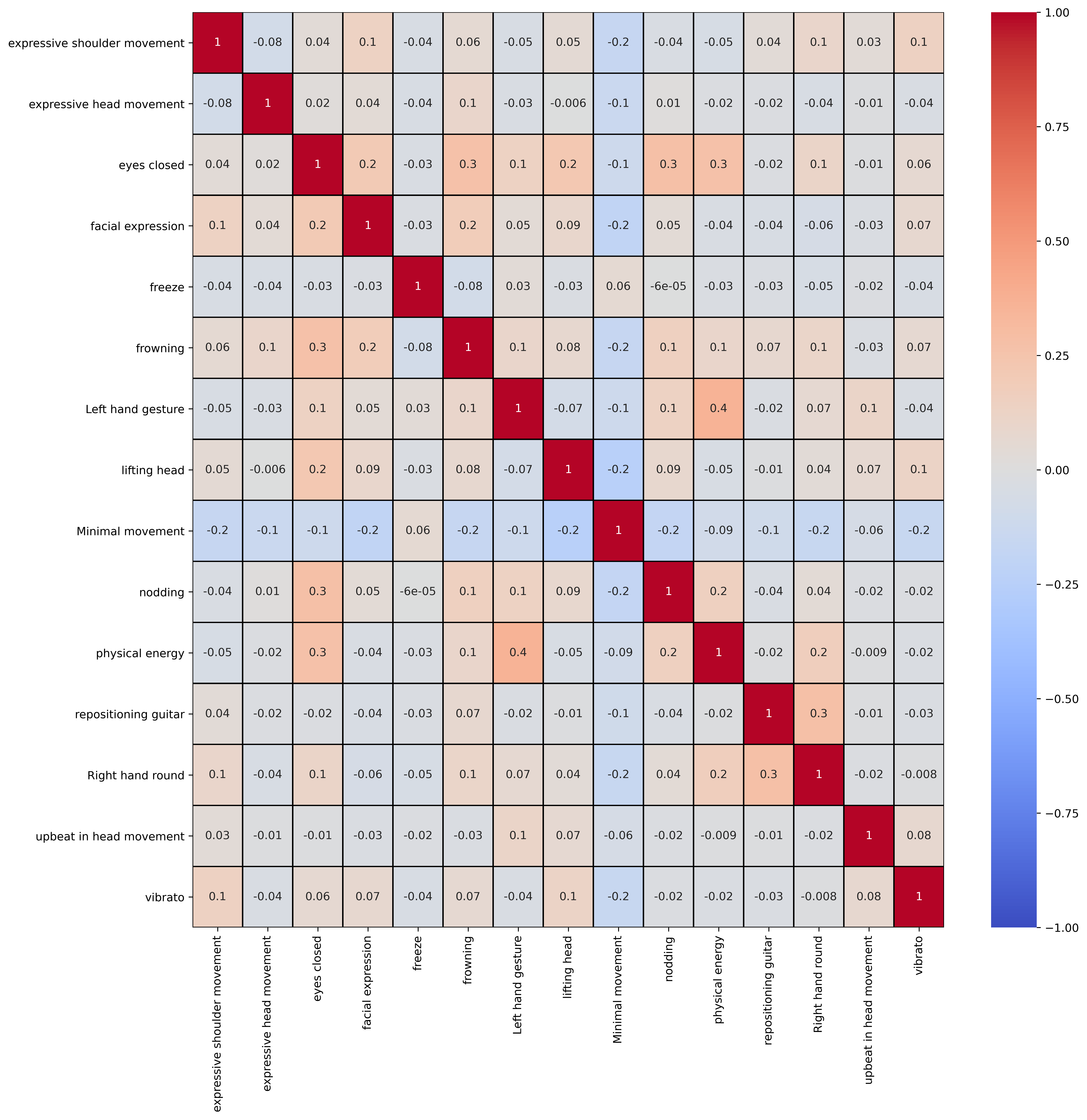}
\caption{Inter-correlation of expressive gestures for video-2}
\label{fig:3r_intercorr}
\end{figure}

\begin{figure}[!ht]
\centering
\includegraphics[width=0.6\linewidth, height=0.6\linewidth]{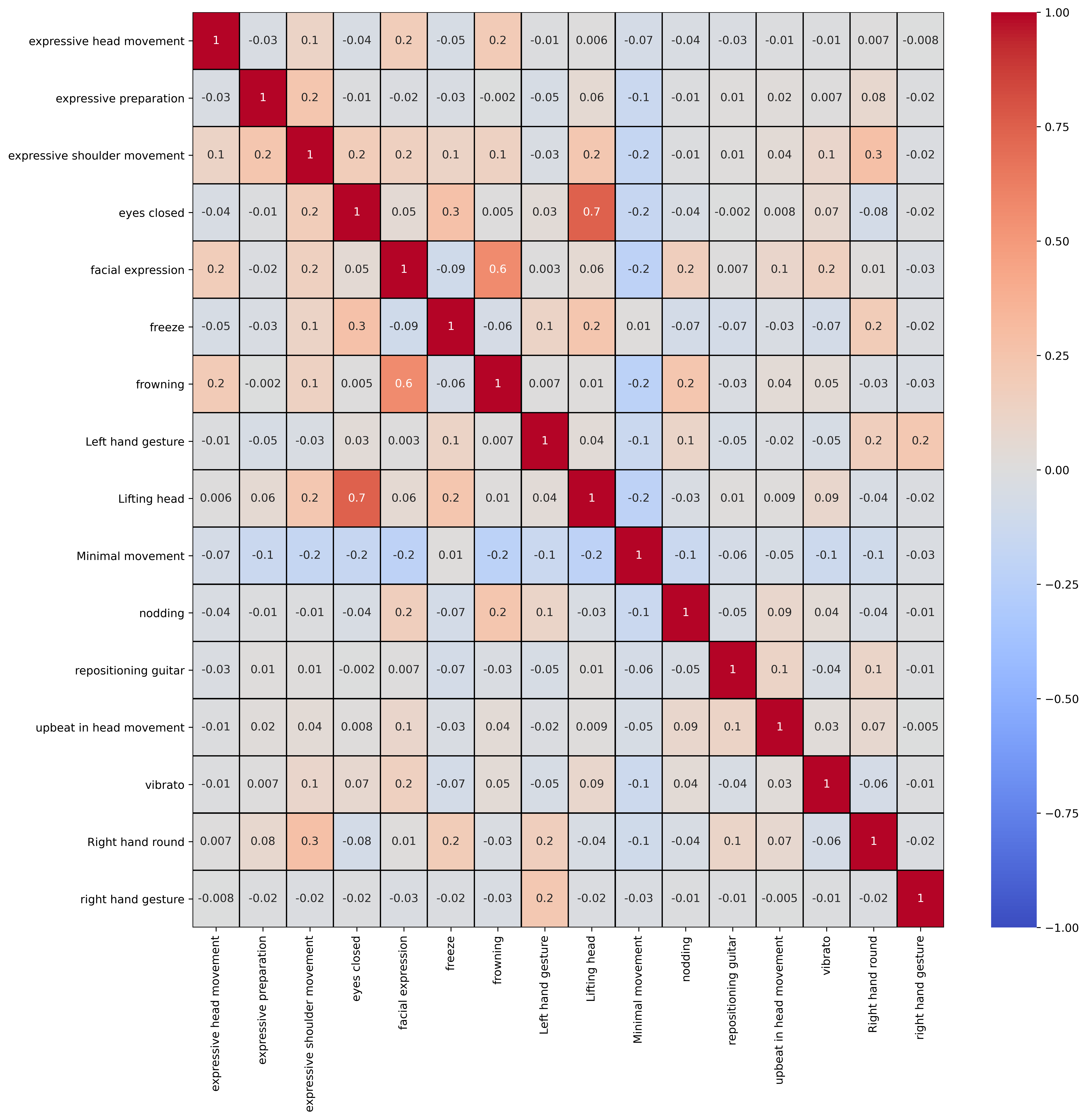}
\caption{Inter-correlation of expressive gestures for video-3}
\label{fig:4c_intercorr}
\end{figure}

%\end{comment}

\begin{comment}
\begin{figure}
\begin{minipage}{\linewidth}
  \centering
  \includegraphics[scale = 0.41]{3r_intercorr}
  \caption{\centering{Inter-correlation of expressive gestures for video 2}}
  \label{fig:3r_intercorr}
\end{minipage}
\vspace{1em} % add some whitespace after the first figure

\begin{minipage}[b]{0.45\linewidth}
  \centering
  \includegraphics[scale = 0.23]{4c_intercorr}
    \caption{\centering{Inter-correlation of expressive gestures for video 3}}
  \label{fig:4c_intercorr}
\end{minipage}
\hfill
\begin{minipage}[b]{0.45\linewidth}
  \centering
  \includegraphics[scale = 0.41]{4r_intercorr}
  \caption{\centering{Inter-correlation of expressive gestures for video 4}}
  \label{fig:4r_intercorr}
\end{minipage}
\end{figure}
\end{comment}

% \usepackage{lscape}
\begin{landscape}
\begin{table}[!ht]
\centering
\caption{Evaluation results for video - 1, 18 gestures}
\label{tab:3TC_18classes}
\begin{tabular}{|l|cccc|cccc|cccc|c|c|c|c|}
\hline
\multicolumn{1}{|r|}{\textbf{Experiments}} & \multicolumn{4}{c|}{\textbf{SM}} & \multicolumn{4}{c|}{\textbf{SM + BB}} & \multicolumn{4}{c|}{\textbf{SM + BB + TS}} & \multicolumn{4}{c|}{\textbf{Bi-modal + SM +}} \\
 & \multicolumn{1}{l}{} & \multicolumn{1}{l}{} & \multicolumn{1}{l}{} & \multicolumn{1}{l|}{} & \multicolumn{1}{l}{} & \multicolumn{1}{l}{} & \multicolumn{1}{l}{} & \multicolumn{1}{l|}{} & \multicolumn{1}{l}{} & \multicolumn{1}{l}{} & \multicolumn{1}{l}{} & \multicolumn{1}{l|}{} & \multicolumn{4}{c|}{\textbf{BB + TS}} \\ \hline
\multicolumn{1}{|r|}{\textbf{Evaluation Metric}} & \multicolumn{2}{c|}{\textbf{Precision}} & \multicolumn{2}{c|}{\textbf{Recall}} & \multicolumn{2}{c|}{\textbf{Precision}} & \multicolumn{2}{c|}{\textbf{Recall}} & \multicolumn{2}{c|}{\textbf{Precision}} & \multicolumn{2}{c|}{\textbf{Recall}} & \multicolumn{2}{c|}{\textbf{Precision}} & \multicolumn{2}{c|}{\textbf{Recall}} \\ \hline
\multicolumn{1}{|r|}{\textbf{Stage}} & \multicolumn{1}{c|}{\textbf{Train}} & \multicolumn{1}{c|}{\textbf{Test}} & \multicolumn{1}{c|}{\textbf{Train}} & \textbf{Test} & \multicolumn{1}{c|}{\textbf{Train}} & \multicolumn{1}{c|}{\textbf{Test}} & \multicolumn{1}{c|}{\textbf{Train}} & \textbf{Test} & \multicolumn{1}{c|}{\textbf{Train}} & \multicolumn{1}{c|}{\textbf{Test}} & \multicolumn{1}{c|}{\textbf{Train}} & \textbf{Test} & \textbf{Train} & \textbf{Test} & \textbf{Train} & \textbf{Test} \\ \hline
\textbf{Gestures} & \multicolumn{1}{c|}{} & \multicolumn{1}{c|}{} & \multicolumn{1}{c|}{} &  & \multicolumn{1}{c|}{} & \multicolumn{1}{c|}{} & \multicolumn{1}{c|}{} &  & \multicolumn{1}{c|}{} & \multicolumn{1}{c|}{} & \multicolumn{1}{c|}{} &  &  &  &  &  \\ \hline
{\color[HTML]{404040} \textbf{Facial expression}} & \multicolumn{1}{c|}{1.00} & \multicolumn{1}{c|}{0.88} & \multicolumn{1}{c|}{1.00} & 0.82 & \multicolumn{1}{c|}{0.80} & \multicolumn{1}{c|}{0.88} & \multicolumn{1}{c|}{0.98} & 0.82 & \multicolumn{1}{c|}{0.84} & \multicolumn{1}{c|}{0.88} & \multicolumn{1}{c|}{0.94} & 0.82 & 0.76 & 1.00 & 0.91 & 0.76 \\ \hline
{\color[HTML]{404040} \textbf{Nodding}} & \multicolumn{1}{c|}{0.95} & \multicolumn{1}{c|}{0.94} & \multicolumn{1}{c|}{0.90} & 0.94 & \multicolumn{1}{c|}{0.94} & \multicolumn{1}{c|}{1.00} & \multicolumn{1}{c|}{0.99} & 0.94 & \multicolumn{1}{c|}{0.88} & \multicolumn{1}{c|}{0.86} & \multicolumn{1}{c|}{0.85} & 0.96 & 0.80 & 0.92 & 0.85 & 0.92 \\ \hline
{\color[HTML]{404040} \textbf{Right hand round}} & \multicolumn{1}{c|}{0.89} & \multicolumn{1}{c|}{1.00} & \multicolumn{1}{c|}{0.84} & 0.78 & \multicolumn{1}{c|}{0.90} & \multicolumn{1}{c|}{1.00} & \multicolumn{1}{c|}{0.96} & 0.67 & \multicolumn{1}{c|}{0.82} & \multicolumn{1}{c|}{1.00} & \multicolumn{1}{c|}{0.88} & 0.89 & 0.68 & 1.00 & 0.76 & 0.78 \\ \hline
{\color[HTML]{404040} \textbf{Expressive shoulder movement}} & \multicolumn{1}{c|}{1.00} & \multicolumn{1}{c|}{0.83} & \multicolumn{1}{c|}{1.00} & 0.71 & \multicolumn{1}{c|}{0.87} & \multicolumn{1}{c|}{1.00} & \multicolumn{1}{c|}{0.96} & 0.57 & \multicolumn{1}{c|}{0.91} & \multicolumn{1}{c|}{1.00} & \multicolumn{1}{c|}{0.96} & 0.63 & 0.81 & 0.86 & 0.95 & 0.75 \\ \hline
{\color[HTML]{404040} \textbf{Left hand gesture}} & \multicolumn{1}{c|}{0.95} & \multicolumn{1}{c|}{0.60} & \multicolumn{1}{c|}{0.95} & 1.00 & \multicolumn{1}{c|}{0.91} & \multicolumn{1}{c|}{1.00} & \multicolumn{1}{c|}{0.99} & 1.00 & \multicolumn{1}{c|}{0.89} & \multicolumn{1}{c|}{1.00} & \multicolumn{1}{c|}{0.93} & 1.00 & 0.91 & 0.71 & 0.98 & 1.00 \\ \hline
{\color[HTML]{404040} \textbf{Lifting head}} & \multicolumn{1}{c|}{1.00} & \multicolumn{1}{c|}{1.00} & \multicolumn{1}{c|}{0.98} & 0.50 & \multicolumn{1}{c|}{0.80} & \multicolumn{1}{c|}{0.50} & \multicolumn{1}{c|}{0.93} & 0.50 & \multicolumn{1}{c|}{0.84} & \multicolumn{1}{c|}{0.57} & \multicolumn{1}{c|}{0.94} & 0.80 & 0.87 & 1.00 & 0.97 & 0.60 \\ \hline
{\color[HTML]{404040} \textbf{Minimal movement}} & \multicolumn{1}{c|}{0.00} & \multicolumn{1}{c|}{0.00} & \multicolumn{1}{c|}{0.00} & 0.00 & \multicolumn{1}{c|}{0.00} & \multicolumn{1}{c|}{0.00} & \multicolumn{1}{c|}{0.00} & 0.00 & \multicolumn{1}{c|}{0.00} & \multicolumn{1}{c|}{0.00} & \multicolumn{1}{c|}{0.00} & 0.00 & 0.00 & 0.00 & 0.00 & 0.00 \\ \hline
{\color[HTML]{404040} \textbf{Eyes closed}} & \multicolumn{1}{c|}{0.89} & \multicolumn{1}{c|}{0.67} & \multicolumn{1}{c|}{0.89} & 1.00 & \multicolumn{1}{c|}{0.84} & \multicolumn{1}{c|}{1.00} & \multicolumn{1}{c|}{0.97} & 1.00 & \multicolumn{1}{c|}{0.99} & \multicolumn{1}{c|}{1.00} & \multicolumn{1}{c|}{1.00} & 1.00 & 0.69 & 1.00 & 0.86 & 1.00 \\ \hline
{\color[HTML]{404040} \textbf{Vibrato}} & \multicolumn{1}{c|}{0.68} & \multicolumn{1}{c|}{0.00} & \multicolumn{1}{c|}{0.66} & 0.00 & \multicolumn{1}{c|}{0.98} & \multicolumn{1}{c|}{0.00} & \multicolumn{1}{c|}{0.97} & 0.00 & \multicolumn{1}{c|}{0.62} & \multicolumn{1}{c|}{0.00} & \multicolumn{1}{c|}{0.67} & 0.00 & 0.62 & 0.00 & 0.72 & 0.00 \\ \hline
{\color[HTML]{404040} \textbf{Expressive preparation}} & \multicolumn{1}{c|}{0.93} & \multicolumn{1}{c|}{0.00} & \multicolumn{1}{c|}{0.91} & 0.00 & \multicolumn{1}{c|}{0.87} & \multicolumn{1}{c|}{0.00} & \multicolumn{1}{c|}{0.96} & 0.00 & \multicolumn{1}{c|}{0.84} & \multicolumn{1}{c|}{1.00} & \multicolumn{1}{c|}{0.98} & 0.50 & 0.79 & 1.00 & 0.96 & 0.50 \\ \hline
{\color[HTML]{404040} \textbf{Freeze}} & \multicolumn{1}{c|}{0.89} & \multicolumn{1}{c|}{0.00} & \multicolumn{1}{c|}{0.88} & 0.00 & \multicolumn{1}{c|}{0.93} & \multicolumn{1}{c|}{0.00} & \multicolumn{1}{c|}{1.00} & 0.00 & \multicolumn{1}{c|}{0.85} & \multicolumn{1}{c|}{0.00} & \multicolumn{1}{c|}{0.88} & 0.00 & 0.69 & 0.00 & 0.83 & 0.00 \\ \hline
{\color[HTML]{404040} \textbf{Expressive head movement}} & \multicolumn{1}{c|}{0.37} & \multicolumn{1}{c|}{0.00} & \multicolumn{1}{c|}{0.35} & 0.00 & \multicolumn{1}{c|}{0.64} & \multicolumn{1}{c|}{0.00} & \multicolumn{1}{c|}{0.75} & 0.00 & \multicolumn{1}{c|}{0.47} & \multicolumn{1}{c|}{0.00} & \multicolumn{1}{c|}{0.47} & 0.00 & 0.51 & 0.00 & 0.53 & 0.00 \\ \hline
{\color[HTML]{404040} \textbf{Frowning}} & \multicolumn{1}{c|}{0.99} & \multicolumn{1}{c|}{0.92} & \multicolumn{1}{c|}{1.00} & 0.92 & \multicolumn{1}{c|}{0.80} & \multicolumn{1}{c|}{0.75} & \multicolumn{1}{c|}{0.98} & 0.92 & \multicolumn{1}{c|}{0.84} & \multicolumn{1}{c|}{0.68} & \multicolumn{1}{c|}{0.96} & 0.93 & 0.76 & 0.68 & 0.94 & 0.93 \\ \hline
{\color[HTML]{404040} \textbf{Physical energy}} & \multicolumn{1}{c|}{1.00} & \multicolumn{1}{c|}{0.88} & \multicolumn{1}{c|}{1.00} & 0.88 & \multicolumn{1}{c|}{0.78} & \multicolumn{1}{c|}{1.00} & \multicolumn{1}{c|}{0.88} & 0.75 & \multicolumn{1}{c|}{0.86} & \multicolumn{1}{c|}{1.00} & \multicolumn{1}{c|}{0.94} & 0.89 & 0.87 & 1.00 & 0.95 & 0.78 \\ \hline
{\color[HTML]{404040} \textbf{Upbeat in head movement}} & \multicolumn{1}{c|}{0.00} & \multicolumn{1}{c|}{0.00} & \multicolumn{1}{c|}{0.00} & 0.00 & \multicolumn{1}{c|}{0.00} & \multicolumn{1}{c|}{0.00} & \multicolumn{1}{c|}{0.00} & 0.00 & \multicolumn{1}{c|}{0.00} & \multicolumn{1}{c|}{0.00} & \multicolumn{1}{c|}{0.00} & 0.00 & 0.00 & 0.00 & 0.00 & 0.00 \\ \hline
{\color[HTML]{404040} \textbf{repositioning guitar}} & \multicolumn{1}{c|}{0.00} & \multicolumn{1}{c|}{0.00} & \multicolumn{1}{c|}{0.00} & 0.00 & \multicolumn{1}{c|}{0.00} & \multicolumn{1}{c|}{0.00} & \multicolumn{1}{c|}{0.00} & 0.00 & \multicolumn{1}{c|}{0.00} & \multicolumn{1}{c|}{0.00} & \multicolumn{1}{c|}{0.00} & 0.00 & 0.00 & 0.00 & 0.00 & 0.00 \\ \hline
{\color[HTML]{404040} \textbf{sympathetic body movement}} & \multicolumn{1}{c|}{0.00} & \multicolumn{1}{c|}{0.00} & \multicolumn{1}{c|}{0.00} & 0.00 & \multicolumn{1}{c|}{0.00} & \multicolumn{1}{c|}{0.00} & \multicolumn{1}{c|}{0.00} & 0.00 & \multicolumn{1}{c|}{0.00} & \multicolumn{1}{c|}{0.00} & \multicolumn{1}{c|}{0.00} & 0.00 & 0.00 & 0.00 & 0.00 & 0.00 \\ \hline
{\color[HTML]{404040} \textbf{normal play}} & \multicolumn{1}{c|}{1.00} & \multicolumn{1}{c|}{0.91} & \multicolumn{1}{c|}{1.00} & 1.00 & \multicolumn{1}{c|}{0.98} & \multicolumn{1}{c|}{0.84} & \multicolumn{1}{c|}{0.98} & 0.97 & \multicolumn{1}{c|}{0.98} & \multicolumn{1}{c|}{1.00} & \multicolumn{1}{c|}{0.96} & 0.90 & 0.95 & 0.86 & 0.95 & 0.90 \\ \hline
\end{tabular}
\end{table}
\end{landscape}

\begin{sidewaystable}
% \usepackage{lscape}
%\begin{landscape}
%\begin{table}[ht!]
%\centering
%\begin{table}[ht!]
%\begin{sideways}
%\begin{minipage}[h!]{2\linewidth}
%\centering

%\begin{sidewaystable}[ht!]
\caption{Evaluation result of video - 1, for 7 super-classes}
\label{tab:3TC_7classes}
\begin{adjustbox}{scale=1, valign=t, raise={8cm}{8cm}{0pt}}
\begin{tabular}{|l|cccc|cccc|cccc|c|c|c|c|}
\hline
\multicolumn{1}{|r|}{\textbf{Experiments}} & \multicolumn{4}{c|}{\textbf{SM}} & \multicolumn{4}{c|}{\textbf{SM + BB}} & \multicolumn{4}{c|}{\textbf{SM + BB + TS}} & \multicolumn{4}{c|}{\textbf{Bi-modal + SM}} \\
 & \multicolumn{1}{l}{} & \multicolumn{1}{l}{} & \multicolumn{1}{l}{} & \multicolumn{1}{l|}{} & \multicolumn{1}{l}{} & \multicolumn{1}{l}{} & \multicolumn{1}{l}{} & \multicolumn{1}{l|}{} & \multicolumn{1}{l}{} & \multicolumn{1}{l}{} & \multicolumn{1}{l}{} & \multicolumn{1}{l|}{} & \multicolumn{4}{c|}{\textbf{BB + TS}} \\ \hline
\textbf{Evaluation Metric} & \multicolumn{2}{c|}{\textbf{Precision}} & \multicolumn{2}{c|}{\textbf{Recall}} & \multicolumn{2}{c|}{\textbf{Precision}} & \multicolumn{2}{c|}{\textbf{Recall}} & \multicolumn{2}{c|}{\textbf{Precision}} & \multicolumn{2}{c|}{\textbf{Recall}} & \multicolumn{2}{c|}{\textbf{Precision}} & \multicolumn{2}{c|}{\textbf{Recall}} \\ \hline
\textbf{Stage} & \multicolumn{1}{c|}{\textbf{Train}} & \multicolumn{1}{c|}{\textbf{Test}} & \multicolumn{1}{c|}{\textbf{Train}} & \textbf{Test} & \multicolumn{1}{c|}{\textbf{Train}} & \multicolumn{1}{c|}{\textbf{Test}} & \multicolumn{1}{c|}{\textbf{Train}} & \textbf{Test} & \multicolumn{1}{c|}{\textbf{Train}} & \multicolumn{1}{c|}{\textbf{Test}} & \multicolumn{1}{c|}{\textbf{Train}} & \textbf{Test} & \textbf{Train} & \textbf{Test} & \textbf{Train} & \textbf{Test} \\ \hline
{\color[HTML]{404040} \textbf{Facial   Expression}} & \multicolumn{1}{c|}{1.00} & \multicolumn{1}{c|}{0.84} & \multicolumn{1}{c|}{1.00} & 0.89 & \multicolumn{1}{c|}{0.89} & \multicolumn{1}{c|}{0.75} & \multicolumn{1}{c|}{0.95} & 0.83 & \multicolumn{1}{c|}{0.98} & \multicolumn{1}{c|}{0.95} & \multicolumn{1}{c|}{0.96} & 0.94 & 0.72 & 0.46 & 0.96 & 0.95 \\ \hline
{\color[HTML]{404040} \textbf{Head related   action}} & \multicolumn{1}{c|}{1.00} & \multicolumn{1}{c|}{0.90} & \multicolumn{1}{c|}{0.99} & 0.86 & \multicolumn{1}{c|}{0.89} & \multicolumn{1}{c|}{0.81} & \multicolumn{1}{c|}{1.00} & 0.81 & \multicolumn{1}{c|}{0.97} & \multicolumn{1}{c|}{0.91} & \multicolumn{1}{c|}{0.95} & 0.96 & 0.50 & 0.78 & 0.93 & 0.93 \\ \hline
{\color[HTML]{404040} \textbf{Right hand   action}} & \multicolumn{1}{c|}{0.98} & \multicolumn{1}{c|}{1.00} & \multicolumn{1}{c|}{0.99} & 0.78 & \multicolumn{1}{c|}{0.79} & \multicolumn{1}{c|}{1.00} & \multicolumn{1}{c|}{0.95} & 1.00 & \multicolumn{1}{c|}{0.91} & \multicolumn{1}{c|}{0.85} & \multicolumn{1}{c|}{0.91} & 0.86 & 0.85 & 1.00 & 0.91 & 1.00 \\ \hline
{\color[HTML]{404040} \textbf{Left hand   action}} & \multicolumn{1}{c|}{0.90} & \multicolumn{1}{c|}{1.00} & \multicolumn{1}{c|}{1.00} & 1.00 & \multicolumn{1}{c|}{0.97} & \multicolumn{1}{c|}{0.67} & \multicolumn{1}{c|}{1.00} & 1.00 & \multicolumn{1}{c|}{0.97} & \multicolumn{1}{c|}{0.85} & \multicolumn{1}{c|}{0.99} & 0.80 & 0.90 & 1.00 & 0.91 & 1.00 \\ \hline
{\color[HTML]{404040} \textbf{Relative   stillness}} & \multicolumn{1}{c|}{0.81} & \multicolumn{1}{c|}{0.00} & \multicolumn{1}{c|}{0.82} & 0.00 & \multicolumn{1}{c|}{0.74} & \multicolumn{1}{c|}{0.00} & \multicolumn{1}{c|}{0.84} & 0.00 & \multicolumn{1}{c|}{0.95} & \multicolumn{1}{c|}{0.64} & \multicolumn{1}{c|}{0.95} & 0.66 & 0.74 & 0.00 & 0.83 & 0.00 \\ \hline
{\color[HTML]{404040} \textbf{Upper body   movement}} & \multicolumn{1}{c|}{1.00} & \multicolumn{1}{c|}{0.88} & \multicolumn{1}{c|}{1.00} & 0.88 & \multicolumn{1}{c|}{0.99} & \multicolumn{1}{c|}{1.00} & \multicolumn{1}{c|}{0.99} & 0.75 & \multicolumn{1}{c|}{0.98} & \multicolumn{1}{c|}{0.72} & \multicolumn{1}{c|}{0.96} & 0.72 & 0.73 & 0.55 & 0.98 & 0.75 \\ \hline
{\color[HTML]{404040} \textbf{Normal play}} & \multicolumn{1}{c|}{0.99} & \multicolumn{1}{c|}{0.91} & \multicolumn{1}{c|}{1.00} & 1.00 & \multicolumn{1}{c|}{0.99} & \multicolumn{1}{c|}{0.90} & \multicolumn{1}{c|}{0.99} & 0.93 & \multicolumn{1}{c|}{0.99} & \multicolumn{1}{c|}{0.93} & \multicolumn{1}{c|}{0.98} & 0.96 & 0.85 & 0.93 & 0.96 & 0.93 \\ \hline
\end{tabular}
%\end{table}
%\end{landscape}
%\end{minipage}
%\end{sideways}
%\end{table}
\end{adjustbox}
%\end{sidewaystable}
\medskip
% \usepackage{lscape}
%\begin{landscape}
%\begin{table}[ht!]
%\centering
%\begin{sidewaystable}[ht!]
\caption{Evaluation result of all videos (video-1 + video-2 + video-3 + video-4), for 7 super-classes}
\label{tab:4videos_7classes}

\begin{adjustbox}{scale=1,valign=t, raise={8cm}{8cm}{0pt}}
\begin{tabular}{|l|cccc|cccc|cccc|c|c|c|c}
\hline
\multicolumn{1}{|r|}{\textbf{Experiments}} & \multicolumn{4}{c|}{\textbf{SM}} & \multicolumn{4}{c|}{\textbf{SM + BB}} & \multicolumn{4}{c|}{\textbf{SM + BB + TS}} & \multicolumn{4}{c|}{\textbf{Bi-modal  + SM +}} \\
 & \multicolumn{1}{l}{} & \multicolumn{1}{l}{} & \multicolumn{1}{l}{} & \multicolumn{1}{l|}{} & \multicolumn{1}{l}{} & \multicolumn{1}{l}{} & \multicolumn{1}{l}{} & \multicolumn{1}{l|}{} & \multicolumn{1}{l}{} & \multicolumn{1}{l}{} & \multicolumn{1}{l}{} & \multicolumn{1}{l|}{} & \multicolumn{4}{c|}{\textbf{BB + TS}} \\ \hline
\textbf{Evaluation Metric} & \multicolumn{2}{c|}{\textbf{Precision}} & \multicolumn{2}{c|}{\textbf{Recall}} & \multicolumn{2}{c|}{\textbf{Precision}} & \multicolumn{2}{c|}{\textbf{Recall}} & \multicolumn{2}{c|}{\textbf{Precision}} & \multicolumn{2}{c|}{\textbf{Recall}} & \multicolumn{2}{c|}{\textbf{Precision}} & \multicolumn{2}{c|}{\textbf{Recall}} \\ \hline
\textbf{Stage} & \multicolumn{1}{c|}{\textbf{Train}} & \multicolumn{1}{c|}{\textbf{Test}} & \multicolumn{1}{c|}{\textbf{Train}} & \textbf{Test} & \multicolumn{1}{c|}{\textbf{Train}} & \multicolumn{1}{c|}{\textbf{Test}} & \multicolumn{1}{c|}{\textbf{Train}} & \textbf{Test} & \multicolumn{1}{c|}{\textbf{Train}} & \multicolumn{1}{c|}{\textbf{Test}} & \multicolumn{1}{c|}{\textbf{Train}} & \textbf{Test} & \textbf{Train} & \textbf{Test} & \textbf{Train} & \multicolumn{1}{c|}{\textbf{Test}} \\ \hline
{\color[HTML]{404040} \textbf{Facial   Expression}} & \multicolumn{1}{c|}{0.97} & \multicolumn{1}{c|}{0.88} & \multicolumn{1}{c|}{0.97} & 0.90 & \multicolumn{1}{c|}{0.80} & \multicolumn{1}{c|}{0.83} & \multicolumn{1}{c|}{0.98} & 0.90 & \multicolumn{1}{c|}{0.98} & \multicolumn{1}{c|}{0.95} & \multicolumn{1}{c|}{0.95} & 0.86 & 0.56 & 0.62 & 0.90 & \multicolumn{1}{c|}{0.95} \\ \hline
{\color[HTML]{404040} \textbf{Head related   action}} & \multicolumn{1}{c|}{0.95} & \multicolumn{1}{c|}{0.85} & \multicolumn{1}{c|}{0.92} & 0.76 & \multicolumn{1}{c|}{0.85} & \multicolumn{1}{c|}{0.75} & \multicolumn{1}{c|}{0.96} & 0.89 & \multicolumn{1}{c|}{0.95} & \multicolumn{1}{c|}{0.93} & \multicolumn{1}{c|}{0.96} & 0.90 & 0.53 & 0.59 & 0.92 & \multicolumn{1}{c|}{0.90} \\ \hline
{\color[HTML]{404040} \textbf{Right hand   action}} & \multicolumn{1}{c|}{0.85} & \multicolumn{1}{c|}{0.89} & \multicolumn{1}{c|}{0.79} & 0.80 & \multicolumn{1}{c|}{0.80} & \multicolumn{1}{c|}{0.70} & \multicolumn{1}{c|}{0.93} & 0.95 & \multicolumn{1}{c|}{0.88} & \multicolumn{1}{c|}{0.77} & \multicolumn{1}{c|}{0.89} & 0.95 & 0.77 & 0.90 & 0.87 & \multicolumn{1}{c|}{0.86} \\ \hline
{\color[HTML]{404040} \textbf{Left hand   action}} & \multicolumn{1}{c|}{0.93} & \multicolumn{1}{c|}{0.76} & \multicolumn{1}{c|}{0.84} & 0.81 & \multicolumn{1}{c|}{0.85} & \multicolumn{1}{c|}{0.95} & \multicolumn{1}{c|}{0.95} & 0.86 & \multicolumn{1}{c|}{0.94} & \multicolumn{1}{c|}{0.86} & \multicolumn{1}{c|}{0.94} & 0.93 & 0.77 & 0.79 & 0.82 & \multicolumn{1}{c|}{0.69} \\ \hline
{\color[HTML]{404040} \textbf{Relative   stillness}} & \multicolumn{1}{c|}{0.95} & \multicolumn{1}{c|}{0.63} & \multicolumn{1}{c|}{0.98} & 1.00 & \multicolumn{1}{c|}{0.90} & \multicolumn{1}{c|}{0.83} & \multicolumn{1}{c|}{0.98} & 1.00 & \multicolumn{1}{c|}{0.99} & \multicolumn{1}{c|}{1.00} & \multicolumn{1}{c|}{0.96} & 1.00 & 0.74 & 0.36 & 0.93 & \multicolumn{1}{c|}{0.71} \\ \hline
{\color[HTML]{404040} \textbf{Upper body   movement}} & \multicolumn{1}{c|}{0.94} & \multicolumn{1}{c|}{0.85} & \multicolumn{1}{c|}{0.94} & 0.95 & \multicolumn{1}{c|}{0.80} & \multicolumn{1}{c|}{0.59} & \multicolumn{1}{c|}{0.96} & 0.97 & \multicolumn{1}{c|}{0.96} & \multicolumn{1}{c|}{0.95} & \multicolumn{1}{c|}{0.95} & 0.90 & 0.73 & 0.70 & 0.91 & \multicolumn{1}{c|}{0.93} \\ \hline
{\color[HTML]{404040} \textbf{Normal play}} & \multicolumn{1}{c|}{0.95} & \multicolumn{1}{c|}{0.89} & \multicolumn{1}{c|}{0.99} & 0.93 & \multicolumn{1}{c|}{0.94} & \multicolumn{1}{c|}{0.92} & \multicolumn{1}{c|}{0.98} & 0.94 & \multicolumn{1}{c|}{0.96} & \multicolumn{1}{c|}{0.88} & \multicolumn{1}{c|}{0.96} & 0.96 & 0.74 & 0.83 & 0.93 & \multicolumn{1}{c|}{0.94} \\ \hline
\end{tabular}
\end{adjustbox}
%\end{table}
%\end{landscape}
%\end{sidewaystable}
\end{sidewaystable}

% \usepackage{lscape}
\begin{landscape}
\begin{table}[!ht]
\centering
\caption{Evaluation result of all videos (video-1 + video-2 + video-3 + video-4), for 18 super-classes}
\label{tab:allvideos18classes}
\begin{tabular}{|l|cccc|cccc|cccc|c|c|c|c|}
\hline
\multicolumn{1}{|r|}{\textbf{Experiments}} & \multicolumn{4}{c|}{\textbf{SM}} & \multicolumn{4}{c|}{\textbf{SM + BB}} & \multicolumn{4}{c|}{\textbf{Sm + BB + TS}} & \multicolumn{4}{c|}{\textbf{Bi-modal + SM +}} \\
 & \multicolumn{1}{l}{} & \multicolumn{1}{l}{} & \multicolumn{1}{l}{} & \multicolumn{1}{l|}{} & \multicolumn{1}{l}{} & \multicolumn{1}{l}{} & \multicolumn{1}{l}{} & \multicolumn{1}{l|}{} & \multicolumn{1}{l}{} & \multicolumn{1}{l}{} & \multicolumn{1}{l}{} & \multicolumn{1}{l|}{} & \multicolumn{4}{c|}{\textbf{BB + TS}} \\ \hline
\multicolumn{1}{|r|}{\textbf{Evaluation Metric}} & \multicolumn{2}{c|}{\textbf{Precision}} & \multicolumn{2}{c|}{\textbf{Recall}} & \multicolumn{2}{c|}{\textbf{Precision}} & \multicolumn{2}{c|}{\textbf{Recall}} & \multicolumn{2}{c|}{\textbf{Precision}} & \multicolumn{2}{c|}{\textbf{Recall}} & \multicolumn{2}{c|}{\textbf{Precision}} & \multicolumn{2}{c|}{\textbf{Recall}} \\ \hline
\multicolumn{1}{|r|}{\textbf{Stage}} & \multicolumn{1}{c|}{\textbf{Train}} & \multicolumn{1}{c|}{\textbf{Test}} & \multicolumn{1}{c|}{\textbf{Train}} & \textbf{Test} & \multicolumn{1}{c|}{\textbf{Train}} & \multicolumn{1}{c|}{\textbf{Test}} & \multicolumn{1}{c|}{\textbf{Train}} & \textbf{Test} & \multicolumn{1}{c|}{\textbf{Train}} & \multicolumn{1}{c|}{\textbf{Test}} & \multicolumn{1}{c|}{\textbf{Train}} & \textbf{Test} & \textbf{Train} & \textbf{Test} & \textbf{Train} & \textbf{Test} \\ \hline
\textbf{Gestures} & \multicolumn{1}{l|}{} & \multicolumn{1}{l|}{} & \multicolumn{1}{l|}{} & \multicolumn{1}{l|}{} & \multicolumn{1}{l|}{} & \multicolumn{1}{l|}{} & \multicolumn{1}{l|}{} & \multicolumn{1}{l|}{} & \multicolumn{1}{l|}{} & \multicolumn{1}{l|}{} & \multicolumn{1}{l|}{} & \multicolumn{1}{l|}{} & \multicolumn{1}{l|}{} & \multicolumn{1}{l|}{} & \multicolumn{1}{l|}{} & \multicolumn{1}{l|}{} \\ \hline
{\color[HTML]{404040} \textbf{Facial expression}} & \multicolumn{1}{c|}{0.82} & \multicolumn{1}{c|}{0.85} & \multicolumn{1}{c|}{0.86} & 0.92 & \multicolumn{1}{c|}{0.96} & \multicolumn{1}{c|}{0.93} & \multicolumn{1}{c|}{0.96} & 0.87 & \multicolumn{1}{c|}{0.90} & \multicolumn{1}{c|}{0.84} & \multicolumn{1}{c|}{0.93} & 0.88 & 0.40 & 0.62 & 0.89 & 0.84 \\ \hline
{\color[HTML]{404040} \textbf{Nodding}} & \multicolumn{1}{c|}{0.64} & \multicolumn{1}{c|}{{\color[HTML]{404040} 0.68}} & \multicolumn{1}{c|}{0.67} & 0.81 & \multicolumn{1}{c|}{{\color[HTML]{404040} 0.82}} & \multicolumn{1}{c|}{0.87} & \multicolumn{1}{c|}{0.81} & 0.73 & \multicolumn{1}{c|}{0.83} & \multicolumn{1}{c|}{0.81} & \multicolumn{1}{c|}{0.88} & 0.83 & 0.51 & 0.66 & 0.83 & 0.87 \\ \hline
{\color[HTML]{404040} \textbf{Right hand round}} & \multicolumn{1}{c|}{0.75} & \multicolumn{1}{c|}{{\color[HTML]{404040} 0.94}} & \multicolumn{1}{c|}{0.75} & 0.85 & \multicolumn{1}{c|}{{\color[HTML]{404040} 0.86}} & \multicolumn{1}{c|}{0.94} & \multicolumn{1}{c|}{0.85} & 0.85 & \multicolumn{1}{c|}{0.84} & \multicolumn{1}{c|}{0.86} & \multicolumn{1}{c|}{0.90} & 0.78 & 0.56 & 0.56 & 0.84 & 0.87 \\ \hline
{\color[HTML]{404040} \textbf{Expressive shoulder movement}} & \multicolumn{1}{c|}{0.96} & \multicolumn{1}{c|}{{\color[HTML]{404040} 0.97}} & \multicolumn{1}{c|}{0.93} & 0.84 & \multicolumn{1}{c|}{{\color[HTML]{404040} 0.97}} & \multicolumn{1}{c|}{0.87} & \multicolumn{1}{c|}{0.98} & 0.73 & \multicolumn{1}{c|}{0.91} & \multicolumn{1}{c|}{0.81} & \multicolumn{1}{c|}{0.97} & 0.98 & 0.44 & 0.37 & 0.92 & 0.95 \\ \hline
{\color[HTML]{404040} \textbf{Left hand gesture}} & \multicolumn{1}{c|}{0.75} & \multicolumn{1}{c|}{{\color[HTML]{404040} 0.89}} & \multicolumn{1}{c|}{0.68} & 0.89 & \multicolumn{1}{c|}{{\color[HTML]{404040} 0.81}} & \multicolumn{1}{c|}{0.93} & \multicolumn{1}{c|}{0.76} & 0.93 & \multicolumn{1}{c|}{0.83} & \multicolumn{1}{c|}{0.83} & \multicolumn{1}{c|}{0.87} & 0.97 & 0.72 & 0.59 & 0.86 & 0.87 \\ \hline
{\color[HTML]{404040} \textbf{Lifting head}} & \multicolumn{1}{c|}{0.91} & \multicolumn{1}{c|}{{\color[HTML]{404040} 0.76}} & \multicolumn{1}{c|}{0.89} & 0.80 & \multicolumn{1}{c|}{{\color[HTML]{404040} 0.95}} & \multicolumn{1}{c|}{0.94} & \multicolumn{1}{c|}{0.96} & 0.75 & \multicolumn{1}{c|}{0.93} & \multicolumn{1}{c|}{0.71} & \multicolumn{1}{c|}{0.95} & 0.83 & 0.50 & 0.49 & 0.91 & 0.83 \\ \hline
{\color[HTML]{404040} \textbf{Minimal movement}} & \multicolumn{1}{c|}{0.96} & \multicolumn{1}{c|}{{\color[HTML]{404040} \textbf{0.00}}} & \multicolumn{1}{c|}{0.97} & 0.00 & \multicolumn{1}{c|}{{\color[HTML]{404040} 0.96}} & \multicolumn{1}{c|}{0.00} & \multicolumn{1}{c|}{0.97} & 0.00 & \multicolumn{1}{c|}{0.98} & \multicolumn{1}{c|}{0.00} & \multicolumn{1}{c|}{0.98} & 0.00 & 0.68 & 0.00 & 0.95 & 0.00 \\ \hline
{\color[HTML]{404040} \textbf{Eyes closed}} & \multicolumn{1}{c|}{0.87} & \multicolumn{1}{c|}{0.80} & \multicolumn{1}{c|}{0.85} & 0.53 & \multicolumn{1}{c|}{0.84} & \multicolumn{1}{c|}{1.00} & \multicolumn{1}{c|}{0.84} & 0.67 & \multicolumn{1}{c|}{0.85} & \multicolumn{1}{c|}{0.75} & \multicolumn{1}{c|}{0.87} & 0.79 & 0.61 & 0.74 & 0.81 & 0.74 \\ \hline
{\color[HTML]{404040} \textbf{Vibrato}} & \multicolumn{1}{c|}{0.40} & \multicolumn{1}{c|}{0.50} & \multicolumn{1}{c|}{0.36} & 0.25 & \multicolumn{1}{c|}{0.68} & \multicolumn{1}{c|}{1.00} & \multicolumn{1}{c|}{0.65} & 0.50 & \multicolumn{1}{c|}{0.64} & \multicolumn{1}{c|}{0.73} & \multicolumn{1}{c|}{0.69} & 0.69 & 0.38 & 0.52 & 0.60 & 0.69 \\ \hline
{\color[HTML]{404040} \textbf{Expressive preparation}} & \multicolumn{1}{c|}{0.48} & \multicolumn{1}{c|}{0.00} & \multicolumn{1}{c|}{0.42} & 0.00 & \multicolumn{1}{c|}{0.66} & \multicolumn{1}{c|}{1.00} & \multicolumn{1}{c|}{0.64} & 0.20 & \multicolumn{1}{c|}{0.66} & \multicolumn{1}{c|}{0.71} & \multicolumn{1}{c|}{0.75} & 0.92 & 0.50 & 0.47 & 0.74 & 0.62 \\ \hline
{\color[HTML]{404040} \textbf{Freeze}} & \multicolumn{1}{c|}{0.84} & \multicolumn{1}{c|}{0.71} & \multicolumn{1}{c|}{0.83} & 0.83 & \multicolumn{1}{c|}{0.88} & \multicolumn{1}{c|}{0.67} & \multicolumn{1}{c|}{0.86} & 1.00 & \multicolumn{1}{c|}{0.86} & \multicolumn{1}{c|}{0.86} & \multicolumn{1}{c|}{0.85} & 0.86 & 0.75 & 1.00 & 0.90 & 0.86 \\ \hline
{\color[HTML]{404040} \textbf{Expressive head movement}} & \multicolumn{1}{c|}{0.17} & \multicolumn{1}{c|}{0.50} & \multicolumn{1}{c|}{0.14} & 0.50 & \multicolumn{1}{c|}{0.55} & \multicolumn{1}{c|}{1.00} & \multicolumn{1}{c|}{0.53} & 0.70 & \multicolumn{1}{c|}{0.52} & \multicolumn{1}{c|}{1.00} & \multicolumn{1}{c|}{0.56} & 0.91 & 0.50 & 0.32 & 0.63 & 0.82 \\ \hline
{\color[HTML]{404040} \textbf{Frowning}} & \multicolumn{1}{c|}{0.89} & \multicolumn{1}{c|}{0.80} & \multicolumn{1}{c|}{0.94} & 0.90 & \multicolumn{1}{c|}{0.96} & \multicolumn{1}{c|}{0.91} & \multicolumn{1}{c|}{0.98} & 0.86 & \multicolumn{1}{c|}{0.89} & \multicolumn{1}{c|}{0.83} & \multicolumn{1}{c|}{0.93} & 0.96 & 0.43 & 0.53 & 0.83 & 0.89 \\ \hline
{\color[HTML]{404040} \textbf{Physical energy}} & \multicolumn{1}{c|}{0.34} & \multicolumn{1}{c|}{0.93} & \multicolumn{1}{c|}{0.32} & 0.70 & \multicolumn{1}{c|}{0.41} & \multicolumn{1}{c|}{0.87} & \multicolumn{1}{c|}{0.41} & 0.89 & \multicolumn{1}{c|}{0.41} & \multicolumn{1}{c|}{0.91} & \multicolumn{1}{c|}{0.43} & 0.89 & 0.35 & 0.68 & 0.46 & 0.93 \\ \hline
{\color[HTML]{404040} \textbf{Upbeat in head movement}} & \multicolumn{1}{c|}{0.00} & \multicolumn{1}{c|}{0.00} & \multicolumn{1}{c|}{0.00} & 0.00 & \multicolumn{1}{c|}{0.02} & \multicolumn{1}{c|}{0.00} & \multicolumn{1}{c|}{0.02} & 0.00 & \multicolumn{1}{c|}{0.16} & \multicolumn{1}{c|}{0.00} & \multicolumn{1}{c|}{0.16} & 0.00 & 0.09 & 0.00 & 0.16 & 0.00 \\ \hline
{\color[HTML]{404040} \textbf{repositioning guitar}} & \multicolumn{1}{c|}{0.64} & \multicolumn{1}{c|}{0.00} & \multicolumn{1}{c|}{0.62} & 0.00 & \multicolumn{1}{c|}{0.76} & \multicolumn{1}{c|}{0.00} & \multicolumn{1}{c|}{0.77} & 0.00 & \multicolumn{1}{c|}{0.75} & \multicolumn{1}{c|}{0.00} & \multicolumn{1}{c|}{0.76} & 0.00 & 0.56 & 0.00 & 0.69 & 0.00 \\ \hline
{\color[HTML]{404040} \textbf{sympathetic body movement}} & \multicolumn{1}{c|}{0.00} & \multicolumn{1}{c|}{0.00} & \multicolumn{1}{c|}{0.00} & 0.00 & \multicolumn{1}{c|}{0.00} & \multicolumn{1}{c|}{0.00} & \multicolumn{1}{c|}{0.00} & 0.00 & \multicolumn{1}{c|}{0.00} & \multicolumn{1}{c|}{0.00} & \multicolumn{1}{c|}{0.00} & 0.00 & 0.00 & 0.00 & 0.00 & 0.00 \\ \hline
{\color[HTML]{404040} \textbf{normal play}} & \multicolumn{1}{c|}{0.96} & \multicolumn{1}{c|}{0.90} & \multicolumn{1}{c|}{0.98} & 0.95 & \multicolumn{1}{c|}{0.97} & \multicolumn{1}{c|}{0.83} & \multicolumn{1}{c|}{0.98} & 0.97 & \multicolumn{1}{c|}{0.95} & \multicolumn{1}{c|}{0.90} & \multicolumn{1}{c|}{0.95} & 0.92 & 0.72 & 0.81 & 0.89 & 0.95 \\ \hline
\end{tabular}
\end{table}
\end{landscape}
\begin{landscape}
\begin{table}[!ht]
%\centering
\caption{Evaluation results when trained on three videos and testing on the remaining one}
\label{tab:leaveoneout}
\begin{tabular}{|l|c|c|c|c|c|c|c|c|c|c|c|c|c|c|c|c|}
\hline
\multicolumn{1}{|r|}{\textbf{Experiments}} & \multicolumn{4}{c|}{\textbf{\begin{tabular}[c]{@{}c@{}}Train on \\ video 1,2 \& 3, \\ test on video 4\end{tabular}}} & \multicolumn{4}{c|}{\textbf{\begin{tabular}[c]{@{}c@{}}Train on \\ video 2, 3 \& 4, \\ test on video 1\end{tabular}}} & \multicolumn{4}{c|}{\textbf{\begin{tabular}[c]{@{}c@{}}Train on \\ video 1, 3\& 4, \\ test on video 2\end{tabular}}} & \multicolumn{4}{c|}{\textbf{\begin{tabular}[c]{@{}c@{}}Train on \\ video 1,2 \& 4, \\ test on video 3\end{tabular}}} \\ \hline
 & \multicolumn{16}{c|}{\textbf{Single Model + Batch balancing + Temporal Smoothing}} \\ \hline
\multicolumn{1}{|r|}{\textbf{Evaluation Metric}} & \multicolumn{2}{c|}{\textbf{Precision}} & \multicolumn{2}{c|}{\textbf{Recall}} & \multicolumn{2}{c|}{\textbf{Precision}} & \multicolumn{2}{c|}{\textbf{Recall}} & \multicolumn{2}{c|}{\textbf{Precision}} & \multicolumn{2}{c|}{\textbf{Recall}} & \multicolumn{2}{c|}{\textbf{Precision}} & \multicolumn{2}{c|}{\textbf{Recall}} \\ \hline
\multicolumn{1}{|r|}{\textbf{Stage}} & \textbf{Train} & \textbf{Test} & \textbf{Train} & \textbf{Test} & \textbf{Train} & \textbf{Test} & \textbf{Train} & \textbf{Test} & \textbf{Train} & \textbf{Test} & \textbf{Train} & \textbf{Test} & \textbf{Train} & \textbf{Test} & \textbf{Train} & \textbf{Test} \\ \hline
\textbf{Gestures} & \multicolumn{1}{l|}{} & \multicolumn{1}{l|}{} & \multicolumn{1}{l|}{} & \multicolumn{1}{l|}{} & \multicolumn{1}{l|}{} & \multicolumn{1}{l|}{} & \multicolumn{1}{l|}{} & \multicolumn{1}{l|}{} & \multicolumn{1}{l|}{} & \multicolumn{1}{l|}{} & \multicolumn{1}{l|}{} & \multicolumn{1}{l|}{} & \multicolumn{1}{l|}{} & \multicolumn{1}{l|}{} & \multicolumn{1}{l|}{} & \multicolumn{1}{l|}{} \\ \hline
{\color[HTML]{404040} \textbf{Facial   Expression}} & 0.98 & 0.39 & 0.97 & 0.56 & 0.99 & 0.61 & 0.96 & 0.55 & 0.98 & 0.23 & 0.98 & 0.20 & 0.99 & 0.44 & 0.98 & 0.51 \\ \hline
{\color[HTML]{404040} \textbf{Head related   action}} & 0.98 & 0.45 & 0.96 & 0.13 & 0.97 & 0.28 & 0.95 & 0.40 & 0.98 & 0.43 & 0.97 & 0.07 & 0.98 & 0.28 & 0.99 & 0.64 \\ \hline
{\color[HTML]{404040} \textbf{Right hand   action}} & 0.95 & 0.48 & 0.94 & 0.11 & 0.92 & 0.12 & 0.92 & 0.45 & 0.95 & 0.43 & 0.96 & 0.06 & 0.93 & 0.22 & 0.92 & 0.29 \\ \hline
{\color[HTML]{404040} \textbf{Left hand   action}} & 0.96 & 0.24 & 0.96 & 0.07 & 0.94 & 0.33 & 0.93 & 0.25 & 0.97 & 0.47 & 0.98 & 0.14 & 0.98 & 0.19 & 0.98 & 0.39 \\ \hline
{\color[HTML]{404040} \textbf{Relative   stillness}} & 0.98 & 0.60 & 0.97 & 0.03 & 0.98 & 0.02 & 0.97 & 0.08 & 0.98 & 0.54 & 0.97 & 0.07 & 1.00 & 0.40 & 0.98 & 0.25 \\ \hline
{\color[HTML]{404040} \textbf{Upper body   movement}} & 0.99 & 0.48 & 0.99 & 0.04 & 0.97 & 0.21 & 0.97 & 0.72 & 0.99 & 0.19 & 0.98 & 0.11 & 1.00 & 0.17 & 1.00 & 0.39 \\ \hline
{\color[HTML]{404040} \textbf{Normal play}} & 0.98 & 0.24 & 0.98 & 0.76 & 0.98 & 0.38 & 0.98 & 0.07 & 0.98 & 0.28 & 0.98 & 0.87 & 0.99 & 0.26 & 0.98 & 0.09 \\ \hline
\end{tabular}
\end{table}
\end{landscape}

%%%%%%%%%%%%%%%%%%%%%%%%%%%%%%%%%%%%
\begin{figure}[!ht]
\centering
  \begin{subfigure}[b]{0.4\textwidth}
    \includegraphics[scale = 0.2]{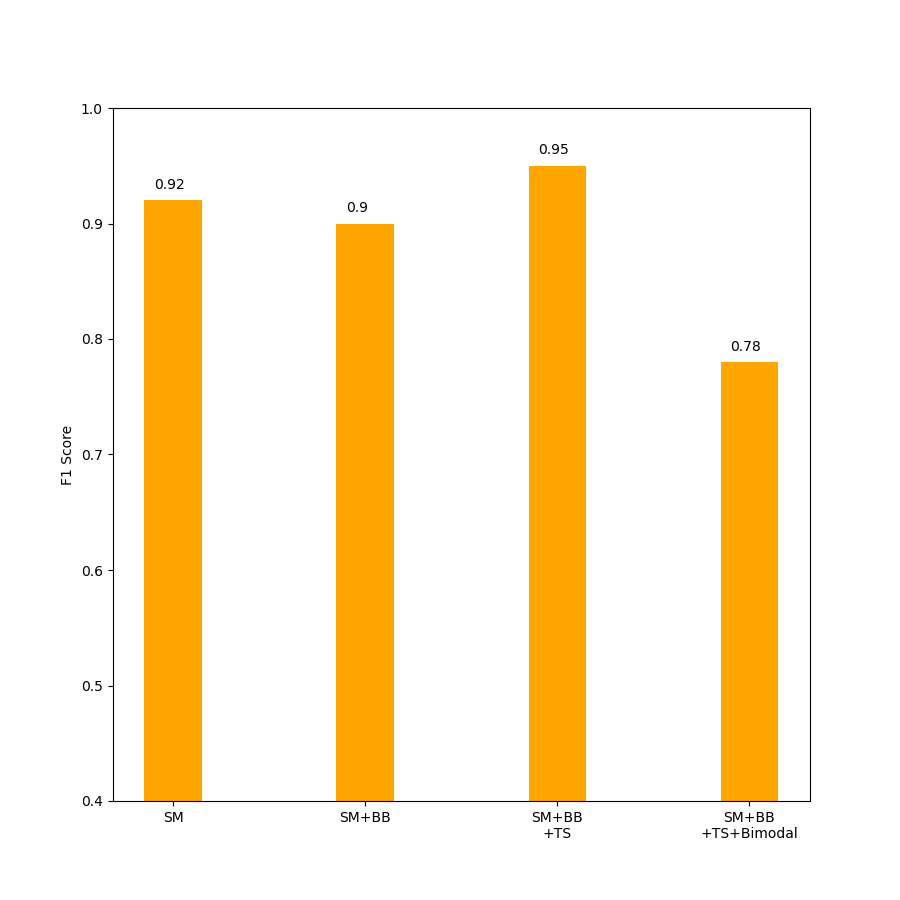}
    \caption{\centering{Evaluation on train set of model trained on all videos}}
    \label{fig:7classes_allvideos_train}
  \end{subfigure}
  \hspace{1em}
  \begin{subfigure}[b]{0.4\textwidth}
    \includegraphics[scale = 0.2]{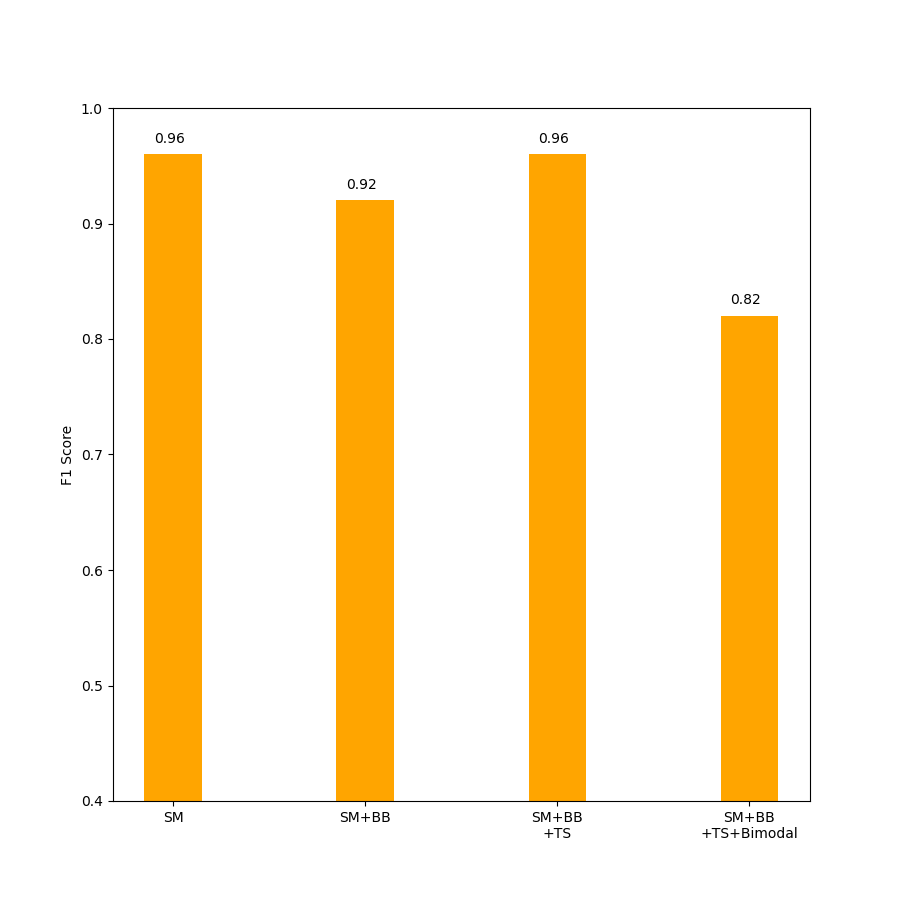}
        \caption{\centering{Evaluation on train set of model trained on video-1}}
    \label{fig:7classes_1video_train}
  \end{subfigure}
  \caption{These plots show the F-1 score evaluation results on the training set when the model is trained on (a) Video-1 + Video-2 + Video-3 + Video-4 (b) only video 1. It displays results on four experiments, that are SM, SM+BB, SM+BB+TS and Bi-modal+SM+BB+TS as detailed in Table \ref{tab:exp} for 7 classes (Table \ref{tab:7classdistribution}) }
\end{figure}
%%%%%%%%%%%%%%%%%%%%%%%%%%%%%%%%%%%%%%%

%%%%%%%%%%%%%%%%%%%%%%%%%%%%%%%%%%%%
\begin{figure}[!ht]
\centering
  \begin{subfigure}[b]{0.45\textwidth}
    \includegraphics[scale = 0.2]{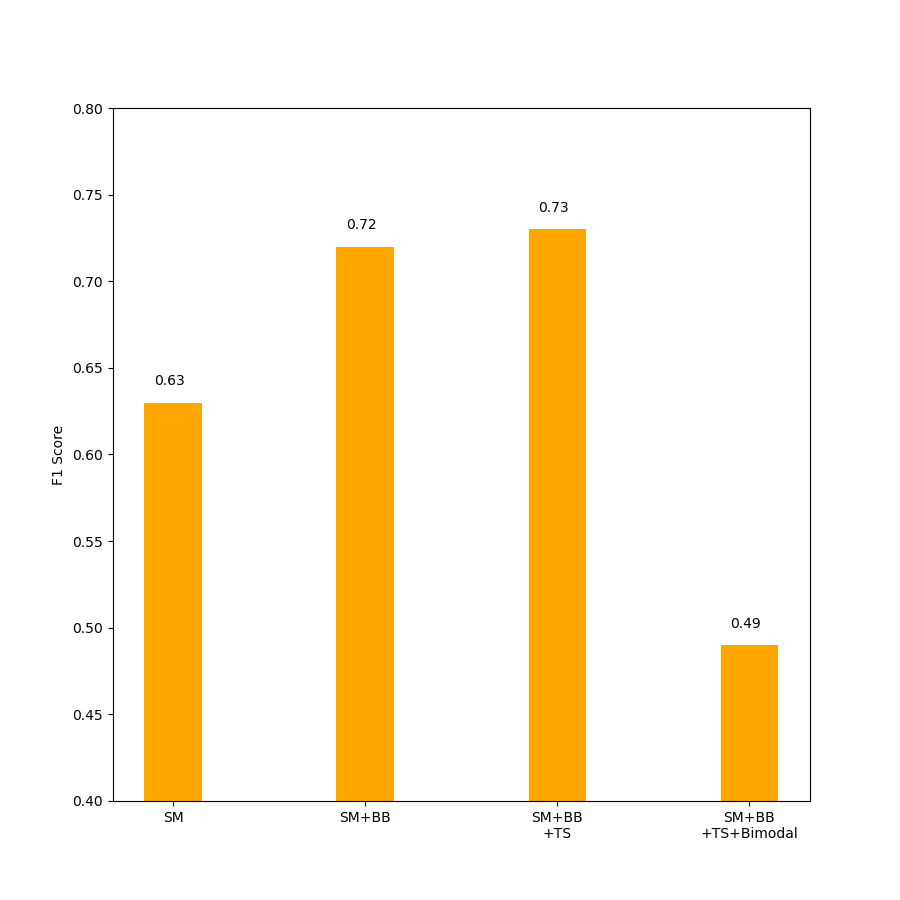}
        \caption{\centering{Evaluation on train set of model trained on all videos}}
    \label{fig:18classes_allvideos_train}
  \end{subfigure}
  \hspace{1em}
  \begin{subfigure}[b]{0.45\textwidth}
    \includegraphics[scale = 0.2]{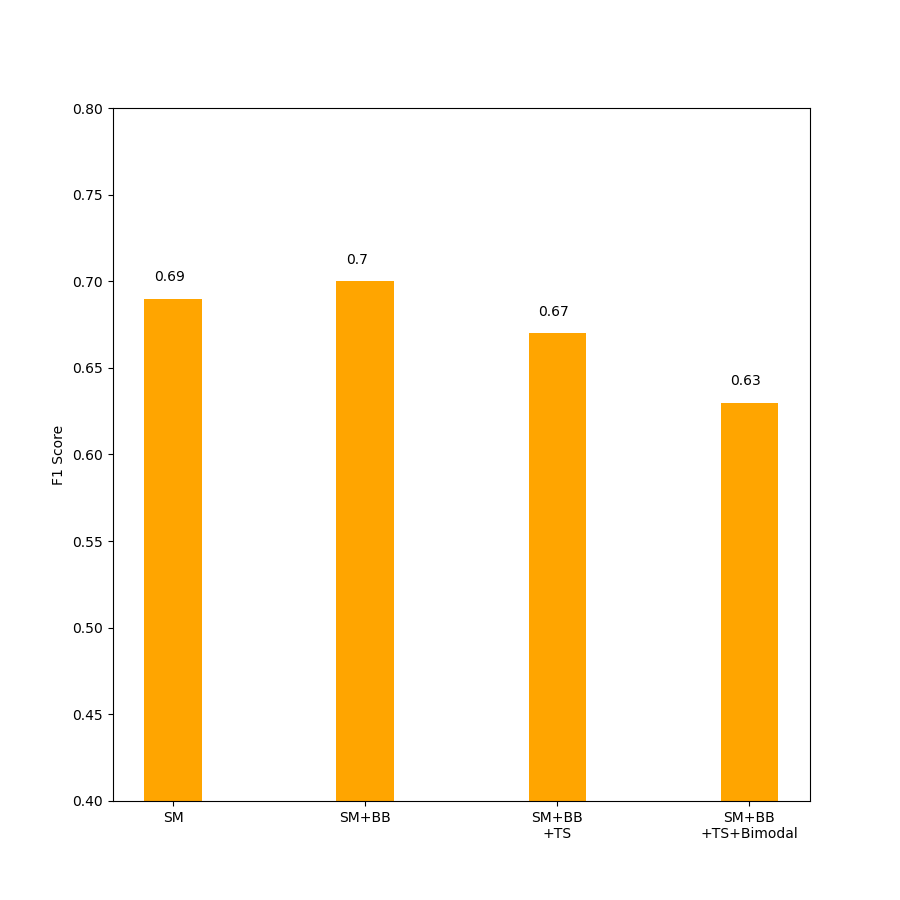}
        \caption{\centering{Evaluation on train set of model trained on video-1}}
    \label{fig:18classes_1video_train}
  \end{subfigure}
  \caption{These plots show the F-1 score evaluation results on the training set when the model is trained on (a) Video-1 + Video-2 + Video-3 + Video-4 (b) only video 1. It displays results on four experiments, that are SM, SM+BB, SM+BB+TS and Bi-modal+SM+BB+TS as detailed in Table \ref{tab:exp} for 18 classes (Table \ref{tab:18classdistribution}) }
\end{figure}
%%%%%%%%%%%%%%%%%%%%%%%%%%%%%%%%%%%%%%%

\end{document}